\documentclass[10pt,twocolumn,letterpaper]{article}

\usepackage{iccv}      %

\usepackage{xcolor}
\usepackage{amsmath}
\usepackage{amssymb}
\usepackage{array}
\usepackage{multirow}
\usepackage{color}
\usepackage{colortbl}
\usepackage{framed}
\usepackage{bm}
\usepackage{bbm}
\usepackage{xspace}
\usepackage{enumitem}
\usepackage{algorithm}
\usepackage{listings}
\usepackage{url}
\usepackage{booktabs}
\usepackage{pifont} 
\usepackage{xfrac}
\usepackage{rotating}
\usepackage{makecell}
\usepackage{stfloats}

\newcommand\blfootnote[1]{%
  \begingroup
  \renewcommand\thefootnote{}\footnote{#1}%
  \addtocounter{footnote}{-1}%
  \endgroup
}

\definecolor{Gray}{gray}{0.9}
\definecolor{DarkGray}{gray}{0.3}
\definecolor{LightCyan}{rgb}{0.88,0.95,1}
\definecolor{LightCyan}{rgb}{0.87, 0.94, 0.84}
\definecolor{LighterGreen}{rgb}{0.94, 0.96, 0.93}

\definecolor{blond}{rgb}{0.98, 0.94, 0.75}
\definecolor{green}{rgb}{0.0, 0.62, 0.42}
\definecolor{green_into_the_wild}{rgb}{0.0, 1.0, 0.0}
\definecolor{light_green_into_the_wild}{rgb}{0.3, 0.7, 0.3}
\definecolor{red_into_the_wild}{rgb}{1.0, 0.0, 0.0}
\definecolor{yellow_into_the_wild}{rgb}{0.7, 0.7, 0.0}
\definecolor{light_blue_into_the_wild}{rgb}{0.0, 0.5, 1.0}
\definecolor{fuchsia_into_the_wild}{rgb}{0.6, 0.0, 0.6}
\definecolor{light_fuchsia_into_the_wild}{rgb}{0.4, 0.2, 0.2}
\definecolor{lightgray}{gray}{0.93}

\newcommand{\cmark}{\ding{51}}%
\newcommand{\xmark}{\ding{55}}%

\newcommand{\ours}{Talk2DINO\xspace}

\newcommand{\tit}[1]{\smallbreak\noindent\textbf{#1.}}
\newcommand{\tinytit}[1]{\noindent\textbf{#1.}}

\DeclareMathOperator*{\argmax}{argmax}

\definecolor{iccvblue}{rgb}{0.21,0.49,0.74}
\usepackage[pagebackref,breaklinks,colorlinks,allcolors=iccvblue]{hyperref}

\def\paperID{10000}
\def\confName{ICCV}
\def\confYear{2025}

\title{\textit{Talking to DINO}: Bridging Self-Supervised Vision Backbones with Language\\for Open-Vocabulary Segmentation}

\author{Luca Barsellotti$^{*1}$, Lorenzo Bianchi$^{*2,3}$, Nicola Messina$^2$, Fabio Carrara$^2$, Marcella Cornia$^1$,\\Lorenzo Baraldi$^1$, Fabrizio Falchi$^2$, Rita Cucchiara$^1$\\
$^1$University of Modena and Reggio Emilia, Italy\quad$^2$ISTI-CNR, Italy\quad$^3$University of Pisa, Italy\\
$^1${\tt\small \{name.surname\}@unimore.it}\quad\quad$^2${\tt\small \{name.surname\}@isti.cnr.it}}

\begin{document}
\maketitle

\begin{abstract}
Open-Vocabulary Segmentation (OVS) aims at segmenting images from free-form textual concepts without predefined training classes. While existing vision-language models such as CLIP can generate segmentation masks by leveraging coarse spatial information from Vision Transformers, they face challenges in spatial localization due to their global alignment of image and text features. Conversely, self-supervised visual models like DINO excel in fine-grained visual encoding but lack integration with language. To bridge this gap, we present \ours, a novel hybrid approach that combines the spatial accuracy of DINOv2 with the language understanding of CLIP. Our approach aligns the textual embeddings of CLIP to the patch-level features of DINOv2 through a learned mapping function without the need to fine-tune the underlying backbones. At training time, we exploit the attention maps of DINOv2 to selectively align local visual patches with textual embeddings. We show that the powerful semantic and localization abilities of \ours can enhance the segmentation process, resulting in more natural and less noisy segmentations, and that our approach can also effectively distinguish foreground objects from the background. Experimental results demonstrate that \ours achieves state-of-the-art performance across several unsupervised OVS benchmarks.
Source code and models are publicly available at: {\small\url{https://lorebianchi98.github.io/Talk2DINO/}}.
\blfootnote{$^*$Equal contribution.}
\end{abstract}
\vspace{-.55cm}
    
\section{Introduction}
\label{sec:intro}

\begin{figure}[t]
    \centering
    \includegraphics[width=0.98\linewidth]{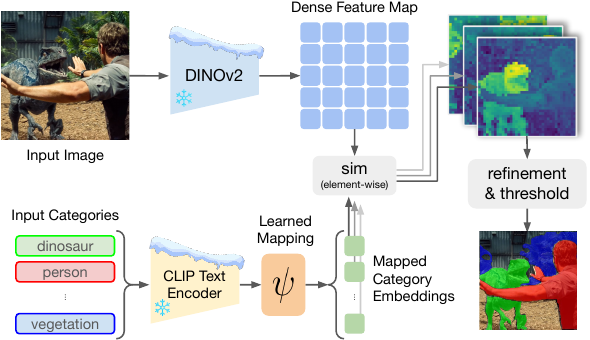}
    \vspace{-.1cm}
    \caption{\textbf{Overview.} Our approach aligns the embedding spaces of CLIP and DINOv2 through a learned mapping function. This results in fine-grained visual encodings, which can be mapped to language to provide natural and less noisy semantic segmentation.}
    \label{fig:first-page}
    \vspace{-.5cm}
\end{figure}

Open-Vocabulary Segmentation (OVS)~\cite{zhao2017open} is a fundamental task in Computer Vision that aims to partition an input image into a set of coherent regions based on concepts provided at inference time~\cite{ghiasi2022scaling,li2022languagedriven,bucher2019zero}. The set of concepts used to partition the image is usually provided in freeform natural language, which effectively unchains the methods from working with a specific fixed set of classes provided at training time. Solving this task requires a fine-grained understanding of the semantic interconnections between image pixels and the meaning conveyed by natural language.

While previous works employed pixel-level annotations as a source of supervision~\cite{xie2024sed,xu2023side,zou2023generalized}, a recent trend in the field is to tackle this problem in an unsupervised manner~\cite{xu2022groupvit,cha2023learning,barsellotti2024training} leveraging the correlations learned by state-of-the-art backbones. Contrastive embedding spaces like CLIP~\cite{radford2021learning}, in particular, have demonstrated good performance on tasks that demand a holistic understanding of vision and language modalities~\cite{zhong2022regionclip,li2022grounded,minderer2024scaling,zhou2022detecting}, and have therefore been employed for unsupervised OVS~\cite{zhou2022extract,wang2024sclip,lan2024proxyclip}. Although CLIP-based backbones exhibit strong cross-modal capabilities, they are primarily trained to predict a global similarity score between text and images, which limits their spatial understanding and consequently affects tasks based on dense predictions. Recent efforts have tackled this limitation by introducing architectural modifications~\cite{zhou2022extract,wang2024sclip,hajimiri2025pay}. However, the spatial understanding constraints imposed by the training modality hinder the effectiveness of such backbones in OVS and highlight the potential benefits of exploring alternative models with enhanced perceptual capabilities.

Self-supervised vision-only backbones like DINO and DINOv2~\cite{caron2021emerging,oquab2023dinov2,darcet2024vision} have instead shown remarkable abilities in capturing fine-grained and localized spatial features without the reliance on annotated data. Specifically, the self-attention mechanism in such backbones generates attention maps that consistently pinpoint relevant regions within the image and has been widely leveraged for foreground object segmentation~\cite{simeoni2021localizing,simeoni2023unsupervised,wang2023tokencut,wang2022self,wang2023cut}. While this property makes them a powerful tool for tasks requiring fine-grained spatial understanding, the embedding space derived from visual self-supervised networks is not inherently aligned with textual concepts, making it incompatible with the OVS task.

To close the gap between vision-language and self-supervised embedding spaces, we propose \ours, a method that combines the spatial sensitivity of DINOv2 with the text-image alignment capabilities of CLIP, enabling a highly localized multimodal image understanding. Our approach, depicted in Fig.~\ref{fig:first-page}, learns a mapping function that translates the text embeddings of CLIP to interact with the patch-level embeddings of DINOv2 without fine-tuning the underlying backbones. Our alignment mechanism enhances text-image correspondence by exploiting the
self-attention heads of DINOv2 to highlight diverse regions of the image. During training, we weight visual patch embeddings using the attention maps of DINOv2, dynamically selecting the head that best aligns with the provided caption. This embedding is then used to maximize similarity with the caption representation through contrastive learning. At inference time, we calculate the similarity of visual patches to each textual label, including a novel background cleaning procedure that weights class scores using attention maps.

Our approach demonstrates state-of-the-art performance
in unsupervised OVS with minimal parameter learning. Our results show that a self-supervised vision-only encoder can generate embeddings with semantic properties akin to textual representations, opening up new pathways for addressing spatial understanding limitations in CLIP-like models~\cite{tong2024eyes}.
To sum up, our main contributions are as follows:
\begin{itemize}
    \item We propose \ours, the first model that provides language properties to DINOv2 by mapping the CLIP textual embeddings into the DINOv2 space through a non-linear warping function.
    \item Our proposed model employs a novel training schema that selects the most relevant visual self-attention head and does not need fine-tuning on the backbones.
    \item We showcase the capabilities of \ours on unsupervised OVS by devising a computationally efficient inference pipeline that also employs a novel approach based on DINOv2 self-attention to improve distinguishing foreground categories from the background.
    \item Experimentally, we show that \ours achieves state-of-the-art results in standard OVS benchmarks, demonstrating the effectiveness of the proposed approach.
\end{itemize}

\section{Related Work}
\label{sec:related}

\tinytit{Vision-Language Pre-Training} In the last years, vision-language pre-training has gained increasing interest by learning multimodal representations that can be easily transferred to downstream tasks~\cite{fang2023eva,jia2021scaling}. The popular CLIP model~\cite{radford2021learning} is trained on large-scale web-scraped data through a contrastive objective, which maximizes the similarity of the representations of corresponding image-text pairs while minimizing the similarity of the other pairs within a batch. This approach has demonstrated remarkable zero-shot classification and retrieval performance. However, learning a multimodal representation by matching global images and texts poses challenges in localizing regions with their corresponding text, showing limited performance in dense prediction tasks~\cite{mukhoti2023open,ranasinghe2023perceptual,zhong2022regionclip,bica2024improving,li2025closer}.

\tit{Open-Vocabulary Segmentation}
In zero-shot segmentation, a segmentation model is trained on a set of \textit{seen} classes and must generalize to \textit{unseen} classes. The first attempts of OVS inherit this paradigm by training the model on a closed set of classes for which segmentation data is available and exploiting vision-language pre-training to extend their capabilities on an open set of classes through text~\cite{xu2023side,cho2024cat,xie2024sed,han2023open,zou2023generalized}.
The two-step approach represents the most popular framework in this research direction, which proposes class-agnostic regions, trained on segmentation masks, and provides them to CLIP to be aligned with textual classes~\cite{ding2022decoupling,ghiasi2022scaling,liang2023open,xu2022simple,xu2023open}. However, this approach is affected by performance gaps between seen and unseen classes and presents a significant computational overhead.

On the contrary, another research direction investigates how to force the segmentation capabilities to emerge without relying on direct supervision from segmentation data. Cha~\etal~\cite{cha2023learning} identify this setting as unsupervised OVS and propose a unified evaluation protocol. Approaches in this line can be categorized into two main groups: (i) training-free methods that propose architectural adaptations to enable pre-trained models to produce localized multimodal features~\cite{zhou2022extract,lan2024proxyclip,wysoczanska2024clipdiy,bousselham2024grounding,tu2023closer,wang2024sclip,hajimiri2025pay,lan2024clearclip,sun2024clip}, and (ii) weakly-supervised methods that leverage large sets of image-text pairs with dedicated learning strategies that aim to improve the correspondence between regions and texts~\cite{xu2022groupvit,cha2023learning,ranasinghe2023perceptual,xu2023learning,luo2023segclip,ren2023viewco}. Our model lies in the latter category since we exploit a weak supervision to learn how to bridge the CLIP and DINOv2 feature spaces.

\newcommand{\R}{\mathbb{R}}

\tit{Self-Supervised Backbones}
Recent advances in self-supervised learning have led to models that showcase impressive matching and localization capabilities. In particular, the DINO family of models~\cite{caron2021emerging,oquab2023dinov2,darcet2024vision} employs Vision Transformers~\cite{dosovitskiy2021an} trained with self-distillation, and has shown that patch-level features learned through self-supervision can yield semantic information. Also, a strong relationship has been observed between self-attention activations and foreground regions of the image. Hence, researchers have recently focused on the usage of DINO for the unsupervised segmentation task~\cite{simeoni2021localizing,simeoni2023unsupervised,wang2023tokencut,wang2022self,wang2023cut}. 

While the aforementioned models are purely visual, recent works have also focused on connecting self-supervised feature spaces with textual representations for OVS. ReCo~\cite{shin2022reco}, OVDiff~\cite{karazija2023diffusion}, FOSSIL~\cite{barsellotti2024fossil}, and FreeDA~\cite{barsellotti2024training} are training-free approaches that build prototypes in the visual space according to pre-defined textual categories. CLIP-DINOiser~\cite{wysoczanska2024clip} demonstrates that CLIP can be fine-tuned with the supervision of DINO to retain improved localization capabilities. LaVG~\cite{kang2024defense} employs DINO to propose class-agnostic regions and computes the average embedding from CLIP for each region. ProxyCLIP~\cite{lan2024proxyclip} proposes a proxy attention module to integrate DINO features with \textit{values} from the last attention layer of CLIP, thus employing the two visual backbones at prediction time. Our method is closely related to this research field since we aim to combine the multimodal understanding capabilities of CLIP with the localization properties of DINOv2. However, in contrast to these methods, we propose to directly map the textual representations from the textual encoder of CLIP to the DINOv2 space, and demonstrate that our approach sets a new state-of-the-art without relying on multiple visual backbones or on external sources of knowledge.

\begin{figure*}
    \centering
    \includegraphics[width=0.98\linewidth]{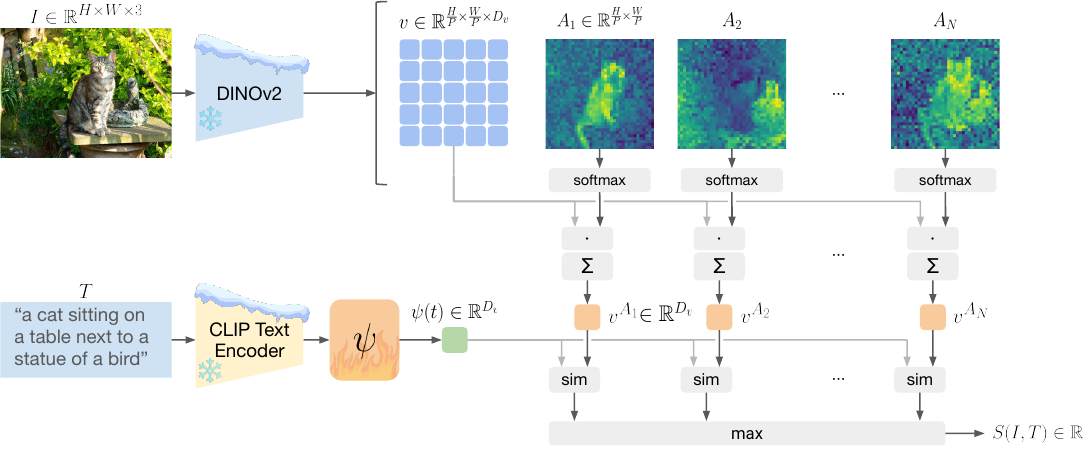}
    \vspace{-0.15cm}
    \caption{\textbf{Overview of the training methodology of \ours.} We learn a projection $\psi(\cdot)$ that maps the CLIP textual embeddings to the visual embedding space of DINOv2. Given the dense feature map and the attention maps extracted from DINOv2, we generate $N$ visual embeddings by computing a weighted average of the feature map with each attention map. We then compute the similarity between each visual embedding and the projected text embedding,
    and use the maximum similarity as the global alignment score.
    }
    \label{fig:training}
    \vspace{-0.35cm}
\end{figure*}

\section{Proposed Method}
\label{sec:method}

\subsection{Preliminaries}
\tinytit{Task Definition}
Open-vocabulary segmentation aims to segment objects of interest defined through natural language at inference time.
Let $I \in \R^{H \times W \times 3}$ be an image and $v(I) \in \R^{\frac{H}{P} \times \frac{W}{P} \times D_v}$ its dense feature map extracted by a Transformer-based visual backbone with input patch size $P$ and dimensionality of embedding space $D_v$. Let $\{T_j\}_{j=1,...,M}$ be a set of arbitrary textual categories and $t(T_j) \in \R^{D_t}$ their embeddings extracted by a pre-trained textual backbone.
To simplify the notation, in the following, we will refer to $v(I)$ as $v$ and to $t(T_j)$ as $t_j$. Assuming a multimodal setting in which $D_t=D_v$, we could define the similarity map $\mathcal{S}(I,T_j) \in \R^{\frac{H}{P} \times \frac{W}{P}}$ for the image $I$ and category $T_j$ as the cosine similarity between $t_j$ and each spatial entry of $v$. Formally, the similarity map is defined as
\begin{equation}
    \label{eq:similarity_map}
    \mathcal{S}(I,T_j)_{[h,w]} = \frac{v_{[h,w]} \cdot t_j^\intercal}{||v_{[h,w]}|| \,||t_j||},
\end{equation}
where $\cdot_{[h,w]}$ represents indexing over spatial axes. The full resolution similarity map $\hat{\mathcal{S}}(I,T) \in \R^{H \times W}$ is recovered by upsampling $\mathcal{S}(I,T)$ (\eg,~via bilinear interpolation). Segmentation masks $\mathcal{M}(I, T_1,...,T_M)$ are then derived by assigning pixels to the category with the highest similarity score, \ie,
\begin{equation}
\label{eq:segmentation_masks}
    \mathcal{M}(I, T_1,...,T_M)_{[h,w]} = \argmax_{j=1,...,M}{\hat{\mathcal{S}}(I,T_j)_{[h,w]}}.
\end{equation}
In order for Eq. \ref{eq:similarity_map}, and therefore the segmentation from Eq. \ref{eq:segmentation_masks},
to work correctly, not only the two $v$ and $t$ spaces should share the same dimensionality, but they should also be constructed so that they also share the same semantics.

\tit{CLIP and DINO Duality}
Existing vision-language models trained on image-text pairs, such as CLIP~\cite{radford2021learning}, can naturally fit the formulation mentioned above, as they provide dense visual and textual embeddings in the same space. However, while CLIP can correctly align global features coming from texts and images (\ie, through the similarities corresponding to \texttt{CLS} tokens), it lacks a precise alignment between the textual feature $t$ and spatial patches $v$.

Conversely, purely visual self-supervised backbones like DINOv2~\cite{oquab2023dinov2} have shown remarkable semantic and local consistency of spatial embeddings, enabling agnostic image segmentation~\cite{simeoni2023unsupervised,simeoni2021localizing,wang2023tokencut}. These abilities occur naturally in the last attention layer of DINOv2, where the attention maps computed between the \texttt{CLS} token and the spatial tokens
align with relevant objects within the image (see Fig.~\ref{fig:training}). Despite the remarkable results observed on image-only tasks, DINOv2 lacks a solid bridge with natural language, making it impossible to directly compute the similarities with the text features, as expressed in Eq.~\ref{eq:similarity_map}.

While DINOv2 and CLIP embedding spaces are traditionally thought as being uncorrelated spaces, we show that the CLIP textual embedding space can be projected into the DINOv2 space through a learnable nonlinear warping.

\subsection{Augmenting DINO with Semantics}
\tit{Warping CLIP Embedding Space}
We learn a projection $\psi: \mathbb{R}^{D_t} \rightarrow \mathbb{R}^{D_v}$ to map textual embeddings $t$ into the space of the visual patch embeddings $v$ of DINOv2, leveraging weak supervision from image-text pairs. We build the
projection $\psi$ applied to textual features by composing two affine transformations with a hyperbolic tangent activation, which provides nonlinear warping. Formally,
\begin{equation}
    \psi(t) = \bm{W}^\intercal_b \left( \text{tanh} (\bm{W}^\intercal_a t + b_a \right)) + b_b,
\end{equation}
where $\bm{W}_a \in \mathbb{R}^{D_t \times D_v}$ and $\bm{W}_b \in \mathbb{R}^{D_v \times D_v}$ are learnable projection matrices and $b_*$ are learnable bias vectors.

\tit{Mapping DINO to the Warped CLIP Space}
To learn the nonlinear projection $\psi$, we exploit the intrinsic segmentation capability of DINOv2 to identify the precise spatial subsets of $v$ to which $\psi(t)$ should be aligned with.

Specifically, we first extract the $N$ attention maps $A_i \in \R^{\frac{H}{P} \times \frac{W}{P}}$ (one for each of the $i = 1, ..., N$ heads) which DINOv2 computes between the \texttt{CLS} vector and its patch features from the last layer. One of the key features of DINOv2 is that each $A_i$ highlights different semantic regions within the image.
For each attention map $A_i$, we compute a visual embedding $v^{A_i} \in \R^{D_v}$ as a weighted average of the dense feature map $v$, emphasizing the spatial areas that $A_i$ highlights. We then calculate the cosine similarity between each $v^{A_i}$ and the projected text embedding $\psi(t)$, resulting in $N$ similarity scores. Formally, the cosine similarity score between a head and the text embedding is defined as
\begin{align}
\text{sim}(v^{A_i}, t) &= \frac{v^{A_i} \cdot \psi(t)^\intercal}{||v^{A_i}|| \, ||\psi(t)||}, \\ 
\label{eq:avg_self_attn} 
\text{with } v^{A_i} &= \sum_{h, w} v_{[h, w]} \text{softmax}(A_{i})_{[h, w]}.
\end{align}
To obtain the most relevant score for alignment, we apply a selection function over the similarity scores obtained for different heads. In particular, we choose the maximum similarity $\max_{i=1, ..., N} \text{sim}(v^{A_i}, t)$ score across all heads, therefore promoting a robust alignment between textual and visual representations that adapts to the most salient visual features corresponding to the text query.

\tit{Training Procedure}
To optimize the alignment between text and visual embeddings, we employ the InfoNCE loss, which leverages similarity scores across a batch of image-text pairs. For each pair $(I_i, T_i)$, we compute similarity scores between the projected text embedding $\psi(t_i)$ and the maximally-activated visual embedding $\tilde{v_i}$, where $\tilde{v_i}$ is the visual embedding derived from the most relevant attention head for the corresponding text $t_i$, \ie, 
\begin{equation}
\tilde{v}_i = v^{A_j}_i \; | \; j=\argmax_{k=1,\dots,N} \text{sim}(v_i^{A_k}, t_i).
\end{equation}

Treating the true image-text pair as the positive instance and the remaining pairs within the batch as negatives, this contrastive approach drives the model to increase similarity for matching pairs and decrease it for non-matching pairs. Formally, the InfoNCE loss $\mathcal{L}_{\text{InfoNCE}}$ for a batch of $B$ image-text pairs is defined as
\begin{align*} \mathcal{L}_{\text{InfoNCE}} = -\frac{1}{2B} \sum_{i=1}^{B} \log \frac{\exp(\text{sim}(\tilde{v}_i, t_i))}{\sum_{j=1}^{B} \exp(\text{sim}(\tilde{v}_j, t_i))} \\ -\frac{1}{2B} \sum_{i=1}^{B} \log \frac{\exp(\text{sim}(\tilde{v}_i, t_i))}{\sum_{j=1}^{B} \exp(\text{sim}(\tilde{v}_i, t_j))}. \end{align*}
This formulation effectively strengthens alignment by maximizing the similarity for true pairs and minimizing it for mismatched pairs across the batch.

\begin{figure}[t]
    \centering
    \includegraphics[width=0.98\linewidth]{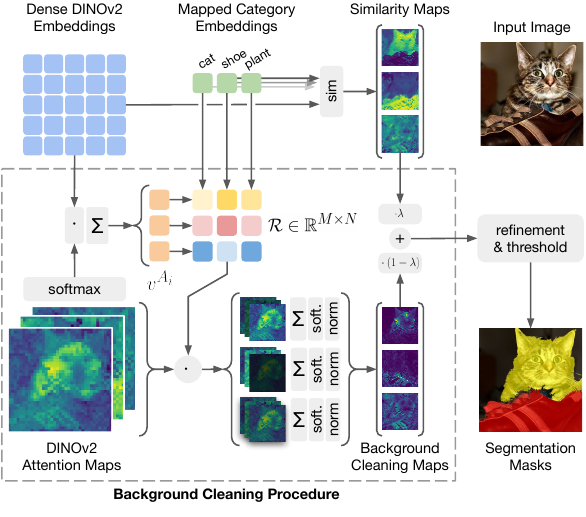}
    \vspace{-.2cm}
    \caption{\textbf{Inference procedure.} 
    At the top, we compute the similarity between mapped text embeddings and the DINOv2 patches to produce the initial similarity maps. In the bottom part, we produce a background cleaning map for each class derived from the different DINOv2 attention heads. We obtain the final enhanced similarity map of each category through a convex combination of the similarity and background cleaning maps. 
    The output segmentation then results from the final refinement and thresholding steps.
    }
    \label{fig:inference_background}
    \vspace{-.2cm}
\end{figure}

\tit{Inference}
The projection learned during the training procedure that warps the CLIP embedding space into the DINOv2 space enables the textual embeddings to be directly comparable with the dense feature embeddings from DINOv2. Hence, at inference time, given an image $I \in \R^{H \times W \times 3}$ and a set of textual arbitrary categories $\{T_j\}_{j=1,...,M}$, we can obtain the segmentation masks as defined in Eq.~\ref{eq:segmentation_masks} by considering the projected text embeddings $\psi(t_j)$ in the similarity map computation from Eq.~\ref{eq:similarity_map}.

\subsection{Identifying Background Regions}
\label{sec:cleaning}
An additional challenge that OVS approaches need to face, especially when tasked with benchmarks like Pascal VOC~\cite{pascal-voc-2012} and COCO Objects~\cite{caesar2018coco}, is that of identifying ``background'' regions, \ie~regions that do not belong to the set of categories considered in the benchmark. The standard approach consists in applying a threshold on the similarity or probability score to identify where the model is not certain about the predicted category and classify these locations as background. However, previous works~\cite{karazija2023diffusion,wysoczanska2024clip,wysoczanska2024clipdiy} have introduced custom approaches to improve the capabilities of the model in recognizing the background.

Following this line, we propose a background cleaning procedure, depicted in Fig.~\ref{fig:inference_background}, that is based on the capabilities of the DINOv2 backbone in focusing on coherent areas and highlighting the foreground through the self-attention heads. Specifically, given $N$ attention maps $A_i \in \R^{\frac{H}{P} \times \frac{W}{P}}$ and $M$ projected textual embeddings of classes $\psi(t_j)$, we first compute the average visual embeddings $v^{A_i}$ as in Eq.~\ref{eq:avg_self_attn}. Similarly to the training procedure, we then compute the similarity between each $v^{A_i}$ and $\psi(t_j)$, resulting in a matrix of similarity scores $\mathcal{R} \in \R^{M \times N}$, which is additionally normalized row-wise through a softmax operation. These scores represent how much each self-attention head is related to each textual category. Formally, $\mathcal{R}$ is defined as
\begin{align}
    \mathcal{R} &= [ \mathcal{R}_1, ..., \mathcal{R}_j, ..., \mathcal{R}_M]^\intercal, \text{ with } \\
    \mathcal{R}_j &= \text{softmax}\left(\text{sim}(v^{A_1},\psi(t_j)), ..., 
    \text{sim}(v^{A_N},\psi(t_j))\right). \nonumber
\end{align}
Then, for each category $T_j$ we compute its average attention map $\mathcal{F}_j \in \R^{\frac{H}{P} \times \frac{W}{P}}$ as
\begin{equation}
    \mathcal{F}_j(A_i, \mathcal{R}_{ij}) = \sum_{i=1}^N \mathcal{R}_{ij} A_{i},
\end{equation}
and normalize $\mathcal{F}$ by applying a softmax normalization over both spatial axes, and linearly re-projecting its values in the range $[\min_{j,h,w}\mathcal{S}(I, T_j)_{[h,w]}, \max_{j,h,w}\mathcal{S}(I, T_j)_{[h,w]}]$, where $\mathcal{S}(\cdot)$ is the similarity map defined in Eq.~\ref{eq:similarity_map}. We exploit the resulting normalized average attention per category to shape the similarity map by activating the foreground region and deactivating the background. The resulting shaped similarity map $\overline{\mathcal{S}} \in \R^{H \times W}$ is defined as
\begin{equation}
    \overline{\mathcal{S}}(I, T_j)_{[h,w]} = \lambda \mathcal{S}(I, T_j)_{[h,w]} + (1-\lambda) \mathcal{F}_{j,[h, w]},
\end{equation}
where $\lambda$ is a hyperparameter representing the relevance of the background shaping in computing the segmentation masks. The background mask is then identified as the collection of pixels for which the shaped similarity map is lower than a threshold across all semantic categories.

\begin{table*}[!t]
\centering
\setlength{\tabcolsep}{.25em}
\resizebox{\textwidth}{!}{
\begin{tabular}{lcc c ccccccccc c ccccccccc}
\toprule
& & & & \multicolumn{9}{c}{\textbf{ViT-Base (mIoU)}} & & \multicolumn{9}{c}{\textbf{ViT-Large (mIoU)}} \\
\cmidrule{5-13} \cmidrule{15-23}
\textbf{Model} & \textbf{Visual Encoder} & \textbf{Frozen} & & V20 & C59 & Stuff & City & ADE & V21 & C60 & Object & \textbf{Avg} & & V20 & C59 & Stuff & City & ADE & V21 & C60 & Object & \textbf{Avg} \\
\midrule
\rowcolor{lightgray}
\multicolumn{2}{l}{\textit{without Mask Refinement}} & & & & & & & & & & & & & & & & & & & & & \\
GroupViT~\cite{xu2022groupvit} & Custom ViT & \xmark & & 79.7 & 23.4 & 15.3 & 11.1 & 9.2 & 50.4 & 18.7 & 27.5 & 29.4 & & - & - & - & - & - & - & - & - & - \\
ReCo~\cite{shin2022reco} & CLIP & \xmark & & 57.7 & 22.3 & 14.8 & 21.1 & 11.2 & 25.1 & 19.9 & 15.7 & 23.5 & & - & - & - & - & - & - & - & - & - \\
TCL~\cite{cha2023learning} & CLIP & \xmark & & 77.5 & 30.3 & 19.6 & 23.1 & 14.9 & 51.2 & 24.3 & 30.4 & 33.9 & & - & - & - & - & - & - & - & - & - \\
SILC~\cite{naeem2024silc} & Custom ViT & \xmark & & 77.5 & 31.6 & 20.8 & 26.9 & 19.3 & - & - & - & - & & - & - & - & - & - & - & - & - & - \\
MaskCLIP~\cite{zhou2022extract} & CLIP & \cmark & & 74.9 & 26.4 & 16.4 & 12.6 & 9.8 & 38.8 & 23.6 & 20.6 & 27.9 & & 29.4 & 12.4 & 8.8 & 11.5 & 7.2 & 23.3 & 11.7 & 7.2 & 13.9 \\
CLIP-DIY~\cite{wysoczanska2024clipdiy} & CLIP+DINO & \cmark & & 79.7 & 19.8 & 13.3 & 11.6 & 9.9 & 59.9 & 19.7 & 31.0 & 30.6 & & - & - & - & - & - & - & - & - & - \\
SCLIP~\cite{wang2024sclip} & CLIP & \cmark & & 80.4 & 34.2 & 22.4 & 32.2 & 16.1 & 59.1 & 30.4 & 30.5 & 38.2 & & 70.6 & 25.2 & 17.6 & 21.3 & 10.9 & 44.0 & 22.3 & 26.9 & 29.9 \\
CLIP-DINOiser~\cite{wysoczanska2024clip} & CLIP & \cmark & & 80.9 & 35.9 & 24.6 & 31.1 & 20.0 & \textbf{62.1} & 32.4 & 34.8 & 40.2 & & - & - & - & - & - & - & - & - & - \\
ClearCLIP~\cite{lan2024clearclip} & CLIP & \cmark & & 80.9 & 35.9 & 23.9 & 30.0 & 16.7 & 51.8 & 32.6 & 33.0 & 38.1 & &  80.0 & 29.6 & 19.9 & 27.9 & 15.0 & - & - & - & - \\
NACLIP~\cite{hajimiri2025pay} & CLIP & \cmark & & 79.7 & 35.2 & 23.3 & 35.5 & 17.4 & 58.9 & 32.2 & 33.2 & 39.4 & & 78.7 & 32.1 & 21.4 & 31.4 & 17.3 & 52.2 & 28.7 & 29.9 & 36.5 \\
\texttt{dino.txt}~\cite{jose2025dinov2} & DINOv2(reg) & \cmark & & -  & -  & -  & -  & -  & -  & -  & - & - & & 62.1 & 30.9 & 20.9 & 32.1 & 20.6 & - & - & - & - \\
FreeDA~\cite{barsellotti2024training} & DINOv2 & \cmark & & 77.1 & 37.1 & 24.9 & 34.0 & 19.5 & 51.7 & 32.6 & 24.4 & 37.7 & & 71.8 & 35.4 & 24.2 & 32.3 & 19.4 & 44.9 & 31.1 & 24.6 & 35.5 \\ 
FreeDA~\cite{barsellotti2024training} & CLIP+DINOv2 & \cmark & & 84.3 & 39.7 & 25.7 & 34.1 & 20.8 & 51.8 & 35.3 & 36.3 & 41.0 & & 85.7 & \textbf{39.7} & 26.3 & 33.6 & 21.4 & 44.1 & 34.8 & 33.9 & 39.9 \\ 
ProxyCLIP~\cite{lan2024proxyclip} & CLIP+DINOv2(reg) & \cmark & & 83.0 & 37.2 & 25.4 & 33.9 & 19.7 & 58.6 & 33.8 & 37.4 & 41.1 & & 85.2 & 36.2 & 24.6 & 35.2 & 21.6 & 56.6 & 33.0 & 36.7 & 41.1 \\
ProxyCLIP~\cite{lan2024proxyclip} & CLIP+DINO & \cmark & & 80.3 & 39.1 & 26.5 & \textbf{38.1} & 20.2 & 61.3 & \textbf{35.3} & 37.5 & 42.3 & & 83.2 & 37.7 & 25.6 & \textbf{40.1} & \textbf{22.6} & \textbf{60.6} & \textbf{34.5} & \textbf{39.2} & \textbf{42.9} \\
\rowcolor{LightCyan}
\textbf{\ours (Ours)} & DINOv2(reg) & \cmark & & \textbf{87.1} & \textbf{39.8} & \textbf{28.1} & 36.6 & \textbf{21.1} & 61.5 & 35.1 & \textbf{41.0} & \textbf{43.8} & & \textbf{87.1} & 39.1 & \textbf{27.0} & 35.8 & 21.1 & 60.1 & 34.2 & 37.6 & 42.8 \\
\midrule
\rowcolor{lightgray}
\multicolumn{2}{l}{\textit{with Mask Refinement}} & & & & & & & & & & & & & & & & & & & & & \\
GroupViT~\cite{xu2022groupvit} & Custom ViT & \xmark & & 81.5 & 23.8 & 15.4 & 11.6 & 9.4 & 51.1 & 19.0 & 27.9 & 30.0 & & - & - & - & - & - & - & - & - & - \\
ReCo~\cite{shin2022reco} & CLIP & \xmark & & 62.4 & 24.7 & 16.3 & 22.8 & 12.4 & 27.2 & 21.9 & 17.3 & 25.6 & & - & - & - & - & - & - & - & - & - \\
TCL~\cite{cha2023learning} & CLIP & \xmark & & 83.2 & 33.9 & 22.4 & 24.0 & 17.1 & 55.0 & 30.4 & 31.6 & 37.2 & & - & - & - & - & - & - & - & - & - \\
MaskCLIP~\cite{zhou2022extract} & CLIP & \cmark & & 72.1 & 25.3 & 15.1 & 11.2 & 9.0 & 37.2 & 22.6 & 18.9 & 26.4 & & - & - & - & - & - & - & - & - & - \\
SCLIP~\cite{wang2024sclip} & CLIP & \cmark & & 83.5 & 36.1 & 23.9 & 34.1 & 17.8 & 61.7 & 31.5 & 32.1 & 40.1 & & 76.3 & 27.4 & 18.7 & 23.9 & 11.8 & 47.8 & 23.8 & 26.9 & 32.1 \\
CLIP-DINOiser~\cite{wysoczanska2024clip} & CLIP & \cmark & & 81.5 & 37.1 & 25.3 & 31.5 & 20.6 & 64.6 & 33.5 & 36.1 & 41.3 & & - & - & - & - & - & - & - & - & - \\
LaVG~\cite{kang2024defense} & CLIP+DINO & \cmark & & 82.5 & 34.7 & 23.2 & 26.2 & 15.8 & 62.1 & 31.6 & 34.2 & 38.3 & & - & - & - & - & - & - & - & - & - \\
NACLIP~\cite{hajimiri2025pay} & CLIP & \cmark & & 83.0 & 38.4 & 25.7 & 38.3 & 19.1 & 64.1 & 35.0 & 36.2 & 42.5 & & 84.5 & 36.4 & 24.6 & 37.1 & 19.6 & 57.9 & 36.4 & 34.6 & 41.4 \\
FreeDA~\cite{barsellotti2024training} & DINOv2 & \cmark & & 79.5 & 40.2 & 27.1 & 34.4 & 20.9 & 52.0 & 35.2 & 25.8 & 39.4 & & 75.2 & 39.0 & 27.0 & 33.1 & 21.3 & 45.3 & 34.3 & 26.7 & 37.7 \\
FreeDA~\cite{barsellotti2024training} & CLIP+DINOv2 & \cmark & & 85.2 & 42.1 & 27.0 & 33.8 & 21.8 & 51.8 & 37.4 & 38.6 & 42.2 & & 87.1 & 42.4 & 28.1 & 33.8 & 22.6 & 55.4 & 37.1 & 36.1 & 42.8 \\ 
ProxyCLIP~\cite{lan2024proxyclip} & CLIP+DINOv2(reg) & \cmark & & 83.1 & 38.9 & 26.6 & 35.4 & 20.3 & 62.0 & 35.2 & 38.7 & 42.5 & & 85.8 & 37.6 & 25.6 & 37.5 & 22.5 & 59.4 & 34.6 & 39.0 & 42.8 \\ 
ProxyCLIP~\cite{lan2024proxyclip} & CLIP+DINO & \cmark & & 80.3 & 39.4 & 26.9 & \underline{\textbf{38.6}} & 20.2 & 60.8 & 35.3 & 37.2 & 42.3 & & 83.2 & 38.0 & 26.2 & \underline{\textbf{41.0}} & 22.6 & 60.7 & 34.7 & 39.4 & 43.2 \\ 
\rowcolor{LightCyan}
\textbf{\ours (Ours)} & DINOv2(reg) & \cmark & & \underline{\textbf{88.5}} & \underline{\textbf{42.4}} & \underline{\textbf{30.2}} & 38.1 & \underline{\textbf{22.5}} & \underline{\textbf{65.8}} & \underline{\textbf{37.7}} & \underline{\textbf{45.1}} & \underline{\textbf{46.3}} & & \underline{\textbf{89.8}} & \underline{\textbf{42.7}} & \underline{\textbf{29.6}} & 38.4 & \underline{\textbf{22.9}} & \underline{\textbf{66.1}} & \underline{\textbf{37.3}} & \underline{\textbf{42.3}} & \underline{\textbf{46.1}}  \\
\bottomrule
\end{tabular}
}
\vspace{-0.15cm}
\caption{Comparison with unsupervised OVS models on Pascal VOC~\cite{pascal-voc-2012}, Pascal Context~\cite{mottaghi2014role}, COCO Stuff~\cite{caesar2018coco}, COCO Object~\cite{caesar2018coco}, Cityscapes~\cite{cordts2016cityscapes}, and ADE20K~\cite{zhou2017scene,zhou2019semantic}. For each method, we specify the visual backbone used, along with whether it is frozen or fine-tuned. We report both the variants with and without background for Pascal VOC (V21 and V20) and Pascal Context (C60 and C59). Best results with and without mask refinement are highlighted in bold, overall best results are underlined.
}
\label{tab:model_comparison}
\vspace{-0.1cm}
\end{table*}

\section{Experiments}
\label{sec:experiments}

\subsection{Experimental Setup}
\tinytit{Datasets} We evaluate our approach on eight widely-used semantic segmentation benchmarks, which we categorize based on the inclusion of a background class. Specifically, we conduct experiments on the validation sets of Pascal VOC 2012~\cite{pascal-voc-2012}, Pascal Context~\cite{mottaghi2014role}, COCO Stuff~\cite{caesar2018coco}, Cityscapes~\cite{cordts2016cityscapes}, and ADE20K~\cite{zhou2017scene,zhou2019semantic}, that contain 20, 59, 171, 150, and 19 semantic categories respectively and do not include the ``background'' class. We report additional experiments on the COCO Objects dataset~\cite{caesar2018coco}, which consists of 80 different foreground object classes, and on modified versions of Pascal VOC 2012 and Pascal Context in which the ``background'' category is, instead, included (\ie, with 21 and 60 semantic categories respectively).

\tit{Implementation Details} For the main experiments, we employ DINOv2 ViT-B/14 as the base model and DINOv2 ViT-L/14 as the large model, both with the CLIP ViT-B/16 text encoder. We use the DINOv2 variant with registers~\cite{darcet2024vision} since our method benefits from the removal of artifacts in self-attention maps.
We train the model with the Adam optimizer, a batch size of $128$, and a learning rate of $1\times10^{-4}$ for a total of 100 epochs on the COCO Captions 2014 training split~\cite{lin2014microsoft}, composed of around 80k images. 

Following previous works~\cite{cha2023learning,karazija2023diffusion,hajimiri2025pay}, we optionally employ a mask refinement stage to counteract any inaccuracies in the final masks. In particular, we adopt Pixel-Adaptive Mask Refinement (PAMR)~\cite{araslanov2020single}, an iterative post-refinement method aimed at enhancing the fidelity of the similarities to the visual characteristics of the image. In our experiments, we use $\lambda$ equal to $\sfrac{5}{6}$ for background cleaning, a threshold of 0.55 on the similarity score to determine which pixels belong to the ``background'' category, and, when using mask refinement, employ PAMR with 10 iterations.

\tit{Evaluation Protocol} We follow the standard evaluation protocol for unsupervised OVS~\cite{cha2023learning}, where prior access to the target data before evaluation is not
allowed, and use the default class names provided by the \texttt{MMSegmentation} toolbox. The images are resized to have a shorter side of 448, using a sliding window approach with a stride of 224 pixels. All models are evaluated using mean Intersection-over-Union (mIoU) on all the classes of each dataset. 

\subsection{Comparison with the State of the Art}
\label{sec:comparison}

We compare \ours with previous state-of-the-art approaches for unsupervised OVS on the five benchmarks that do not include the ``background'' category and the three benchmarks with the ``background'' category. We consider as competitors: (i) prototype-based approaches, such as ReCo~\cite{shin2022reco} and FreeDA~\cite{barsellotti2024training}, which aim to create visual prototypes associated with the textual categories, (ii) CLIP adaptations, as MaskCLIP~\cite{zhou2022extract}, CLIP-DIY~\cite{wysoczanska2024clipdiy}, SCLIP~\cite{wang2024sclip}, ClearCLIP~\cite{lan2024clearclip}, and NACLIP~\cite{hajimiri2025pay}, which propose architectural modifications to enhance its localization properties, (iii) methods trained on sets of image-caption pairs with objectives designed to force the segmentation capabilities to emerge, like GroupViT~\cite{xu2022groupvit}, TCL~\cite{cha2023learning}, SILC~\cite{naeem2024silc}, and \texttt{dino.txt}~\cite{jose2025dinov2}, and (iv) methods that aim to combine the properties of CLIP and DINO, as CLIP-DINOiser~\cite{wysoczanska2024clip}, LaVG~\cite{kang2024defense}, and ProxyCLIP~\cite{lan2024proxyclip}.

Table~\ref{tab:model_comparison} reports the results on the five benchmarks without background (\ie, Pascal VOC-20, Pascal Context-59, COCO Stuff, Cityscapes, and ADE) and the three benchmarks with background (\ie, Pascal VOC-21, Pascal Context-60, and COCO Object). Specifically, we report the performance of both the base and large configurations of both \ours and the competitors, according to their definitions in the original papers. Moreover, we divide the table into two sections depending on whether a mask refinement technique is employed. Specifically, LaVG exploits a custom region proposer combined with DenseCRF~\cite{krahenbuhl2011efficient}, while all the other methods refine their masks with PAMR. Regarding datasets with background, Pascal VOC and COCO Objects present only foreground categories, also referred to as ``things'' in the literature, while Pascal Context presents both categories from foreground and background, also mentioned as ``stuff''. Hence, following CLIP-DINOiser~\cite{wysoczanska2024clip}, we report the performance with the background cleaning procedure described in Sec.~\ref{sec:cleaning} only on Pascal VOC and COCO Objects.

As it can be observed, our approach achieves the best average mIoU on all the configurations and presents a consistent improvement compared to the considered competitors, with and without the mask refinement, across all datasets except Cityscapes. 
The most straightforward comparison is the one with FreeDA without global similarity (\ie, with DINOv2 only as visual backbone). It builds a bridge between DINOv2 and the CLIP text encoder by retrieving from a collection of visual-textual embedding pairs and by building visual prototypes for each textual category. The significant improvement achieved by \ours demonstrates that training a direct projection from the CLIP text encoder to DINOv2 leads to a more accurate bridge between the two embedding spaces without the overhead in computation and memory provided by the retrieval procedure.

\begin{table}[t]
\centering
\setlength{\tabcolsep}{0.38em}
\resizebox{\linewidth}{!}{
\begin{tabular}{lcc ccccc}
\toprule
& & & \multicolumn{5}{c}{\textbf{mIoU}} \\
\cmidrule{4-8}
\multicolumn{2}{l}{\textbf{Visual Backbone}} & & V20 & C59 & Stuff & City & ADE \\
\midrule
DINO & ViT-S & & 27.7 & 13.2 & 7.4 & 14.8 & 5.2 \\
DINOv2 (without registers) & ViT-S & & 83.7 & \textbf{38.3} & \textbf{25.8} & \textbf{32.9} & \textbf{19.8} \\ 
DINOv2 (with registers)& ViT-S & & \textbf{86.9} & 35.3 & 24.5 & 27.2 & 16.9 \\ 
\midrule
MAE & ViT-B & & 9.5 & 4.3 & 2.0 & 4.0 & 1.1 \\ 
CLIP & ViT-B & & 55.4 & 14.9 & 12.2 & 6.2 & 3.8 \\
DINO & ViT-B & & 27.3 & 11.2 & 7.9 & 13.6 & 4.5 \\ 
DINOv2 (without registers)& ViT-B & & 74.2 & 31.9 & 23.0 & 27.9 & 16.5 \\ 
DINOv2 (with registers) & ViT-B & & \textbf{87.1} & \textbf{39.8} & \textbf{28.1} & \textbf{36.6} & \textbf{21.1} \\ 
\midrule
MAE & ViT-L & & 6.7 & 2.7 & 1.4 & 4.6 & 0.9 \\
CLIP & ViT-L & & 16.6 & 5.0 & 7.7 & 0.9 & 2.0 \\ 
DINOv2 (without registers) & ViT-L & &  56.0 & 20.1 & 14.9 & 18.1 & 8.2 \\
DINOv2 (with registers) & ViT-L & & \textbf{87.1} & \textbf{39.1} & \textbf{27.0} & \textbf{35.8} & \textbf{21.1} \\
\bottomrule
\end{tabular}
}
\vspace{-0.15cm}
\caption{Ablation study results using different visual backbones and with different sizes of the ViT architecture.}
\label{tab:backbone}
\vspace{-0.2cm}
\end{table}

\subsection{Ablation Studies and Analyses}

\tinytit{Choosing Different Visual Backbones}
In Table~\ref{tab:backbone} we show the performance of our approach when varying the visual backbone and the size of the employed ViT architecture. 
We observe that backbones that differ from DINOv2 present unsatisfactory results and can not be aligned to the CLIP textual encoder with a learnable mapping. In particular, while DINO achieves the second best performance on average, \ours heavily benefits from the strong semantic representation of the dense features of DINOv2 and on the capabilities of its self-attention heads in highlighting coherent regions of the image -- properties which are not reflected in other visual backbones. We refer to the supplementary materials for more details on self-attention heads.

Moreover, our results emphasize the critical role of registers~\cite{darcet2024vision} in DINOv2, as demonstrated by the comparison between its variants with and without registers. Registers are a recently proposed mechanism to mitigate the presence of artifacts in the feature maps of ViT-based backbones. Artifacts are tokens that exhibit a significantly higher norm with respect to the other tokens and retain less information about their original position in the image. The alignment process in our method relies on high-quality attention maps, and the presence of artifacts poses a challenge by limiting the selection of the most relevant self-attention heads. Interestingly, since register-related artifacts are more pronounced in larger backbones, the ViT-S variant without registers maintains competitive performance compared to its register-enabled counterpart.
Finally, we observe that our approach maintains robust and consistent performance across different ViT sizes, achieving strong results even with the compact ViT-S backbone. This suggests that our method is effective across a range of model sizes, making it adaptable to varying computational constraints. Additional insights and detailed evaluations of the different visual backbones can be found in the supplementary.

\begin{table}[t]
\centering
\setlength{\tabcolsep}{0.4em}
\resizebox{\linewidth}{!}{
\begin{tabular}{lc ccccc}
\toprule
& & \multicolumn{5}{c}{\textbf{mIoU}} \\
\cmidrule{3-7}
 & & V20 & C59 & Stuff & City & ADE \\
\midrule
\rowcolor{lightgray}
\multicolumn{4}{l}{\textit{Effect of Projection}} & & & \\
Linear Projection (text only) & & 85.1 & 37.9 & 26.7 & 35.6 & 20.1 \\
Our mapping (both vision and text) & & 59.2 & 27.3 & 18.9 & 23.5 & 13.5 \\ 
Our mapping (vision only) & & 84.6 & 35.2 & 26.2 & 20.4 & 15.5 \\
Our mapping (text only) & & \textbf{87.1} & \textbf{39.8} & \textbf{28.1} & \textbf{36.6} & \textbf{21.1} \\ 
\midrule
\rowcolor{lightgray}
\multicolumn{7}{l}{\textit{Effect of Self-Attention Selection and Aggregation}} \\
\texttt{CLS} only (without self-attention)  & & 84.5 & 30.6 & 23.0 & 22.6 & 17.2 \\ 
Standard average & & \textbf{89.6} & 36.9 & 25.6 & 33.5 & 19.7 \\ 
\texttt{CLS}-weighted average & & 87.6 & 35.2 & 23.1 & 29.3 & 17.5 \\ 
\texttt{CLS} similarity-weighted sampling & & 88.2 & 32.9 & 22.6 & 27.0 & 17.9 \\ 
Max \texttt{CLS} similarity & & 87.1 & \textbf{39.8} & \textbf{28.1} & \textbf{36.6} & \textbf{21.1} \\
\bottomrule
\end{tabular}
}
\vspace{-0.15cm}
\caption{Ablation study evaluating the impact of the core components of the proposed architecture on the final performance. We report the results using the base model of DINOv2.}
\label{tab:ablation}
\vspace{-0.2cm}
\end{table}

\begin{figure*}[t]
    \centering
    \normalsize
    \resizebox{\linewidth}{!}{
        \setlength{\tabcolsep}{.06em}
        \begin{tabular}{c p{0mm} c p{0mm} c p{0mm} c p{0mm} c p{0mm} c}
            \addlinespace[2mm]
            \textbf{\huge Image} && \textbf{\huge Ground-truth} && \textbf{\huge FreeDA~\cite{barsellotti2024training}} && \textbf{\huge ProxyCLIP~\cite{lan2024proxyclip}} && \textbf{\huge CLIP-DINOiser~\cite{wysoczanska2024clip}} && \textbf{\huge \ours (Ours)} \\
            \addlinespace[2mm]
            \includegraphics[height=0.3\textwidth]{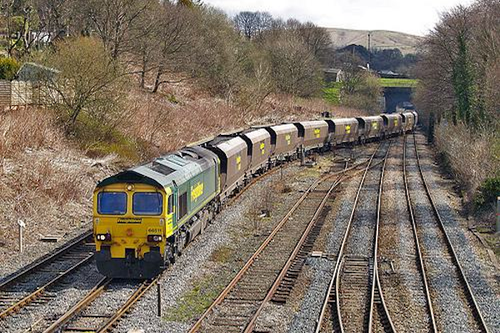} & &
            \includegraphics[height=0.3\textwidth]{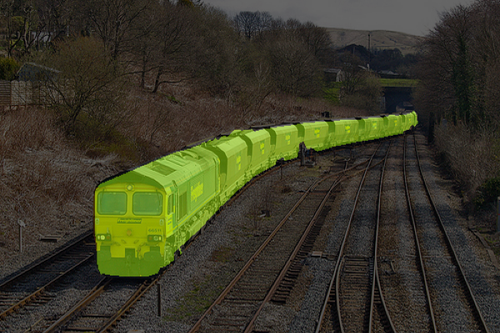} & & \includegraphics[height=0.3\textwidth]{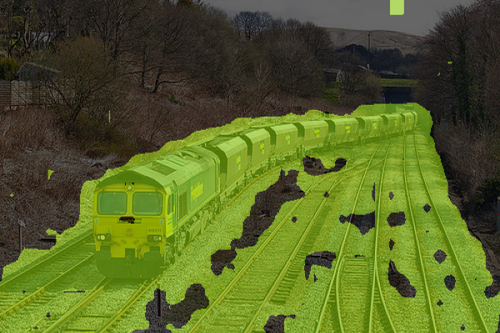} & &
            \includegraphics[height=0.3\textwidth]{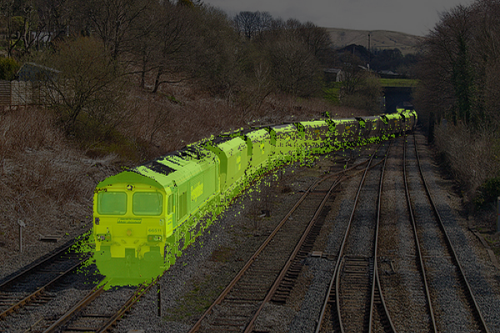} &&
            \includegraphics[height=0.3\textwidth]{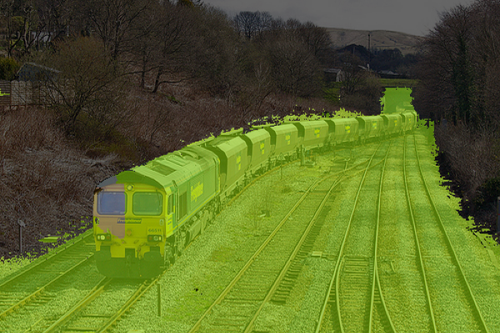} && \includegraphics[height=0.3\textwidth]{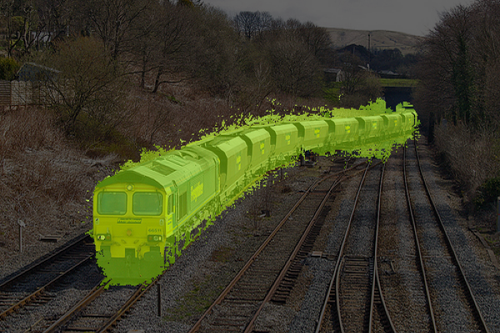} \\
            \includegraphics[height=0.3\textwidth]{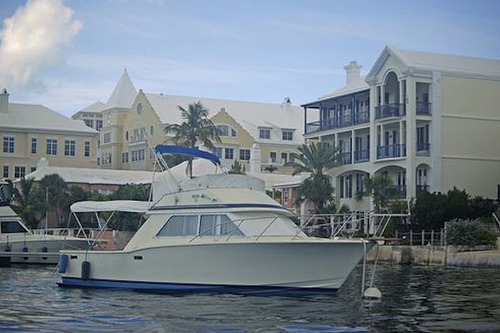} & &
            \includegraphics[height=0.3\textwidth]{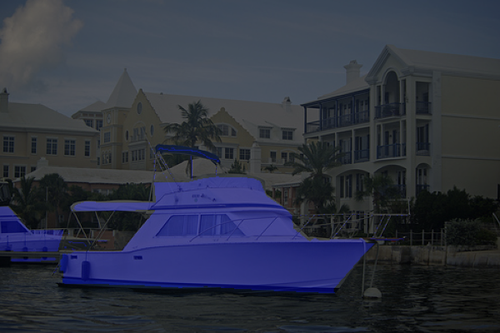} & & \includegraphics[height=0.3\textwidth]{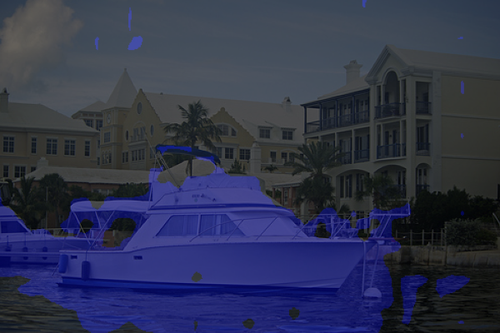} & &
            \includegraphics[height=0.3\textwidth]{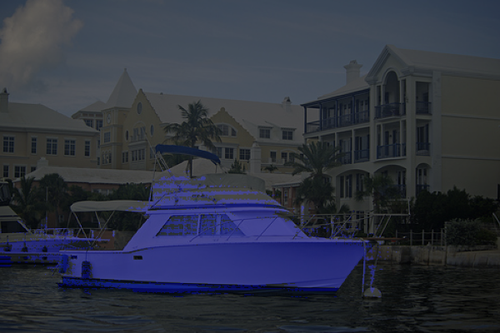} &&
            \includegraphics[height=0.3\textwidth]{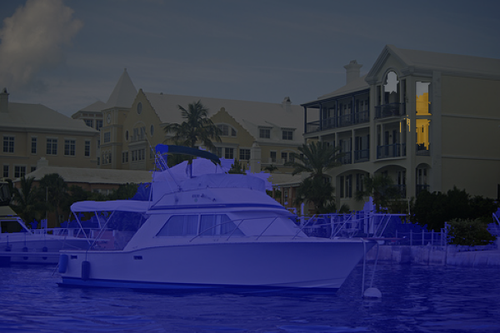} && \includegraphics[height=0.3\textwidth]{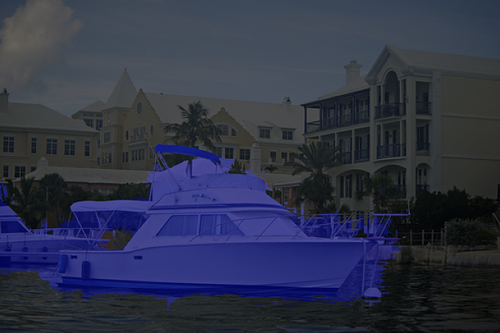} \\
            \includegraphics[height=0.3\textwidth]{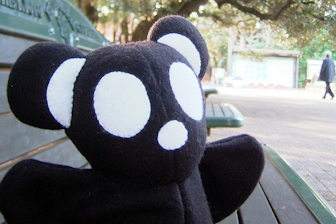} & &
            \includegraphics[height=0.3\textwidth]{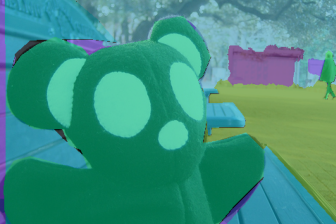} & & \includegraphics[height=0.3\textwidth]{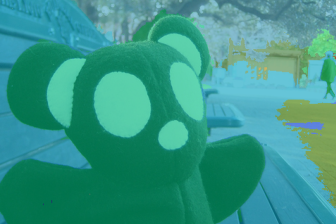} & &
            \includegraphics[height=0.3\textwidth]{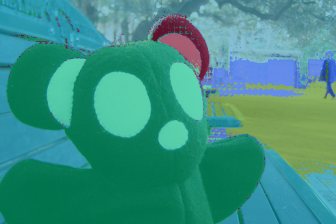} &&
            \includegraphics[height=0.3\textwidth]{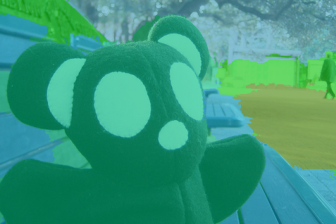} && \includegraphics[height=0.3\textwidth]{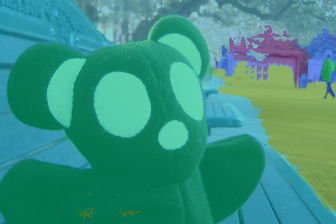} \\
            \includegraphics[height=0.3\textwidth]{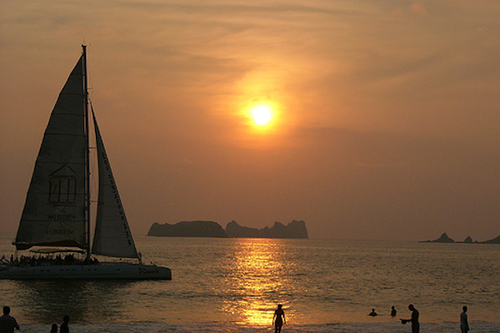} & &
            \includegraphics[height=0.3\textwidth]{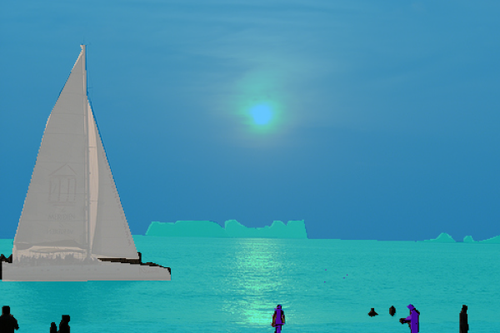} & & \includegraphics[height=0.3\textwidth]{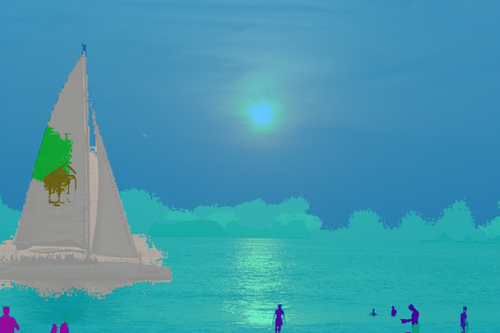} & &
            \includegraphics[height=0.3\textwidth]{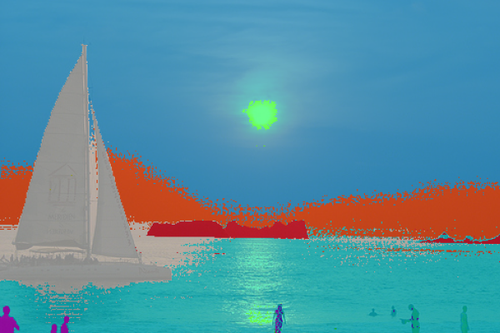} &&
            \includegraphics[height=0.3\textwidth]{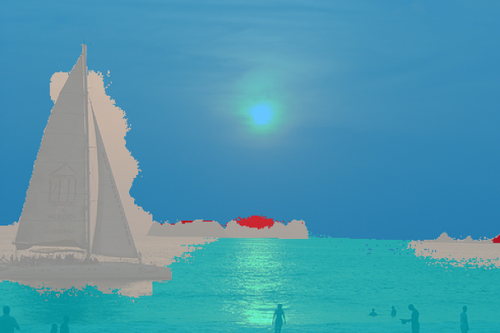} && \includegraphics[height=0.3\textwidth]{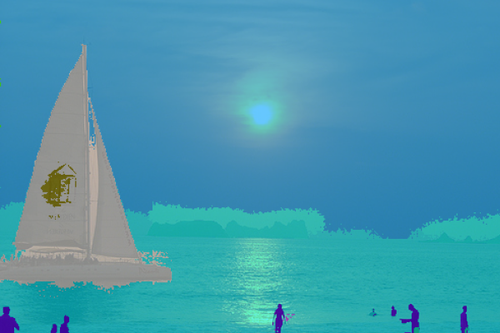} \\

        \end{tabular}
        }
    \vspace{-.15cm}
    \caption{Qualitative results of \ours in comparison with FreeDA~\cite{barsellotti2024training}, ProxyCLIP~\cite{lan2024proxyclip}, and CLIP-DINOiser~\cite{wysoczanska2024clip}.}
    \label{fig:qualitatives}
    \vspace{-.1cm}
\end{figure*}

\tit{Impact of the Proposed Components} Table~\ref{tab:ablation} reports the results of \ours evaluating the impact of its core components on the overall performance. Specifically, in the first section of the table, we analyze the effect of the adopted projection $\psi$. Replacing it with a linear projection leads to a slight performance drop. The good performance obtained by a linear transform evidences how the DINOv2 and CLIP spaces are intrinsically compatible, as the former can be obtained through an affine transform of the latter without losing too much information. Interestingly, applying the proposed projection on top of DINOv2 or using two projections on both spaces significantly lowers performance, confirming the appropriateness of the proposed approach. 

In the second section of the table, we instead study the effect of the selection and aggregation strategy of self-attention heads $A_i, i=1,...,N$ during training. In particular, we test aligning (i) directly the visual \texttt{CLS} token to the textual \texttt{CLS} token, (ii) the visual embedding from the standard average self-attention, (iii) the weighted mean of the head embeddings $v_{A_i}$, where the weights are given by their softmaxed similarity with the textual \texttt{CLS} token, (iv) a strategy in which we sample a single head embedding, where the sampling probability is given by the softmaxed similarity with the \texttt{CLS} token, and (v) our adopted solution in which we select the head embedding which is the most similar to the textual \texttt{CLS} token. The results show that only on the Pascal VOC dataset -- composed mostly by large subjects in foreground -- the embedding from the standard average self-attention presents improved performance. On all other benchmarks, our approach proves to be the most effective, further validating the robustness of our selection method.

\begin{table}[t]
\centering
\setlength{\tabcolsep}{0.42em}
\resizebox{\linewidth}{!}{
\begin{tabular}{lcc cc}
\toprule
&  & & \multicolumn{2}{c}{\textbf{mIoU}} \\
\cmidrule{4-5}
 & \textbf{Mask Refinement} & & V21 & Object \\
\midrule
without background cleaning & \xmark & & 59.9 & 37.1 \\
with background cleaning & \xmark & & \textbf{61.5} & \textbf{41.0} \\
\midrule
without background cleaning & \cmark & & 63.9 & 40.3 \\
with background cleaning & \cmark & & \textbf{65.8} & \textbf{45.1} \\
\bottomrule
\end{tabular}
}
\vspace{-0.15cm}
\caption{Ablation study on the impact of the background cleaning procedure. We report the results using DINOv2 ViT-B.}
\label{tab:ablation_background}
\vspace{-0.1cm}
\end{table}

\tit{Effect of Background Cleaning} Table~\ref{tab:ablation_background} shows how the performance is affected by the background cleaning mechanism and by the usage of PAMR for mask refinement. It can be observed that the background cleaning procedure has a significantly positive impact on Pascal VOC and COCO Object, leading, respectively, to a +1.6 and +3.9 increase in mIoU score. Further, it can be noticed that the effectiveness of the proposed background cleaning procedure is confirmed also when applying the mask refinement. Qualitative results showing the effect of the background cleaning procedure are available in the supplementary material.

\tit{Qualitative Results}
Fig.~\ref{fig:qualitatives} depicts qualitative segmentation results, in which we highlight the segmentation capabilities of \ours along with other state-of-the-art models (\ie, FreeDA~\cite{barsellotti2024training}, ProxyCLIP~\cite{lan2024proxyclip}, and CLIP-DINOiser~\cite{wysoczanska2024clip}). We show two images from Pascal VOC, in which it can also be appreciated how the background cleaning procedure leads to high-quality masks and localization, and two images from COCO Stuff and Pascal Context where \ours effectively segments ``things'' in the scene, such as the \texttt{boat} and \texttt{teddy bear}, and ``stuff'' categories, such as \texttt{sky} and \texttt{road}.

\section{Conclusion}
\label{sec:conclusion}
In this paper, we introduced \ours, a novel approach for OVS that bridges the spatially detailed embeddings of the DINOv2 self-supervised vision backbone with the highly semantic text embeddings of CLIP. Our method achieves fine-grained alignment between textual concepts and visual patches without the need for extensive fine-tuning of the backbone networks, only leveraging self-attention maps from DINOv2 and a lightweight language-to-vision mapping layer. 
\ours achieves state-of-the-art results in standard OVS benchmarks, outperforming previous prototype-based, CLIP adaptation, and contrastive learning approaches.
Our study highlights the potential of integrating vision-only and multimodal models, suggesting broader applications in fine-grained and cross-modal tasks.

\section*{Acknowledgments}
We acknowledge the CINECA award under the ISCRA initiative, for the availability of high-performance computing resources and support. This work has been conducted under a research grant co-funded by Leonardo S.p.A., and supported by the PNRRM4C2 project ``FAIR - Future Artificial Intelligence Research'', by the PRIN 2022-PNRR project ``MUCES'' (CUP E53D23016290001 and B53D23026090001), and by the PNRR project ``ITSERR - Italian Strengthening of Esfri RI Resilience'' (CUP B53C22001770006), all funded by the European Union - NextGenerationEU, and by the SUN XR project funded by the Horizon Europe Research \& Innovation Programme (GA n. 101092612).

{
    \small
    \bibliographystyle{ieeenat_fullname}
    \bibliography{bibliography}
}

\clearpage
\setcounter{page}{1}
\maketitlesupplementary

\appendix
\begin{figure*}[b]
\vspace{-0.2cm}
    \centering
    \includegraphics[width=\linewidth]{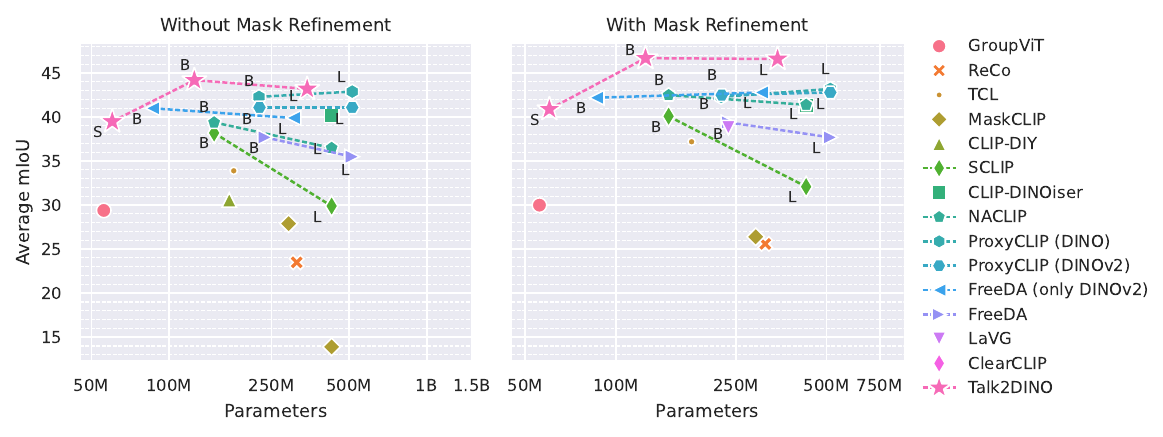}
    \vspace{-0.8cm}
    \caption{\textbf{Performance vs. Parameter Count.} The y-axis denotes the obtained mIoU averaged over all the five benchmarks reported in the main paper. Dashed lines connect methods tested with multiple backbone sizes. These are labeled S/B/L for Small/Base/Large ViT models. \ours offers the best trade-off between performance and number of parameters.}
    \label{fig:perf-vs-params}
\end{figure*}

In the following, we present additional material and experimental analyses of the proposed \ours approach.

\section{Additional Experiments and Analyses}
\tinytit{Analysis of Model Parameters}
Fig.~\ref{fig:perf-vs-params} reports a comparison of the relationship between the performance, in terms of average mIoU, and the number of parameters of the models. As it can be observed, \ours presents a lower number of parameters than the recent competitors FreeDA~\cite{barsellotti2024training} and ProxyCLIP~\cite{lan2024proxyclip}, along with an improved average mIoU. Models with a comparable number of parameters, such as TCL~\cite{cha2023learning}, GroupViT~\cite{xu2022groupvit}, and MaskCLIP~\cite{zhou2022extract}, exhibit a lower performance compared to \ours. Finally, it shall be noted that models such as FreeDA and ReCo~\cite{shin2022reco} require maintaining external sources of knowledge, which increases memory consumption. Further discussion on the comparison between \ours, ProxyCLIP, and FreeDA can be found in the following sections (see "\textit{Comparison with ProxyCLIP and FreeDA}").

\tit{Role of DINO Registers}
The main configuration of \ours, with both the base and large sizes, leverages the variant of DINOv2 with registers. In Fig.~\ref{fig:reg_vs_no_reg_v2} we depict, on the first row, the average self-attentions between the \texttt{CLS} and the other tokens for the ViT-S, ViT-B, and ViT-L architectures with and without registers, while in the following rows, we show the various self-attention heads for each backbone. It can be observed that in the ViT-S the artifacts are not present, and the average self-attention between the model with and without the registers is nearly identical. Instead, the ViT-B exhibits artifacts in the top left corner, resulting in an average self-attention that is especially focused on that portion of the image. This side effect is even more noticeable with the ViT-L, for which the artifact is the only visible token in the average self-attention. These observations align with the results reported in Tab.~\ref{tab:backbone}, that show a downgrade in performance without the registers that is directly related to the presence of the artifacts in the self-attentions. Indeed, the largest difference in performance is measured in the ViT-L architecture, while in the ViT-S case, the backbone without registers performs better on four benchmarks out of five.

\tit{Effect of Training CLIP Last Layer}
Table~\ref{tab:ablation_study_supp} reports a comparison between \ours when training only the $\psi(t)$ projection as proposed in the main paper and when instead unfreezing the last layer of CLIP~\cite{radford2021learning}. Despite this experiment exhibiting a small performance gap between the two configurations, unfreezing the last layer of CLIP, interestingly, leads to worse results. This outcome highlights that the textual representations provided by CLIP, which have been pre-trained to match their visual counterpart, if trained inside a different pipeline, can be harmed and can lose part of their capabilities in multimodal understanding.

\tit{Choosing Different Visual backbones} In Table~\ref{tab:backbone} of the main paper, we report the performance of our approach applied to different visual backbones. The results demonstrated that our training pipeline is especially suitable for DINOv2, while it leads to unsatisfactory performance on DINO, MAE, and CLIP. We attribute this performance gap to two major factors: (i) the quality of the attention maps and (ii) the semantic richness of the patch representations.

For the first point, we qualitatively analyze the self-attention patterns of the different backbones. Fig.~\ref{fig:backbone_attn} showcases the average self-attentions between the \texttt{CLS} token and the other tokens in the first row, breaking down the contributions from the various self-attention heads in the successive rows. We observe that the self-attention heads of CLIP introduce a noise pattern similar to what we observed for DINOv2 without registers, which limits the effectiveness of our training pipeline. On the other hand, the self-attention maps of DINO and MAE appear cleaner and emphasize homogeneous image regions. However, in these cases, the performance gap with DINOv2 can be attributed to the insufficient semantic richness of the extracted dense features.

To quantitatively assess the patch-level semantics of these backbones, we conduct an experiment in which we classify each patch through linear probing on the images of VOC. We determine the ground-truth labels of the patches via majority voting and evaluate accuracy on the validation set (batch size = 16, learning rate = $5 \times 10^{-3}$, for 3 epochs, with 32×32 patches per image, using ViT-B as the backbone). The results, reported in Table~\ref{tab:linear_probing}, align with the overall trends highlighted in the paper: DINOv2 consistently emerges as the best-performing backbone, MAE as the worst, and CLIP and DINO as intermediate. These findings further confirm that the semantic richness of the features extracted by different backbones plays a crucial role in the effectiveness of our approach. Similar conclusions were drawn in the ablation study of FreeDA~\cite{barsellotti2024training}, where a comparable performance drop was observed when using CLIP or DINO instead of DINOv2.

\begin{table}[t]
    \centering
    \resizebox{0.8\linewidth}{!}{
    \renewcommand{\arraystretch}{1.2} 
    \setlength{\tabcolsep}{8pt} 
    \begin{tabular}{lccc}
        \toprule
        \textbf{Visual Backbone} & \textbf{ViT-S} & \textbf{ViT-B} & \textbf{ViT-L} \\ 
        \midrule
        MAE & - & 0.56  & 0.56 \\
        CLIP & - & 0.89  & 0.89 \\
        DINO & 0.62  & 0.73  & - \\
        DINOv2 (without registers) & 0.95  & 0.96  & 0.95  \\
        \rowcolor{Gray} 
        DINOv2 (with registers) & \textbf{0.96}  & \textbf{0.97}  & \textbf{0.96}  \\
        \bottomrule
    \end{tabular}
    }
    \vspace{-0.15cm}
    \caption{Patch linear probing accuracy on VOC for ViT-Small, ViT-Base, and ViT-Large.}
    \label{tab:linear_probing}
    \vspace{-0.35cm}
\end{table}

 \begin{table}[t]
\centering
\setlength{\tabcolsep}{0.25em}
\resizebox{\linewidth}{!}{
\begin{tabular}{lc ccccc}
\toprule
& & \multicolumn{5}{c}{\textbf{mIoU}} \\
\cmidrule{3-7}
 & & V20 & C59 & Stuff & City & ADE \\
\midrule
\rowcolor{lightgray}
\rowcolor{lightgray}
\multicolumn{7}{l}{\textit{Effect of Training CLIP Last Layer}} \\
Trained & & 77.9 & 31.5 & 21.3 & 34.6 & 18.7 \\
Frozen & & \textbf{87.1} & \textbf{39.8} & \textbf{28.1} & \textbf{36.6} & \textbf{21.1} \\ 
\midrule
\rowcolor{lightgray}
\multicolumn{7}{l}{\textit{Effect of Text Token Selection}} \\
Average text tokens & & 84.7 & 37.9 & 25.7 & 33.6 & 20.0 \\ 
Text token to best self-attn map & & 83.9 & 33.8 & 24.2 & 29.5 & 18.1 \\
Text token to best self-attn map (NS) & & 80.8 & 33.9 & 23.7 & 27.5 & 18.6 \\ 
\texttt{CLS} token only & & \textbf{87.1} & \textbf{39.8} & \textbf{28.1} & \textbf{36.6} & \textbf{21.1} \\
\bottomrule
\end{tabular}
}
\vspace{-0.15cm}
\caption{Ablation study on the effect of training the last layer of CLIP and text token selection strategies.}
\label{tab:ablation_study_supp}
\vspace{-0.35cm}
\end{table}

\tit{Using CLIP Text Tokens}
In Table~\ref{tab:ablation_study_supp}, we report the results of utilizing the dense output of the CLIP text encoder instead of its \texttt{CLS} token for alignment. While our primary experiments align the \texttt{CLS} token with the best attention map embedding to target the patches most relevant to the text, we also explore aligning individual text tokens to the best attention map embeddings. This approach is motivated by the hypothesis that each word in the text might correspond to a distinct region in the image. During inference, since we perform the alignment on individual text tokens rather than the \texttt{CLS} token, we average the text tokens to calculate similarity with the visual patches. However, this method yields inferior results compared to using the \texttt{CLS} token. We then refine this approach by aligning only a subset of text tokens selected using nucleus sampling ($\alpha = 0.6$) to filter out potentially irrelevant words, such as stop words. Despite this effort, performance does not improve. 

These observations suggest that the global objective of the training of CLIP, similar to its effect on visual patch embeddings, may not endow text tokens with strong local properties that accurately reflect the specific word each embedding represents. This limitation likely contributes to the noisiness of such alignments. Additionally, we evaluate the use of the average of CLIP text tokens in both training and inference as an alternative to the \texttt{CLS} token. While this approach slightly improves over aligning individual tokens, it still underperforms compared to the \texttt{CLS} token, indicating that it encapsulates the most useful and less noisy information for alignment with DINOv2 patches.

\begin{table*}[!t]
\centering
\setlength{\tabcolsep}{.4em}
\resizebox{\textwidth}{!}{
\begin{tabular}{lc cc c ccccccc cc ccccccc}
\toprule
& & & & & \multicolumn{7}{c}{\textbf{ViT-Base (mIoU)}} & & & \multicolumn{7}{c}{\textbf{ViT-Large (mIoU)}} \\
\cmidrule{6-12} \cmidrule{15-21}
\textbf{Model} & & \textbf{Visual Backbone} & \textbf{Resolution} & & V20 & C59 & Stuff & City & ADE & & \textbf{Avg} & & & V20 & C59 & Stuff & City & ADE & & \textbf{Avg} \\
\midrule
\rowcolor{lightgray}
\multicolumn{2}{l}{\textit{without Mask Refinement}} & & & & & & & & & & & & & & & & & & &  \\
SCLIP~\cite{wang2024sclip} & & CLIP & 336 & & 80.4 & 34.2 & 22.4 & 32.2 & 16.1 & & 37.1 & & & 70.6 & 25.2 & 17.6 & 21.3 & 10.9 & & 29.1 \\ 
NACLIP~\cite{hajimiri2025pay} & & CLIP & 336 & & 79.7 & 35.2 & 23.3 & 35.5 & 17.4 & & 38.2 & & & 78.7 & 32.1 & 21.4 & 31.4 & 17.3 & & 36.2 \\ 
ProxyCLIP~\cite{lan2024proxyclip} & & CLIP+DINOv2 & 336 & & 80.5 & 37.3 & 25.3 & 35.8 & 19.0 & & 39.6 & & & 83.5 & 36.7 & 25.0 & 35.8 & 21.0 & & 40.4 \\ 
ProxyCLIP~\cite{lan2024proxyclip} & & CLIP+DINO & 336 & & 78.2 & 38.8 & 26.2 & \textbf{39.7} & 19.7 & & 40.5 & & & 82.1 & \textbf{38.2} & \textbf{26.2} & \textbf{41.2} & \textbf{22.2} & & \textbf{42.0} \\ 
\rowcolor{LightCyan}
\textbf{\ours (Ours)} & & DINOv2 & 336 & & \textbf{88.3} & \textbf{39.1} & \textbf{27.4} & 38.2 & \textbf{20.2} & & \textbf{42.6} & & & \textbf{86.6} & \textbf{38.2} & 26.0 & 36.4 & 19.3 & & 41.3 \\
\midrule
\rowcolor{lightgray}
\multicolumn{2}{l}{\textit{with Mask Refinement}} & & & & & & & & & & & & & & & & & & & \\
SCLIP~\cite{wang2024sclip} & & CLIP & 336 & & 79.3 & 34.6 & 22.3 & 20.3 & 15.4 & & 34.4 & & & 66.6 & 22.4 & 14.7 & 6.9 & 7.7 & & 23.7 \\
NACLIP~\cite{hajimiri2025pay} & & CLIP & 336 & & 83.0 & 38.4 & 25.7 & 38.3 & 19.1 & & 40.9 & & & 84.5 & 36.4 & 24.6 & 37.1 & 19.6 & & 40.4 \\ 
ProxyCLIP~\cite{lan2024proxyclip} & & CLIP+DINOv2 & 336 & & 80.9 & 39.3 & 26.6 & 37.7 & 19.9 & & 40.9 & & & 83.5 & 36.7 & 26.4 & 38.6 & 22.1 & & 41.5 \\ 
ProxyCLIP~\cite{lan2024proxyclip} & & CLIP+DINO & 336 & & 78.5 & 39.3 & 26.7 & 40.1 & 20.0 & & 40.9 & & & 82.6 & 38.7 & 26.7 & \textbf{42.1} & \textbf{22.5} & & 42.5 \\
\rowcolor{LightCyan}
\textbf{\ours (Ours)} & & DINOv2 & 336 & & \textbf{89.4} & \textbf{41.5} & \textbf{29.4} & \textbf{40.3} & \textbf{21.2} & & \textbf{44.4} & & & \textbf{89.5} & \textbf{41.7} & \textbf{29.8} & 38.7 & 20.8 & & \textbf{44.1} \\
\midrule
\rowcolor{lightgray}
\multicolumn{2}{l}{\textit{without Mask Refinement}} & & & & & & & & & & & & & & & & & & &  \\
SCLIP~\cite{wang2024sclip} & & CLIP & 448 & &77.8	& 33.0 & 	21.1 & 	19.8 & 	14.6 & &	33.3 & & &61.2 &	20.5 & 13.1 &	6.7	& 7.0 & &	21.7 \\ 
NACLIP~\cite{hajimiri2025pay} & & CLIP & 448 & &71.3 &	34.8 &	22.9&	33.7&	17.7& &	36.1 & & & 74.5	&32.6&	21.6&	30.5&	17.8&&	35.4 \\ 
ProxyCLIP~\cite{lan2024proxyclip} & & CLIP+DINOv2 & 448 & & 83.3&	37.8&	25.6&	28.8&	19.1&&	38.9 & & &85.0	&36.6&	25.0&	33.8&	20.6& &	40.2 \\ 
ProxyCLIP~\cite{lan2024proxyclip} & & CLIP+DINO & 448 & & 80.4 &	39.0&	26.2&	31.7&	19.5&&	39.4  & &  &83.1 &	37.8 &	25.9&	\textbf{37.5}&	\textbf{21.6}& &	41.2 \\
\rowcolor{LightCyan}
\textbf{\ours (Ours)} & &  DINOv2 &  448 & &  \textbf{87.1} &  \textbf{39.8} &  \textbf{28.1} &  \textbf{36.6} &  \textbf{21.1} & &  \textbf{42.5} & & &  \textbf{87.1} &  \textbf{39.1} &  \textbf{27.0} &  35.8 &  21.1 & &  \textbf{42.0} \\
\midrule
\rowcolor{lightgray}
\multicolumn{2}{l}{\textit{with Mask Refinement}} & & & & & & & & & & & & & & & & & & & \\
SCLIP~\cite{wang2024sclip} & & CLIP & 448 & & 79.3 &	34.6 &	22.3 &	20.3 &	15.4 & &	34.4 & & & 66.6 &	22.4 &	14.7 & 	6.9 &	7.7 & &	23.7 \\ 
NACLIP~\cite{hajimiri2025pay} & & CLIP & 448 & &74.9&	37.6&	25.2&	36.1&	18.4& &	38.4 & & & 79.8 &	36.8 &	25.0&	35.6&	18.4 &&	39.1 \\ 
ProxyCLIP~\cite{lan2024proxyclip} & & CLIP+DINOv2 & 448 & & 83.1&	39.3&	26.7&	29.5&	19.7& &	39.7 & & & 85.1&	37.8&	25.9&	35.3&	21.4 &&	41.1 \\ 
ProxyCLIP~\cite{lan2024proxyclip} & & CLIP+DINO & 448 & & 80.0	& 39.1&	26.5&	31.7&	19.5 & &	39.4 & & & 82.8&	37.8&	26.2&	26.2&	21.6&&	38.9 \\ 
\rowcolor{LightCyan}
 \textbf{\ours (Ours)} & &  DINOv2 &  448 & &  \textbf{88.5} &  \textbf{42.4} &  \textbf{30.2} &  \textbf{38.1} &  \textbf{22.5} & &  \textbf{44.3} & & &  \textbf{89.8} &  \textbf{42.7} &  \textbf{29.6} &  \textbf{38.4} &  \textbf{22.9} & &  \textbf{44.7} \\

\bottomrule
\end{tabular}
}
\vspace{-0.15cm}
\caption{Comparison with unsupervised OVS models on Pascal VOC~\cite{pascal-voc-2012}, Pascal Context~\cite{mottaghi2014role}, COCO Stuff~\cite{caesar2018coco}, Cityscapes~\cite{cordts2016cityscapes}, and ADE20K~\cite{zhou2017scene,zhou2019semantic} following the evaluation setting proposed in SCLIP~\cite{wang2024sclip} (\textit{resolution 336}) and TCL~\cite{cha2023learning} (\textit{resolution 448}).
}
\label{tab:model_comparison_336}
\vspace{-0.2cm}
\end{table*}

\tit{Impact of Image Resolution}
According to the evaluation protocol introduced in GroupViT~\cite{xu2022groupvit} and standardized in TCL~\cite{cha2023learning}, the images are resized to have a shorter side of 448, and a sliding window approach with a stride of 224 pixels is employed. However, Wang \etal~\cite{wang2024sclip} observed that the approaches based on CLIP benefit from employing a shorter side of 336 with 224 $\times$ 224 windows and a stride of 112 pixels, leading to an equivalent computational effort but better performance. This phenomenon is attributed to two reasons: (i) each window has the same resolution on which CLIP has been originally trained, and (ii) CLIP presents an impressive global understanding but lacks localization capabilities, hence relying on many smaller windows is more advantageous than more patches. This variation of the evaluation setting is not necessary for \ours because DINOv2, which is the frozen underlying visual encoder, has been trained with a 518 $\times$ 518 resolution and presents an outstanding patch-level understanding, However, Tab.~\ref{tab:model_comparison_336} reports the results obtained following the setting used in SCLIP, employing a shorter side of 336 for VOC, Context, COCO-Stuff and ADE, of 560 for Cityscapes, with 224 $\times$ 224 windows and stride 112. Results show that \ours, on average, performs better with a resolution of 448, but the performance slightly varies when changing the setting to 336. This confirms that the semantics of the patch-level features of DINOv2 are robust towards variations of resolution and that our learned bridge is valid for both scenarios. Moreover, for a fair comparison, we also report the results of SCLIP, NACLIP, and ProxyCLIP when adopting the 448 resolution of the standard protocol, in which \ours largely outperforms the competitors.

\tit{Comparison with ProxyCLIP and FreeDA}
GroupViT~\cite{xu2022groupvit} has been the first model to tackle the weakly-supervised OVS. It trains a custom ViT architecture from scratch by hierarchically merging tokens at different layers. Afterward, several works followed this direction, investigating how to let the segmentation capabilities to emerge by training over a large corpora of image-caption pairs. On the contrary, more recent works focused on finding modifications to the architecture of CLIP in order to improve its localization properties. Moreover, some methods consider the usage of further visual encoders with enhanced localization capabilities to help CLIP on dense tasks. Among these methods, ProxyCLIP and FreeDA study how to combine DINO and DINOv2 with CLIP.

FreeDA employs Stable Diffusion to create a huge collection of \textit{localized} images from captions, detecting the area in which each noun of the caption has been generated. This information is used to build a database of textual-visual embedding pairs, in which the textual embedding is obtained with CLIP on each noun and the visual embedding is the average patch-level embedding of DINOv2 from the corresponding area. Then, at inference time, a set of textual embeddings is retrieved for each input category, and the corresponding visual embeddings are averaged to create a prototype for that category in the space of DINOv2. Finally, the CLIP visual encoder runs on the input image to solve ambiguities and remove noise. ProxyCLIP proposes to leverage the semantic coherence of a visual encoder such as DINO or DINOv2 to guide the computation of the patch-level embeddings of CLIP. This guidance is performed inside an attention module, in which the patch-level embeddings of DINO act as queries and keys while those of CLIP act as values.

\begin{table*}[t]
\centering
\newcommand{\g}[1]{ \textcolor{green}{(#1)}}
\renewcommand{\r}[1]{ \textcolor{red}{(#1)}}
\renewcommand{\b}[1]{\textbf{#1}}
\renewcommand{\u}{$\uparrow$}
\renewcommand{\d}{$\downarrow$}
\newcommand{\m}[2]{\multicolumn{#1}{c}{#2}}
\setlength{\tabcolsep}{0.38em}
\resizebox{0.82\linewidth}{!}{
\begin{tabular}{ll ccccc c ccccc}
\toprule
              & \m{5}{\textbf{Image$\to$Text}}              & \m{5}{\textbf{Text$\to$Image}} \\
              \cmidrule(lr){2-6}            \cmidrule(lr){7-11}
              & R@1 \u       & R@5 \u       & R@10 \u & Median \d & Mean \d       & R@1 \u       & R@5 \u       & R@10 \u & Median \d & Mean \d      \\
\midrule
\rowcolor{lightgray}
\multicolumn{2}{l}{\textit{ViT-Base}} &           &           &           &           &            &           &           &           &           \\
CLIP           & \textbf{41.3} & \textbf{65.8} & \textbf{76.3} & \textbf{2} & 13.4          & 22.6 & 44.1 & 54.9 & 8 & 52.5          \\
\rowcolor{LightCyan}
\b{\ours} & 29.5 & 56.0 & 69.0 & 4 & 16.4         & 12.5 & 34.0 & 48.4 & 11 & 38.4          \\ 
\rowcolor{LightCyan}
\quad + Custom Alignment & 28.6 & 58.8 & 72.0 & 4 & \textbf{12.0} & \textbf{28.0} & \textbf{55.6} & \textbf{68.7} & \textbf{4} & \textbf{19.3} \\ 
\midrule
\rowcolor{lightgray}
\multicolumn{2}{l}{\textit{ViT-Large}} &           &           &           &           &                  &           &           &           &     \\
CLIP & \textbf{45.4} & \textbf{71.1} & \textbf{79.2} & \textbf{2} & \textbf{11.0}          & \textbf{26.5} & 48.7 & 59.0 & 6 & 44.2      \\
\rowcolor{LightCyan}
\b{\ours} & 26.5 & 53.7 & 65.6 & 5 & 18.8 & 12.7 & 33.7 & 47.8 & 11 & 43.1        \\
\rowcolor{LightCyan}
\quad + Custom Alignment & 37.9 & 64.7 & 75.1 & 3 & 13.1          & 24.4 & \textbf{50.1} & \textbf{63.2} & \textbf{5} & \textbf{27.8}       \\
\bottomrule
\end{tabular}
}
\vspace{-0.15cm}
\caption{Retrieval performance on the COCO Captions test set.}
\label{tab:retrieval_results}
\vspace{-0.2cm}
\end{table*}

\begin{table*}[t]
    \centering
    \resizebox{0.65\linewidth}{!}{
    \renewcommand{\arraystretch}{1.2} 
    \setlength{\tabcolsep}{7pt}
    \begin{tabular}{lcccc}
        \toprule
         & \textbf{Visual Encoder} & \textbf{Params (M)} & \textbf{FLOPS (G)} & \textbf{Ext. (GiB)} \\ 
        \midrule
        ProxyCLIP & CLIP ViT-B/16 + DINO ViT-B/8 & 172.0  & 521.2 & -\\
        ProxyCLIP & CLIP ViT-B/16 + DINOv2 ViT-B/14 & 172.8  & 180.8 & -\\
        FreeDA & CLIP ViT-B/16 + DINOv2 ViT-B/14  & 172.8  & 125.1 & 12.5 \\
        \rowcolor{LightCyan}
        \textbf{\ours} & DINOv2 ViT-B/14  & 86.6  & 107.4 & -\\
        \bottomrule
    \end{tabular}
    }
    \vspace{-0.15cm}
    \caption{Number of parameters, FLOPS, and the dimension of the external knowledge for ProxyCLIP, FreeDA, and \ours.}
    \label{tab:flops}
    \vspace{-0.3cm}
\end{table*}

\begin{table*}[t]
\centering
\resizebox{0.8\linewidth}{!}{
\begin{tabular}{
    lc c
    >{\columncolor{white}}c
    >{\columncolor{lightgray}}c
    >{\columncolor{white}}c
    >{\columncolor{lightgray}}c
    >{\columncolor{white}}c
    >{\columncolor{lightgray}}c
    >{\columncolor{white}}c
    >{\columncolor{lightgray}}c
    >{\columncolor{white}}c
}
\toprule
\textbf{Model} & \textbf{Visual Encoder} & & 
\cellcolor{white}V20 & 
\cellcolor{lightgray}C59 & 
\cellcolor{white}Stuff & 
\cellcolor{lightgray}City & 
\cellcolor{white}ADE & 
\cellcolor{lightgray}V21 & 
\cellcolor{white}C60 & 
\cellcolor{lightgray}Object & 
\cellcolor{white}\textbf{Avg}\\
\midrule
\rowcolor{lightgray}
\multicolumn{12}{l}{\textit{DINOv2 ViT-B/14 with registers (without Mask Refinement)}} \\
FreeDA & DINOv2 & & 83.4 & 39.5 & 25.9 & 35.2 & 20.7 & 50.1~$\triangleright$~43.6 & 34.3 & 23.8~$\triangleright$~24.7 & 39.1~$\triangleright$~38.4 \\ 
FreeDA & CLIP+DINOv2 & & 87.0 & \textbf{40.6} & 25.7 & 34.2 & \textbf{21.2} & 49.3~$\triangleright$~41.8 & \textbf{35.7} & 34.8~$\triangleright$~34.7 & 41.1~$\triangleright$~40.1 \\ 
ProxyCLIP & CLIP+DINOv2 & & 83.0 & 37.2 & 25.4 & 33.9 & 19.7 & 58.6~$\triangleright$~60.0 & 33.8 & 37.4~$\triangleright$~37.3 & 41.1~$\triangleright$~41.3 \\ 
\rowcolor{LightCyan}
\textbf{\ours} & DINOv2 & & \textbf{87.1} & 39.8 & \textbf{28.1} & \textbf{39.6} & 21.1 & 59.9~$\triangleright$~\textbf{61.5} & 35.1 & 37.1~$\triangleright$~\textbf{41.0} & 43.5~$\triangleright$~\textbf{44.2} \\
\midrule
\rowcolor{lightgray}
\multicolumn{12}{l}{\textit{DINOv2 ViT-B/14 with registers (with Mask Refinement)}} \\
FreeDA & DINOv2 & & 84.9 & 42.3 & 27.7 & 36.8 & 22.0 & 50.2~$\triangleright$~43.7 & 36.7 & 24.5~$\triangleright$~25.5 & 40.6~$\triangleright$~40.0\\ 
FreeDA & CLIP+DINOv2 & & 87.4 & \underline{\textbf{42.4}} & 26.6 & 34.8 & 22.1 & 49.4~$\triangleright$~41.7 & 37.2 & 36.6~$\triangleright$~36.7 & 42.1~$\triangleright$~41.1 \\ 
ProxyCLIP & CLIP+DINOv2 & & 83.1 & 38.9 & 26.6 & 35.4 & 20.3 & 62.0~$\triangleright$~63.4 & 35.2 & 38.7~$\triangleright$~38.6 & 42.5~$\triangleright$~42.7 \\ 
\rowcolor{LightCyan}
\textbf{\ours} & DINOv2 & & \underline{\textbf{88.5}} & \underline{\textbf{42.4}} & \underline{\textbf{30.2}} & \underline{\textbf{41.6}} & \underline{\textbf{22.5}} & 63.9~$\triangleright$~\underline{\textbf{65.8}} & \underline{\textbf{37.7}} & 40.3~$\triangleright$~\underline{\textbf{45.1}} & 45.9~$\triangleright$~\underline{\textbf{46.7}}\\
\bottomrule
\end{tabular}
}
\vspace{-0.15cm}
\caption{Comparison between FreeDA, ProxyCLIP, and \ours when using DINOv2 with and without registers. For VOC21, Object, and the average, we report the results without background cleaning on the left and with background cleaning on the right.}
\label{tab:model_comparison_reg}
\vspace{-0.3cm}
\end{table*}

\ours, similarly to FreeDA and ProxyCLIP, investigates how to leverage DINOv2 to compensate for CLIP. However, we propose to employ contrastive learning over a large set of image-caption pairs based on maximum similarity between the attention head embeddings and texts, to learn a functional mapping that bridges the CLIP text embeddings into the DINOv2 space. Our approach demonstrates that the two spaces can be directly connected to set the new state-of-the-art in the unsupervised OVS field.
Table~\ref{tab:flops} shows a quantitative comparison in terms of the number of parameters and FLOPS of the visual encoders and the dimension of the external knowledge (\ie, the database of FreeDA), when assuming an input image with a resolution of 448 $\times$ 448. The results highlight that our method is more practical and less
demanding in computation and memory, while presenting improved results against all competitors.

In Tab.~\ref{tab:model_comparison}, we followed the original configurations of the competitors and, hence, ProxyCLIP uses DINOv2 with registers while FreeDA does not. We report a comparison with and without registers in Tab.~\ref{tab:model_comparison_reg}. The registers present the greatest impact on \ours, because, as described in "\textit{Role of DINO Registers}", the presence of anomaly tokens leads all the self-attention heads to focus only on them, preventing the selection of diverse areas during training and, hence, limiting the efficacy of our proposal. Moreover, in Tab.~\ref{tab:model_comparison_reg} we report the effect of the background cleaning also on FreeDA and ProxyCLIP. This approach is effective only on \ours due to the learned alignment between text and average embeddings of the self-attention heads, while it leads to lower results when applied to the other methods.

\tit{ViT-B vs ViT-L}
Tab.~\ref{tab:model_comparison} of the main paper shows that, without mask refinement, the results achieved by \ours with DINOv2 ViT-B as vision encoder are slightly better than the ones achieved with ViT-L, while the opposite should be expected. However, when we apply the PAMR for mask refinement, the results of ViT-L significantly improve, surpassing the ViT-B on five benchmarks out of eight. A similar phenomenon can be observed in other competitors, such as MaskCLIP, SCLIP, ClearCLIP, and NACLIP, while in FreeDA and ProxyCLIP we cannot establish an encoder size that prevails on the other. Even from the experiment in Tab.~\ref{tab:linear_probing} on patch-level linear probing, we can observe that ViT-B performs slightly better than ViT-L. These results suggest that DINOv2 ViT-L has a comparable semantic understanding with respect to ViT-B, but presents inferior localization properties, which are compensated through PAMR. We hypothesize that training the model with a form of weak- or self-supervision by exploiting the innate capabilities of pre-trained backbones lacks a direct relation between performance and model size. Indeed, the impressive semantic and localized understanding of DINOv2 is a consequence of its training procedure but not the direct objective. From Figure~\ref{fig:reg_vs_no_reg_v2}, it is noteworthy that the activations of ViT-S, ViT-B, and ViT-L have very different behaviors, impacting the results of \ours.

\section{Image-Text Matching Results}
While \ours is primarily designed for OVS, we also assess its performance on image-text retrieval to evaluate its capabilities in global image understanding. For this task, we adopt the same text encoding approach used in segmentation, projecting the CLIP text embedding. The global image representation is derived by averaging the embeddings computed from each DINOv2 attention map. Specifically, for each attention map $A_i$, we calculate a visual embedding $v^{A_i} \in \mathbb{R}^{D_v}$ as the weighted average of the dense feature map $v$. The final global image representation is then obtained by taking the mean of all $v^{A_i}$ embeddings.

In Table~\ref{tab:retrieval_results}, we assess the retrieval performance on the COCO Captions test set~\cite{lin2014microsoft} using both ViT-B and ViT-L configurations. While \ours generally performs slightly below CLIP across most metrics, it demonstrates a notable advantage in the mean rank for the text-to-image retrieval task. This result underscores the ability of \ours to better address extreme failures compared to CLIP, indicating improved robustness in handling challenging or outlier queries.  In addition to computing text-image similarities using cosine similarity between a global text token and a global image token, we experiment with a similarity function that mirrors the one used during training. Specifically, instead of representing the image with the mean of the $v^{A_i}$ embeddings and calculating similarity as the cosine similarity between this representation and the text encoding, we represent the image using all $v^{A_i}$ embeddings. We compute the similarity as $\max_{i=1, \dots, N} \text{sim}(v^{A_i}, t)$, taking the maximum similarity score across all heads. This alternative similarity function leads to significant performance improvements, allowing \ours to surpass CLIP on several metrics. This enhancement is likely due to the ability of the model to evaluate captions at a finer granularity. Captions often describe multiple aspects of an image, including both foreground and background elements. By individually examining different regions of the image as detected by distinct attention heads, the model can assign more precise scores, ultimately boosting retrieval accuracy.

\section{Activation Map Visualizations}
\begin{figure}[t]
    \centering
    \includegraphics[width=\linewidth]{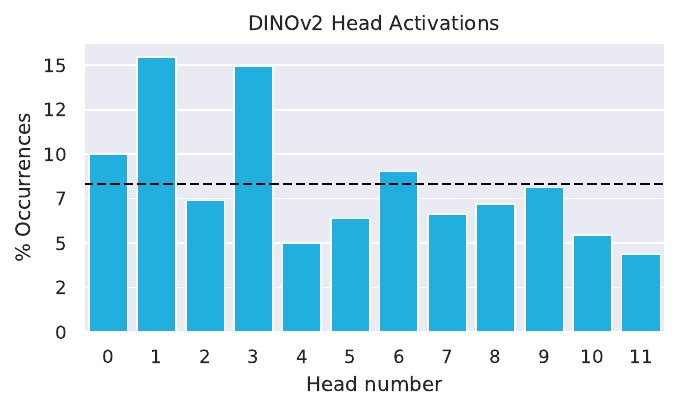}
    \vspace{-0.8cm}
    \caption{Percentage of times each attention head of ViT-B backbone is selected for alignment to textual embeddings on the final epoch of training. The dashed line denotes uniform distribution.}
    \label{fig:activations_head_occurences}
    \vspace{-0.3cm}
\end{figure}

In Fig.~\ref{fig:activations_head_occurences}, we show the distribution of attention heads selected for alignment with the text input during the final epoch of training. 
The results indicate that certain heads, particularly heads 1 and 3, are more often aligned with the text than others. However, aside from these, the remaining heads are relatively evenly distributed.
These findings are noteworthy because they suggest that some heads specialize in capturing features that align more closely with the input caption, while all heads contribute meaningfully during training. Notably, no head shows a negligible activation frequency, highlighting the importance of the entire set of attention heads in the alignment process.

\begin{figure*}[t]
        \newcommand{\colwid}{0.15\linewidth}
        \centering
        \sffamily

        \begin{subfigure}[t]{\colwid}
            \includegraphics[width=\linewidth]{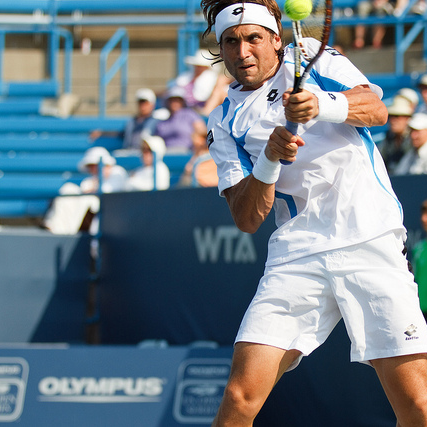}
            \tiny\raggedright
        \end{subfigure}
        \hfill
        \begin{subfigure}[t]{\colwid}
            \includegraphics[width=\linewidth]{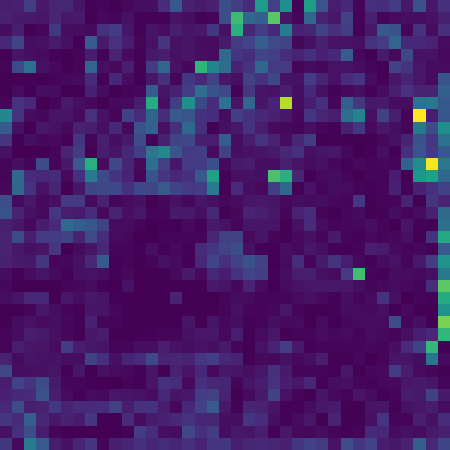}
            \tiny\raggedright
            \textbf{Head 1}: A professional tennis player hits a ball as fans watch. \\
        \end{subfigure}
        \hfill
        \begin{subfigure}[t]{\colwid}
            \includegraphics[width=\linewidth]{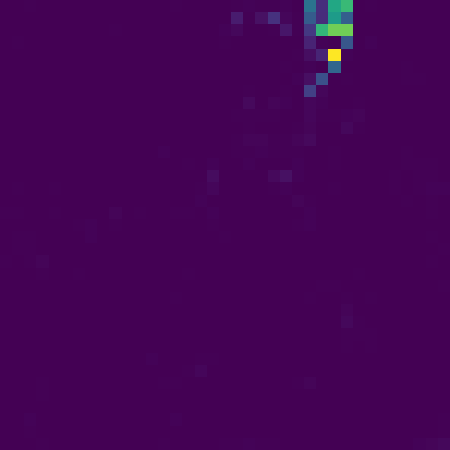}
            \tiny\raggedright
\textbf{Head 5}: champion tennis player swats at the ball hoping to win \\
        \end{subfigure}
        \hfill
        \begin{subfigure}[t]{\colwid}
            \includegraphics[width=\linewidth]{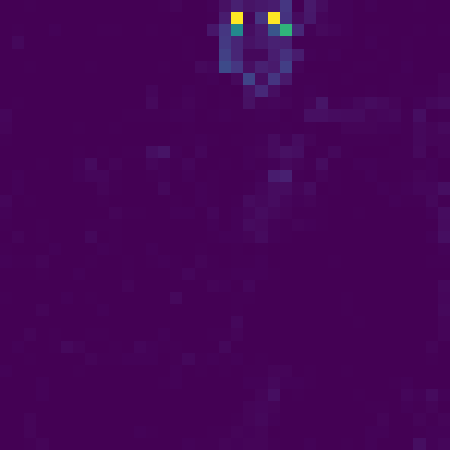}
            \tiny\raggedright
\textbf{Head 7}: A man hitting a tennis ball with a racquet. \\
        \end{subfigure}
        \hfill
        \begin{subfigure}[t]{\colwid}
            \includegraphics[width=\linewidth]{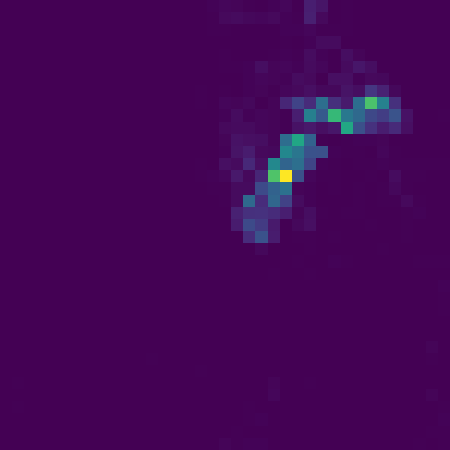}
            \tiny\raggedright
\textbf{Head 10}: A man is hitting his tennis ball with a recket on the court. \\
        \end{subfigure}
        \hfill
        \begin{subfigure}[t]{\colwid}
            \includegraphics[width=\linewidth]{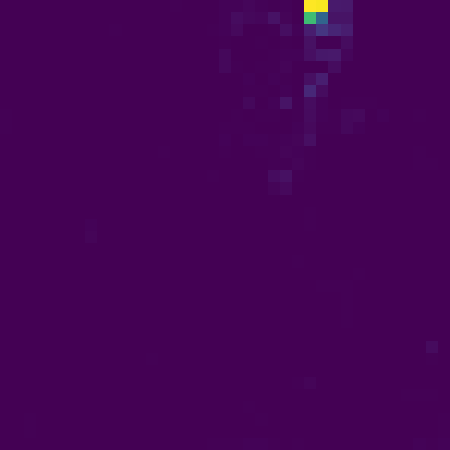}
            \tiny\raggedright
\textbf{Head 11}: a tennis player on a court with a racket \\
        \end{subfigure}
        \begin{subfigure}[t]{\colwid}
            \includegraphics[width=\linewidth]{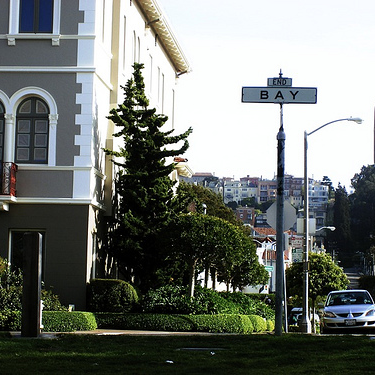}
            \tiny\raggedright
        \end{subfigure}
        \hfill
        \begin{subfigure}[t]{\colwid}
            \includegraphics[width=\linewidth]{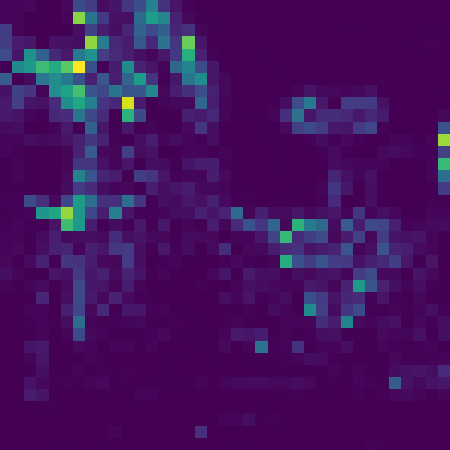}
            \tiny\raggedright
            \textbf{Head 0}: A gray and white building on the corner of bay street with building behind it. \\
        \end{subfigure}
        \hfill
        \begin{subfigure}[t]{\colwid}
            \includegraphics[width=\linewidth]{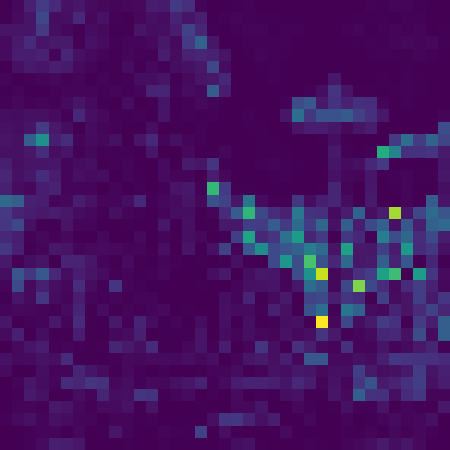}
            \tiny\raggedright
\textbf{Head 1}: A street in the ciy San franscisco on a good day \\
        \end{subfigure}
        \hfill
        \begin{subfigure}[t]{\colwid}
            \includegraphics[width=\linewidth]{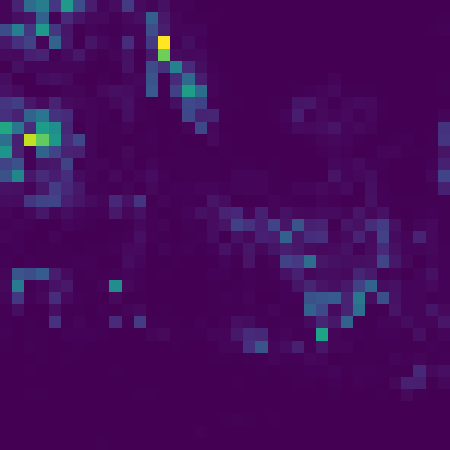}
            \tiny\raggedright
\textbf{Head 2}: A tree leaning on a building on Bay Street. \\
        \end{subfigure}
        \hfill
        \begin{subfigure}[t]{\colwid}
            \includegraphics[width=\linewidth]{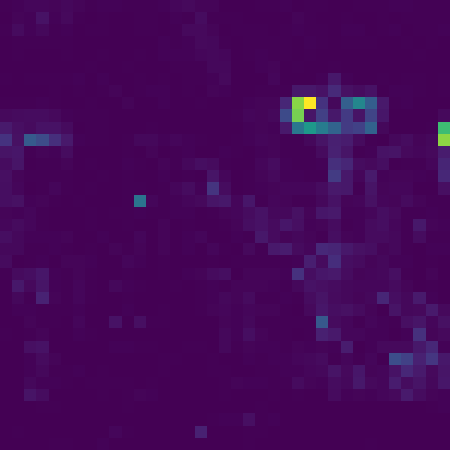}
            \tiny\raggedright
\textbf{Head 5}: A car driving down a street next to a tall building. \\
        \end{subfigure}
        \hfill
        \begin{subfigure}[t]{\colwid}
            \includegraphics[width=\linewidth]{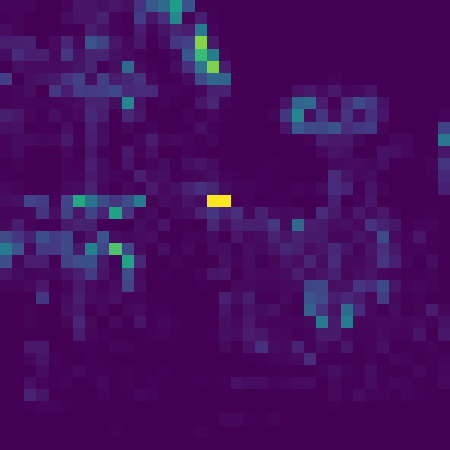}
            \tiny\raggedright
\textbf{Head 9}: A street sign for Bay Street in a residential neighborhood. \\
        \end{subfigure}

        \begin{subfigure}[t]{\colwid}
            \includegraphics[width=\linewidth]{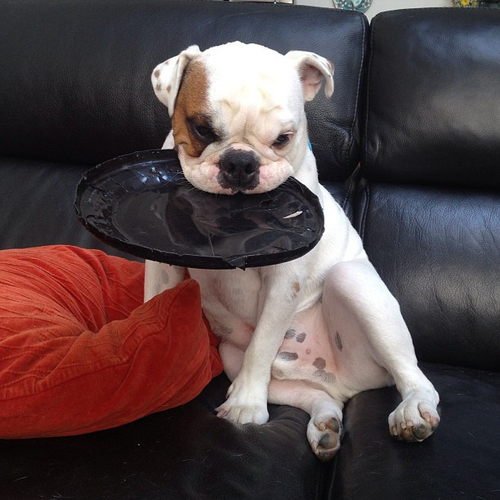}
            \tiny\raggedright
        \end{subfigure}
        \hfill
        \begin{subfigure}[t]{\colwid}
            \includegraphics[width=\linewidth]{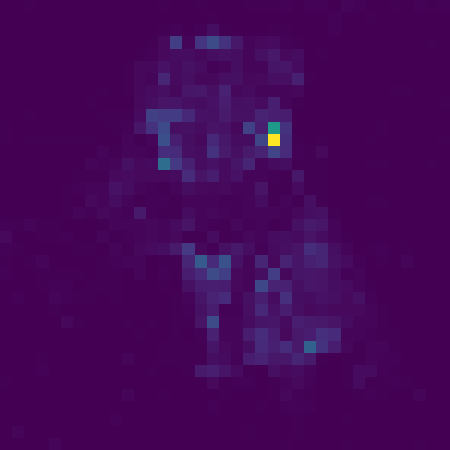}
            \tiny\raggedright
\textbf{Head 1}: A puppy holding a black disk on the couch \\
        \end{subfigure}
        \hfill
        \begin{subfigure}[t]{\colwid}
            \includegraphics[width=\linewidth]{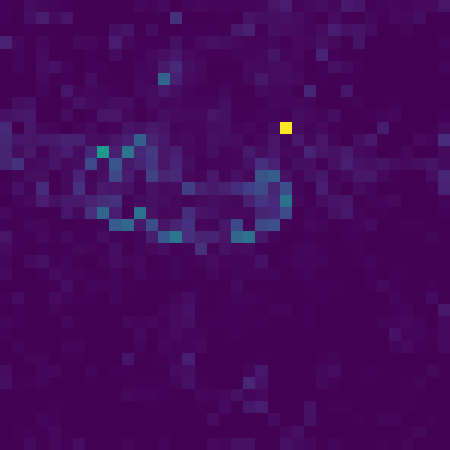}
            \tiny\raggedright
\textbf{Head 2}: A dog sitting on a couch holding a frisbee in its mouth. \\
        \end{subfigure}
        \hfill
        \begin{subfigure}[t]{\colwid}
            \includegraphics[width=\linewidth]{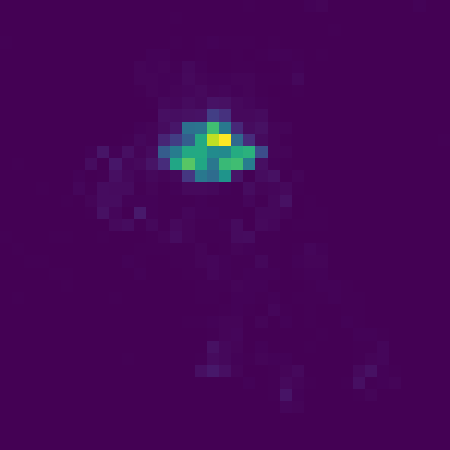}
            \tiny\raggedright
\textbf{Head 3}: A  dog holding a plate while sitting on a chair. \\
        \end{subfigure}
        \hfill
        \begin{subfigure}[t]{\colwid}
            \includegraphics[width=\linewidth]{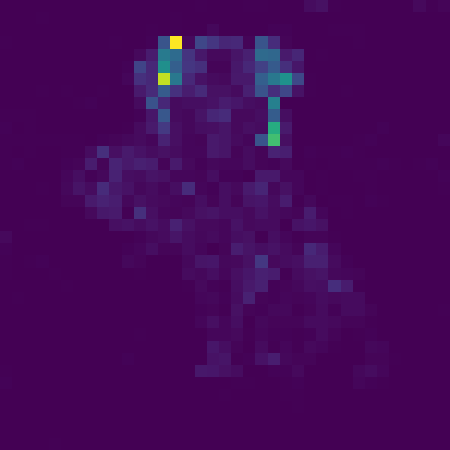}
            \tiny\raggedright
\textbf{Head 9}: A bull dog holding a frisbee in it's mouth. \\
        \end{subfigure}
        \hfill
        \begin{subfigure}[t]{\colwid}
            \includegraphics[width=\linewidth]{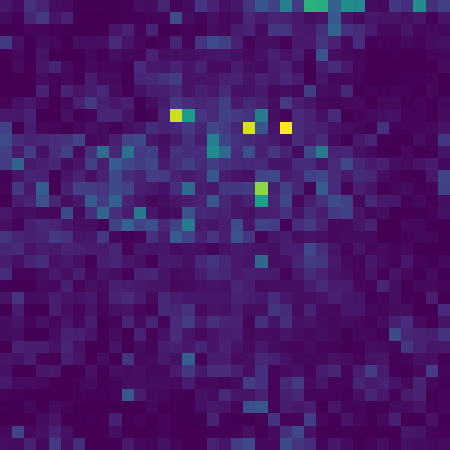}
            \tiny\raggedright
\textbf{Head 10}: a brown black and white dog and a black frisbee \\
        \end{subfigure}

        \begin{subfigure}[t]{\colwid}
            \includegraphics[width=\linewidth]{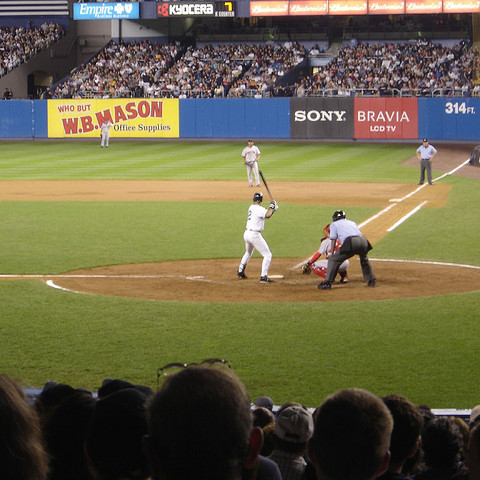}
            \tiny\raggedright
        \end{subfigure}
        \hfill
        \begin{subfigure}[t]{\colwid}
            \includegraphics[width=\linewidth]{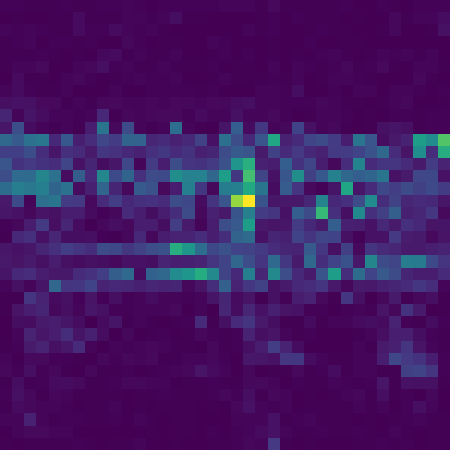}
            \tiny\raggedright
            \textbf{Head 0}: several of the baseball players are in view \\
        \end{subfigure}
        \hfill
        \begin{subfigure}[t]{\colwid}
            \includegraphics[width=\linewidth]{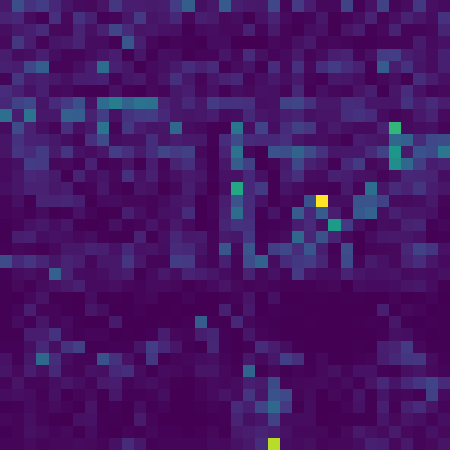}
            \tiny\raggedright
            \textbf{Head 1}: A large crowd is watching a baseball game. \\
        \end{subfigure}
        \hfill
        \begin{subfigure}[t]{\colwid}
            \includegraphics[width=\linewidth]{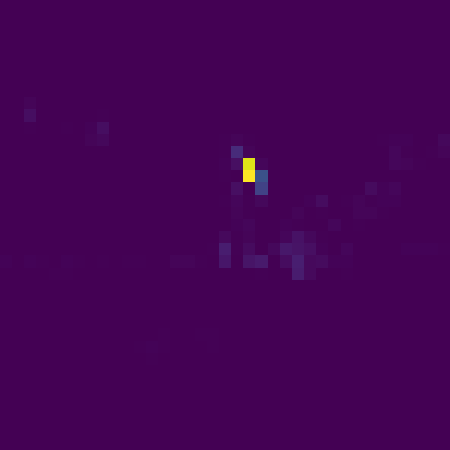}
            \tiny\raggedright
            \textbf{Head 3}: Players on a baseball field during a game. \\
        \end{subfigure}
        \hfill
        \begin{subfigure}[t]{\colwid}
            \includegraphics[width=\linewidth]{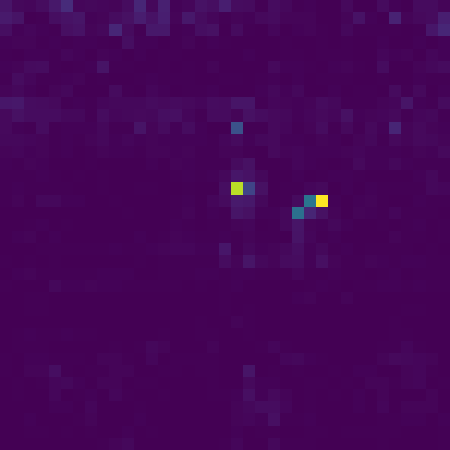}
            \tiny\raggedright
            \textbf{Head 5}: A man is up to bat at a professional baseball game. \\
        \end{subfigure}
        \hfill
        \begin{subfigure}[t]{\colwid}
            \includegraphics[width=\linewidth]{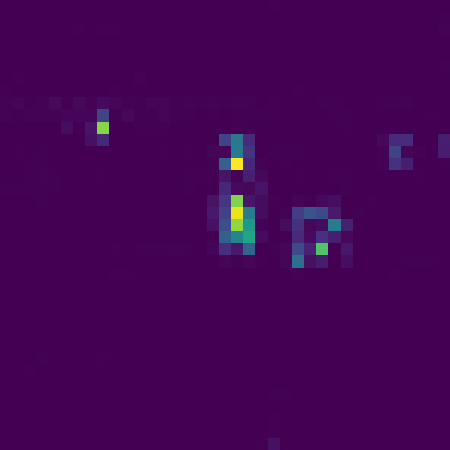}
            \tiny\raggedright
            \textbf{Head 6}: a man holding up his baseball bat during a baseball game  \\
        \end{subfigure}

 \begin{subfigure}[t]{\colwid}
    \includegraphics[width=\linewidth]{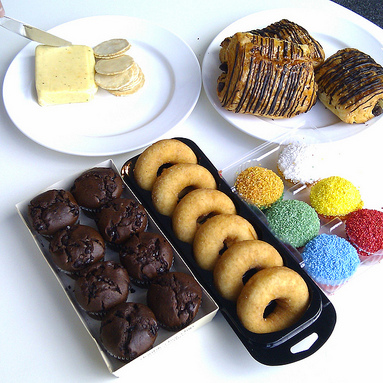}
    \tiny\raggedright
\end{subfigure}
\hfill
\begin{subfigure}[t]{\colwid}
    \includegraphics[width=\linewidth]{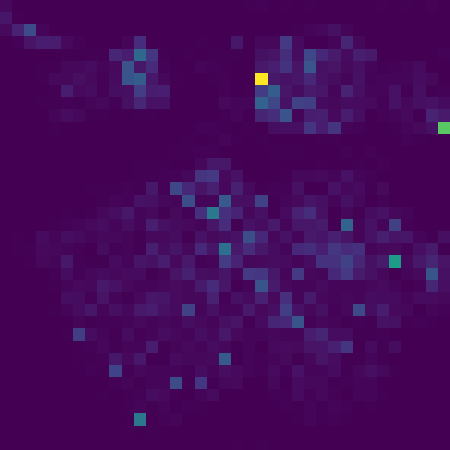}
    \tiny\raggedright
    \textbf{Head 1}: A white round table filled with some assorted treats. \\
\end{subfigure}
\hfill
\begin{subfigure}[t]{\colwid}
    \includegraphics[width=\linewidth]{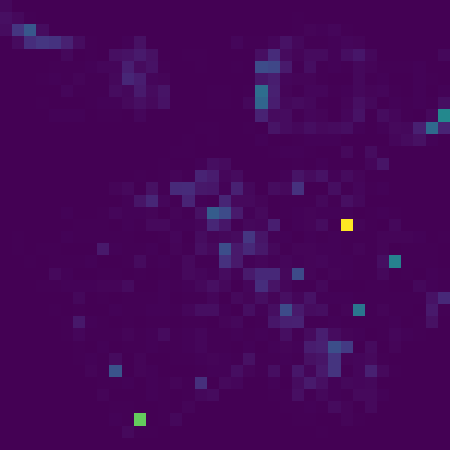}
    \tiny\raggedright
    \textbf{Head 6}: A number of pastry items on a table \\
\end{subfigure}
\hfill
\begin{subfigure}[t]{\colwid}
    \includegraphics[width=\linewidth]{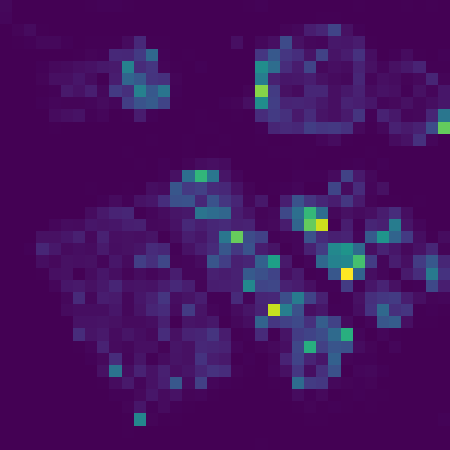}
    \tiny\raggedright
    \textbf{Head 7}: A dessert tray with donuts, cupcakes, and muffins \\
\end{subfigure}
\hfill
\begin{subfigure}[t]{\colwid}
    \includegraphics[width=\linewidth]{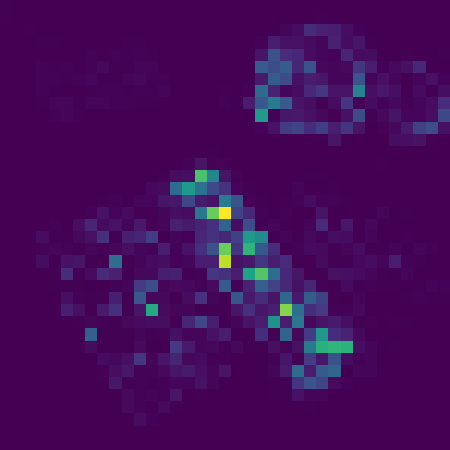}
    \tiny\raggedright
    \textbf{Head 10}: A white plate topped with chocolates, donuts, and other goodies. \\
\end{subfigure}
\hfill
\begin{subfigure}[t]{\colwid}
    \includegraphics[width=\linewidth]{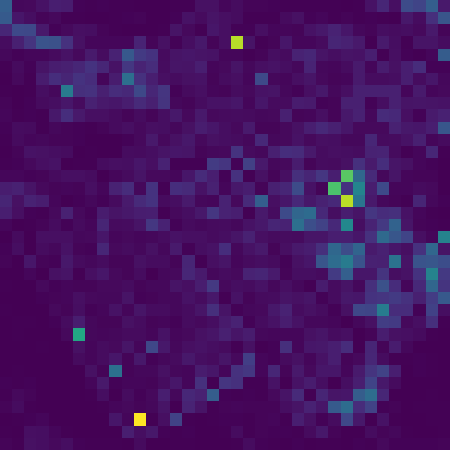}
    \tiny\raggedright
    \textbf{Head 11}: Donuts and other goodies on a table \\
\end{subfigure}

        \vspace{-0.3cm}
        \caption{Sample images from the training set paired with their corresponding captions and the attention maps selected for alignment during the last epoch of training.}
        \label{fig:attn_maps_matches}
        \vspace{-0.3cm}
    \end{figure*}

Fig.~\ref{fig:attn_maps_matches} presents examples from the training set, showcasing images paired with their corresponding captions and the attention maps selected for alignment. Despite describing the same scene, variations in the captions lead the alignment procedure to focus on different regions of the image. For instance, in the first row, the caption mentioning the fans also focuses on the background, while captions that reference only the player, the ball, and the racket do not.

\section{Additional Qualitative Results}

\begin{figure*}[t]
    \centering
    \normalsize
    \resizebox{\linewidth}{!}{
        \setlength{\tabcolsep}{.08em}
        \begin{tabular}{c p{0mm} c p{0mm} c p{0mm} c p{2mm} c p{0mm} c p{0mm} c p{0mm} c}
            \addlinespace[2mm]
             &&&& \textbf{\huge with} && \textbf{\huge without} &&&&&& \textbf{\huge with} && \textbf{\huge without} \\
             \addlinespace[2mm]
            \textbf{\huge Image} && \textbf{\huge Ground-truth} && \textbf{\huge Bkg. Cleaning} && \textbf{\huge Bkg. Cleaning} &&  \textbf{\huge Image} && \textbf{\huge Ground-truth} && \textbf{\huge Bkg. Cleaning} && \textbf{\huge Bkg. Cleaning} \\
            \addlinespace[2mm]
            \includegraphics[height=0.3\textwidth]{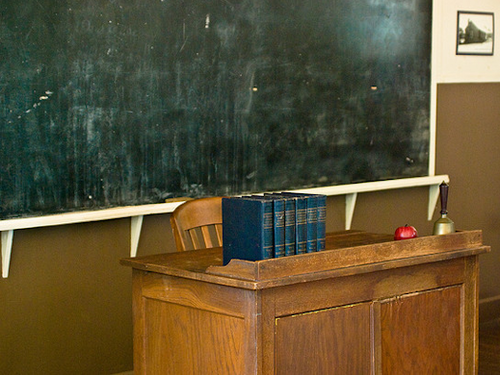} &&
            \includegraphics[height=0.3\textwidth]{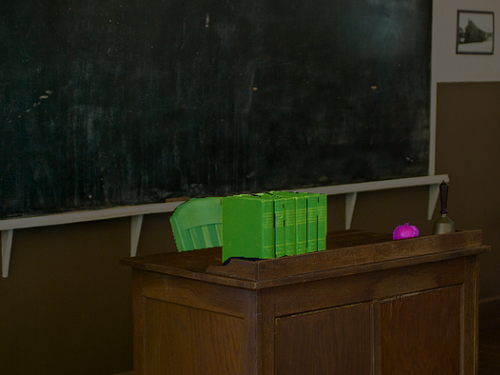} && \includegraphics[height=0.3\textwidth]{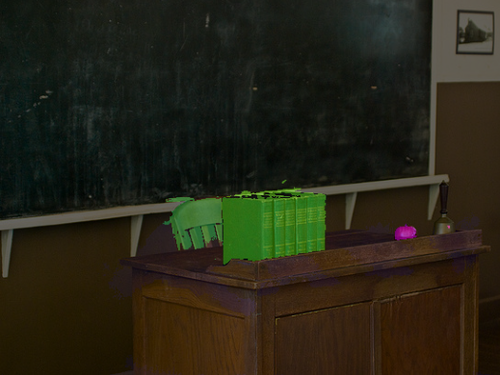} && \includegraphics[height=0.3\textwidth]{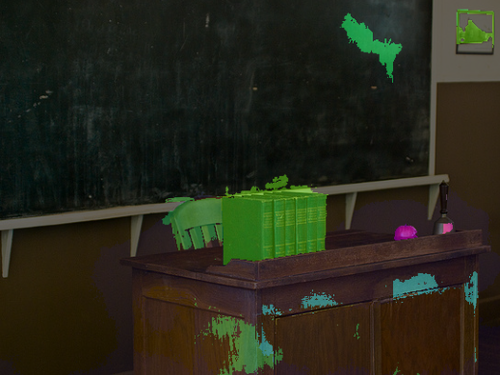} &&
            \includegraphics[height=0.3\textwidth]{figures/qualitatives/obj_bg_clean/2_image.jpg} &&
            \includegraphics[height=0.3\textwidth]{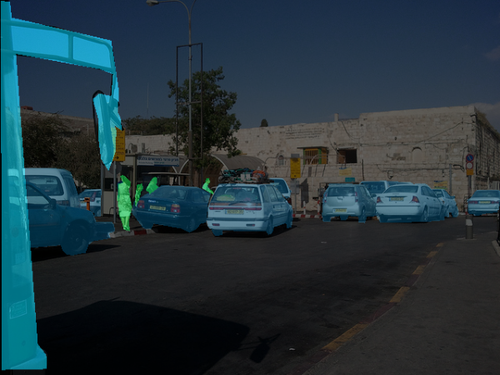} && \includegraphics[height=0.3\textwidth]{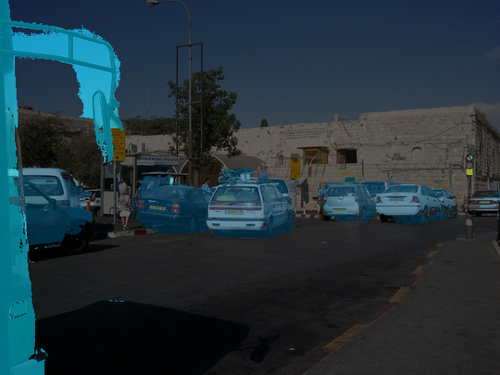} && \includegraphics[height=0.3\textwidth]{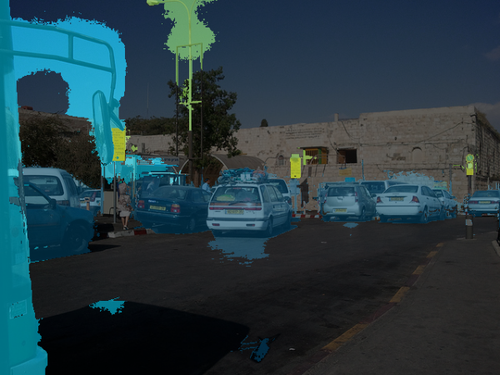} \\
            \includegraphics[height=0.3\textwidth]{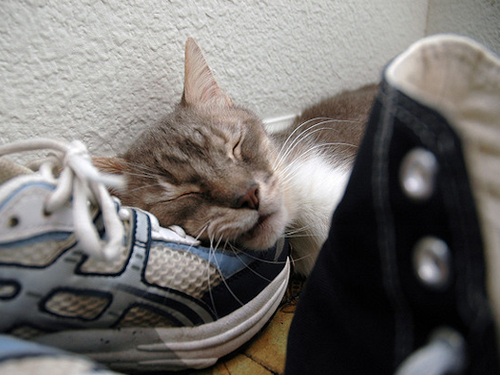} &&
            \includegraphics[height=0.3\textwidth]{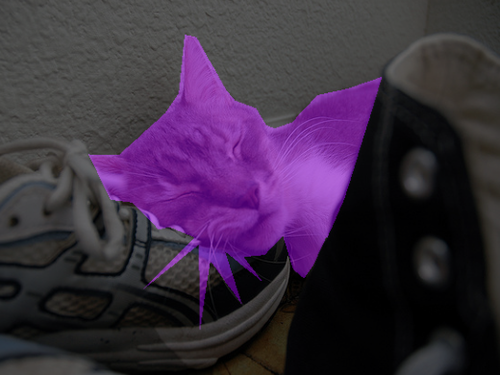} && \includegraphics[height=0.3\textwidth]{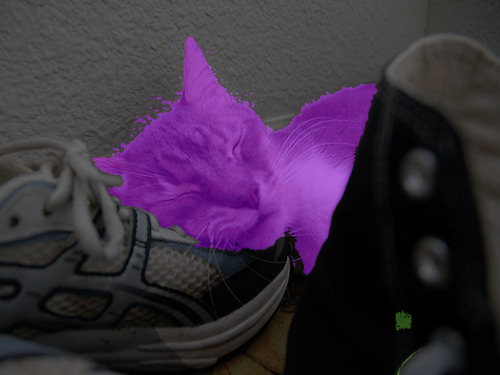} && \includegraphics[height=0.3\textwidth]{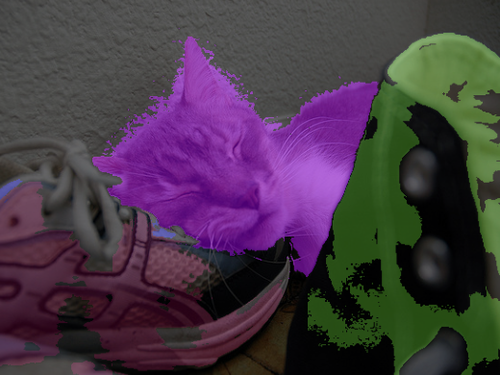} &&
            \includegraphics[height=0.3\textwidth]{figures/qualitatives/obj_bg_clean/4_image.jpg} &&
            \includegraphics[height=0.3\textwidth]{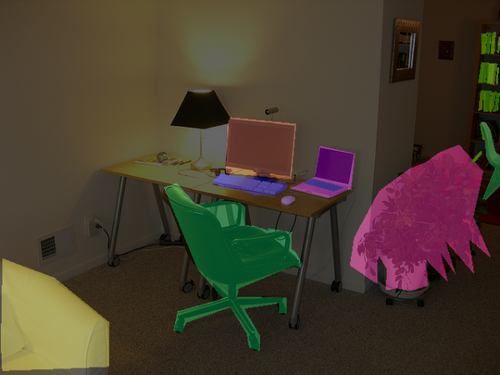} && \includegraphics[height=0.3\textwidth]{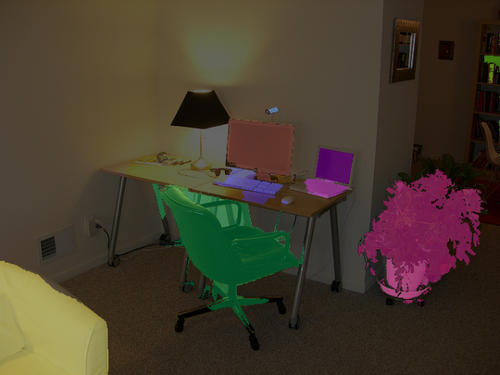} && \includegraphics[height=0.3\textwidth]{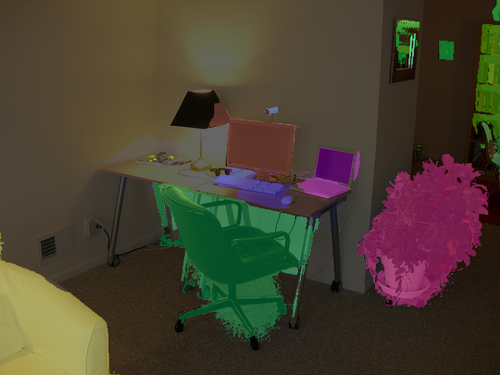} \\
            \includegraphics[height=0.3\textwidth]{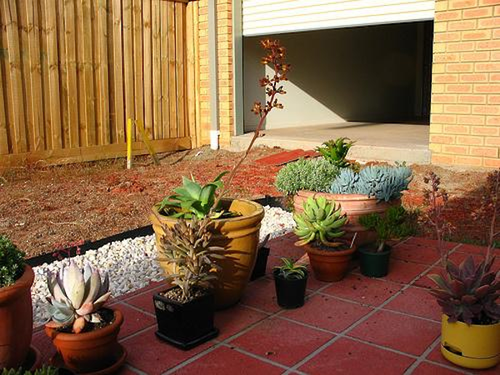} &&
            \includegraphics[height=0.3\textwidth]{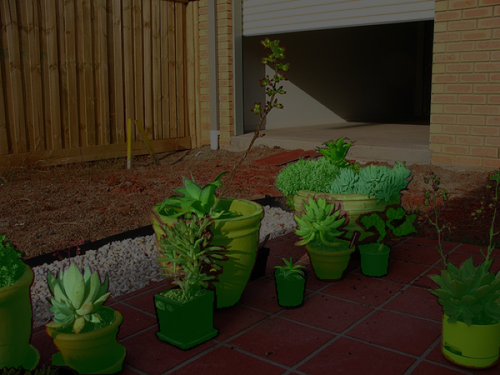} && \includegraphics[height=0.3\textwidth]{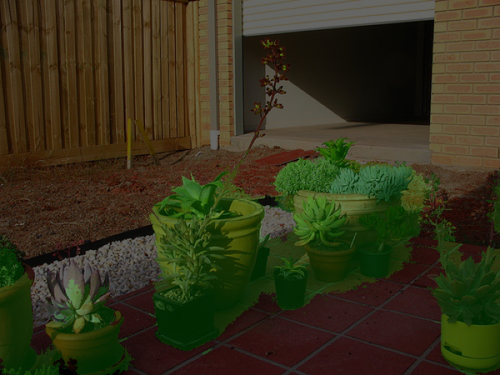} && \includegraphics[height=0.3\textwidth]{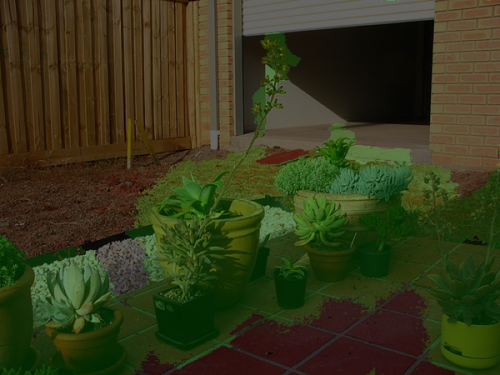} &&
            \includegraphics[height=0.3\textwidth]{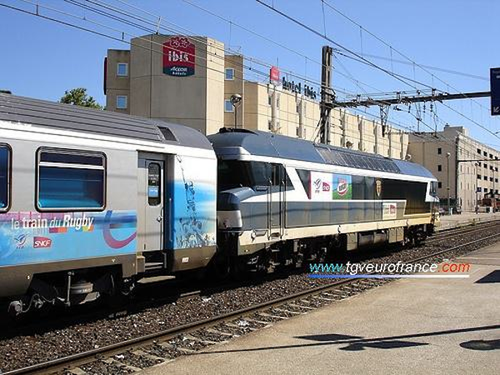} &&
            \includegraphics[height=0.3\textwidth]{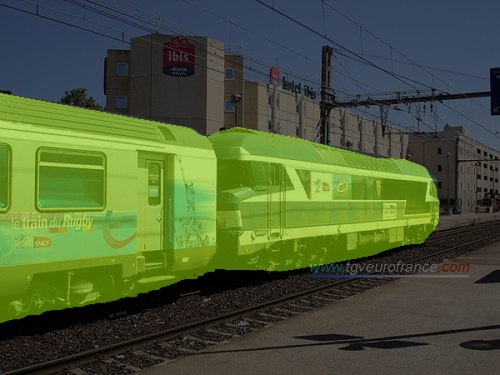} && \includegraphics[height=0.3\textwidth]{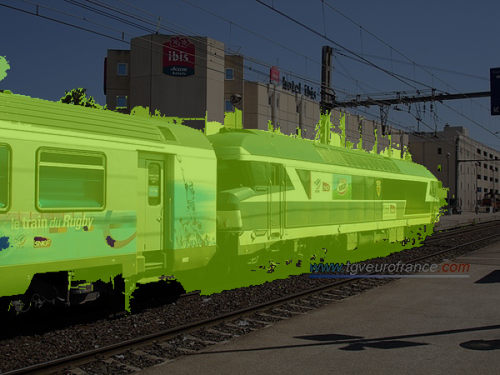} && \includegraphics[height=0.3\textwidth]{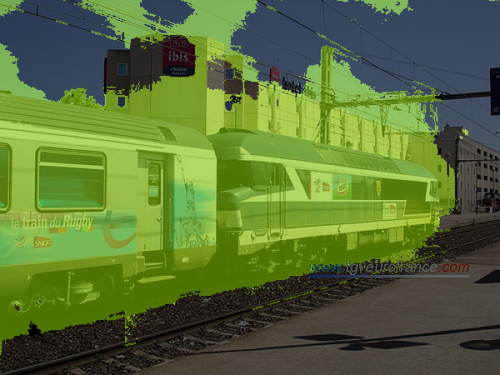} \\
            \includegraphics[height=0.3\textwidth]{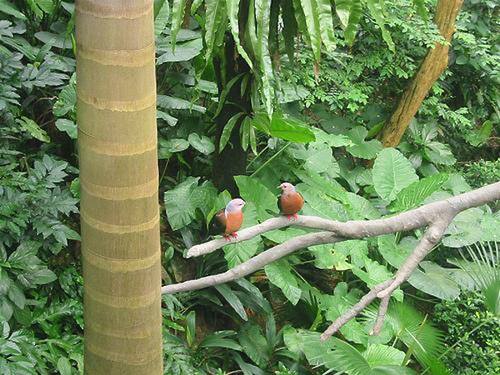} &&
            \includegraphics[height=0.3\textwidth]{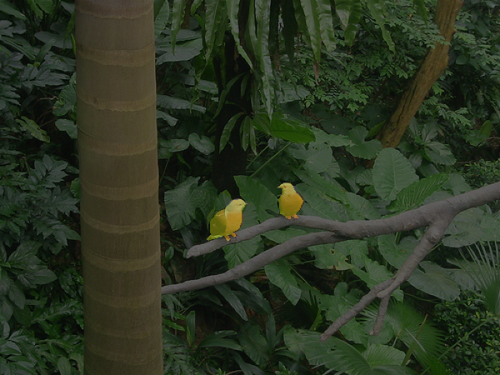} && \includegraphics[height=0.3\textwidth]{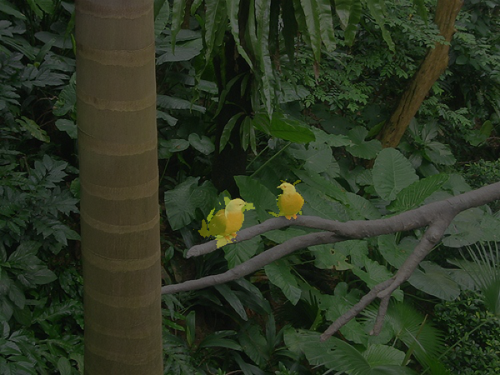} && \includegraphics[height=0.3\textwidth]{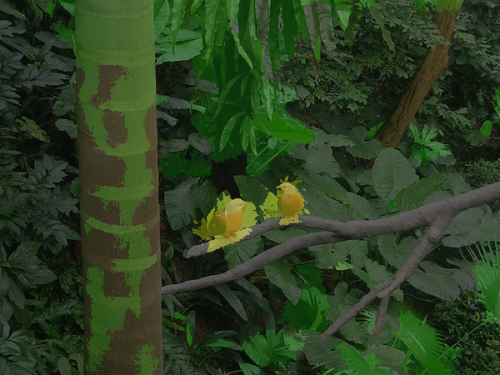} &&
            \includegraphics[height=0.3\textwidth]{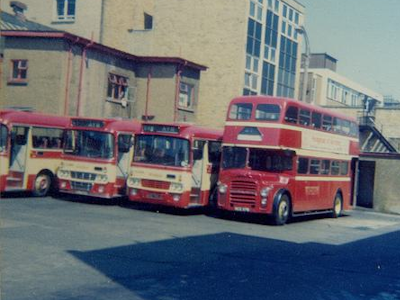} &&
            \includegraphics[height=0.3\textwidth]{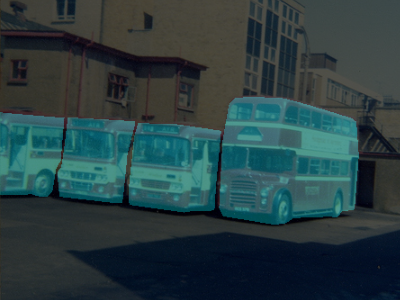} && \includegraphics[height=0.3\textwidth]{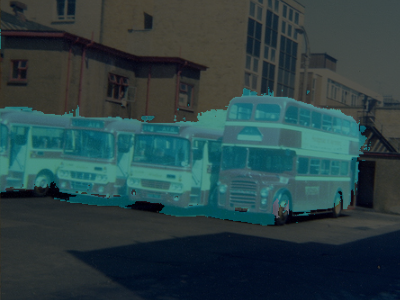} && \includegraphics[height=0.3\textwidth]{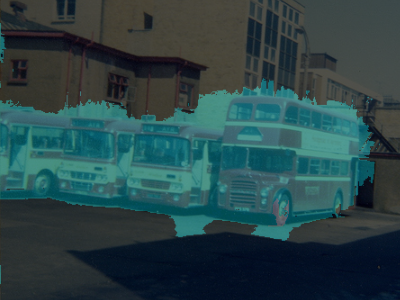} \\
        \end{tabular}
        }
    \vspace{-.2cm}
    \caption{Qualitative results obtained with and without the proposed background cleaning strategy, on COCO Object and Pascal VOC.}
    \label{fig:qualitatives_background}
    \vspace{-.2cm}
\end{figure*}

\tinytit{Effect of Background Cleaning}
Fig.~\ref{fig:qualitatives_background} shows a set of qualitative results in which we highlight the advantages of using the proposed background cleaning procedure with respect to directly thresholding the similarities with the input categories to detect the background. In particular, the first two rows show four qualitatives on images from COCO Object and the last two rows from VOC. These results demonstrate that background cleaning removes the noise in the background from the image and improves the fitting of the masks on the foreground objects. These findings are reflected in the results reported in Table~\ref{tab:ablation_background} of the main paper.

\begin{figure*}[t]
    \centering
    \large
    \resizebox{\linewidth}{!}{
        \setlength{\tabcolsep}{.08em}
        \begin{tabular}{c p{2mm} c p{2mm} c p{2mm} c p{2mm} c p{2mm} c}
            \addlinespace[2mm]
            \includegraphics[height=0.3\textwidth]{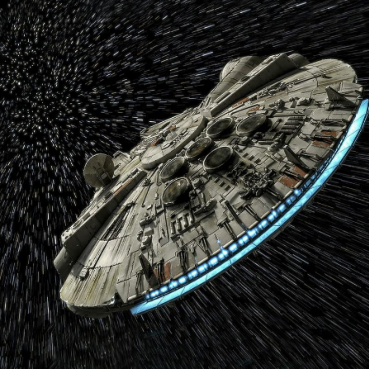} && \includegraphics[height=0.3\textwidth]{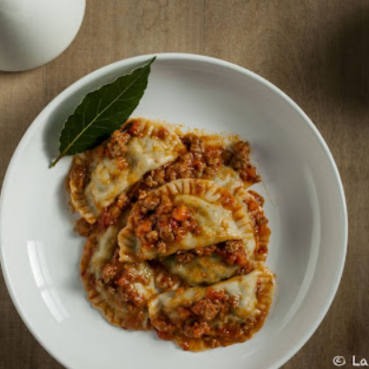} && \includegraphics[height=0.3\textwidth]{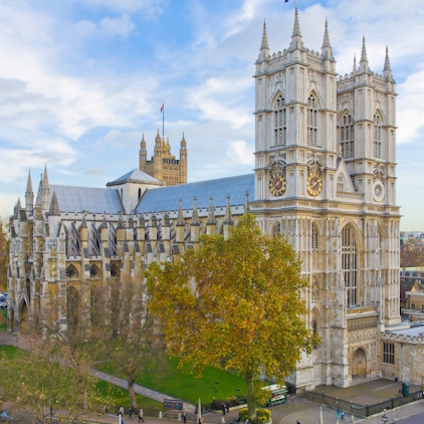} &&
            \includegraphics[height=0.3\textwidth]{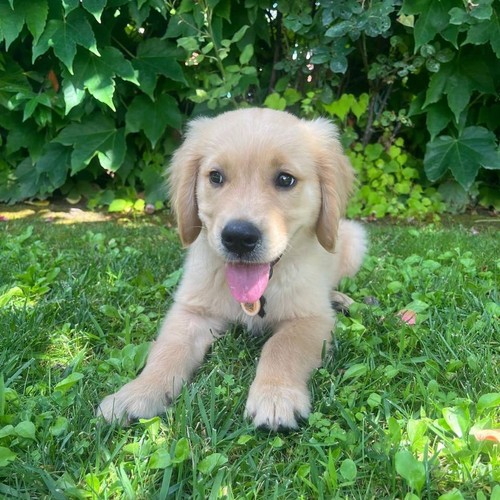} &&
            \includegraphics[height=0.3\textwidth]{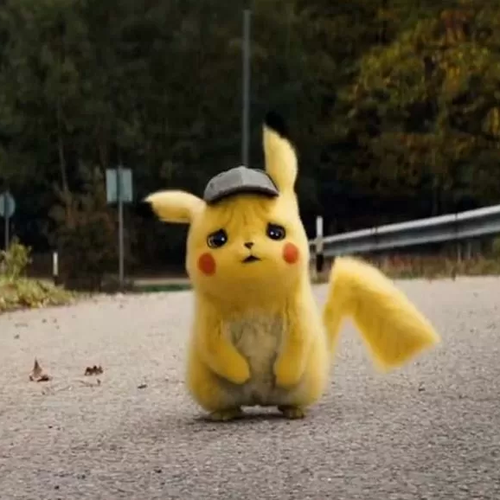} && \includegraphics[height=0.3\textwidth]{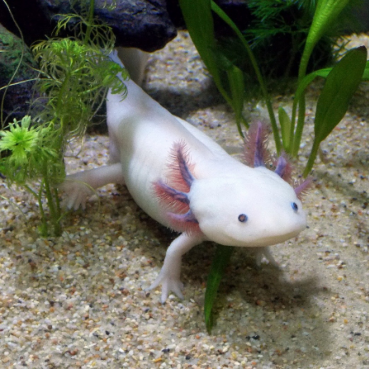} \\
            \includegraphics[height=0.3\textwidth]{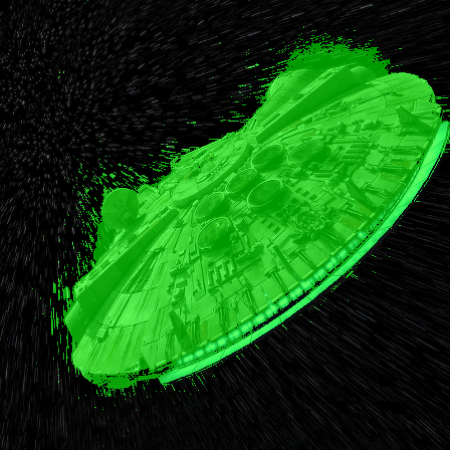} && \includegraphics[height=0.3\textwidth]{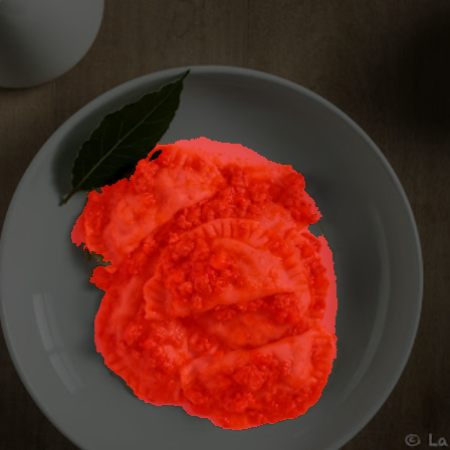} &&
            \includegraphics[height=0.3\textwidth]{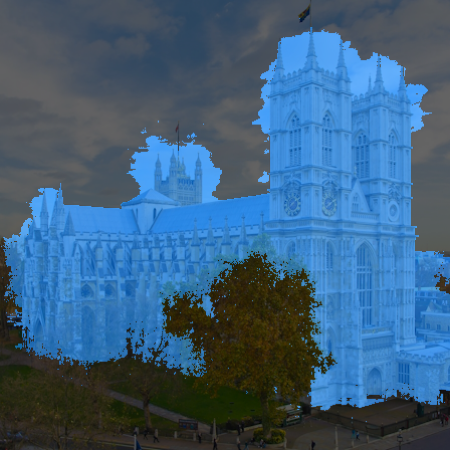} &&
            \includegraphics[height=0.3\textwidth]{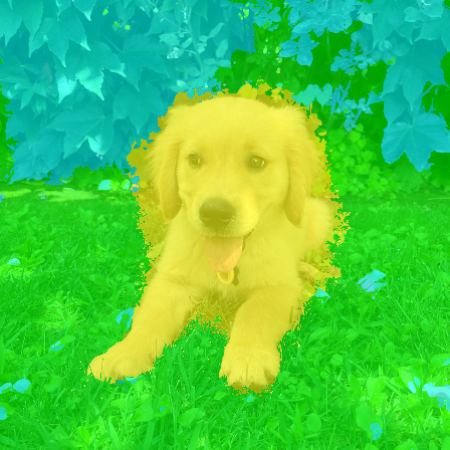} &&
            \includegraphics[height=0.3\textwidth]{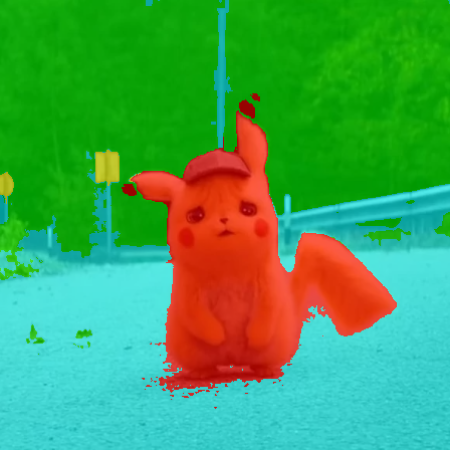} && \includegraphics[height=0.3\textwidth]{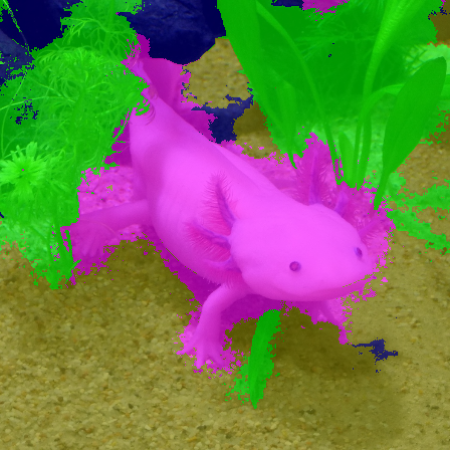} \\
            \makecell{\texttt{\textbf{\color{green_into_the_wild}millennium falcon}}} && \makecell{\texttt{\textbf{\color{red_into_the_wild}tordelli}}} && \makecell{\texttt{\textbf{\color{light_blue_into_the_wild}westminster abbey}}} &&
            \makecell{\texttt{\textbf{\color{yellow_into_the_wild}golden retriever}} \\ \texttt{\textbf{\color{yellow_into_the_wild} puppy}}\\ \texttt{\textbf{\color{green_into_the_wild}grass field}} \\ \texttt{\textbf{\color{light_blue_into_the_wild}vegetation}}}
            && \makecell{\texttt{\textbf{\color{red_into_the_wild}pikachu}} \\ \texttt{\textbf{\color{yellow_into_the_wild}traffic sign}} \\ \texttt{\textbf{\color{green_into_the_wild}forest}} \\ \texttt{\textbf{\color{light_blue_into_the_wild}route}}} && \makecell{\texttt{\textbf{\color{fuchsia_into_the_wild}axolotl}} \\ \texttt{\textbf{\color{light_blue_into_the_wild}stones}} \\ \texttt{\textbf{\color{green_into_the_wild}leaves}}\\ \texttt{\textbf{\color{yellow_into_the_wild}gravel}}}\\

        \end{tabular}
        }
    \vspace{-.3cm}
    \caption{"In-the-wild" segmentation results obtained by prompting \ours with uncommon textual categories on images retrieved from the web.}
    \label{fig:into_the_wild}
    \vspace{-.4cm}
\end{figure*}

\tit{In-the-Wild Qualitative Examples} Fig.~\ref{fig:into_the_wild} depicts a few examples of ``in-the-wild" segmentation, obtained by providing to \ours sample images from the web and asking it to segment uncommon categories, such as ``\texttt{pikachu}", ``\texttt{millennium falcon}", and ``\texttt{westminster abbey}", and free-form text, like ``\texttt{golden retriever puppy}". On the left, we show three examples in which we task the model with also finding the background, while exploiting the background cleaning procedure, and, on the right, three examples in which the model has to assign a provided category to each pixel. The high quality of the resulting masks demonstrates the efficacy of our approach, even on out-of-domain images. From these examples, we can appreciate the capabilities of the model in combining the knowledge from CLIP with the semantic localization of DINOv2 on unconventional concepts, such as fictional character names and proper nouns of historical buildings.

\begin{figure*}[t]
    \centering
    \normalsize
    \resizebox{\linewidth}{!}{
        \setlength{\tabcolsep}{.06em}
        \begin{tabular}{ c p{0mm} c p{0mm} c p{0mm} c p{0mm} c p{0mm} c p{0mm} c}
            \addlinespace[2mm]
            && \textbf{\huge Image} && \textbf{\huge Ground-truth} && \textbf{\huge FreeDA~\cite{barsellotti2024training}} && \textbf{\huge ProxyCLIP~\cite{lan2024proxyclip}} && \textbf{\huge CLIP-DINOiser~\cite{wysoczanska2024clip}} && \textbf{\huge \ours (Ours)} \\
            \addlinespace[3mm]
            \parbox[t]{10mm}{\multirow{2}{*}[4em]{\rotatebox[origin=c]{90}{\huge\textbf{Pascal VOC}}}} &&
            \includegraphics[height=0.3\textwidth]{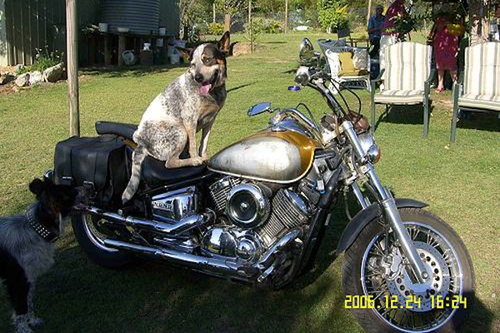} & &
            \includegraphics[height=0.3\textwidth]{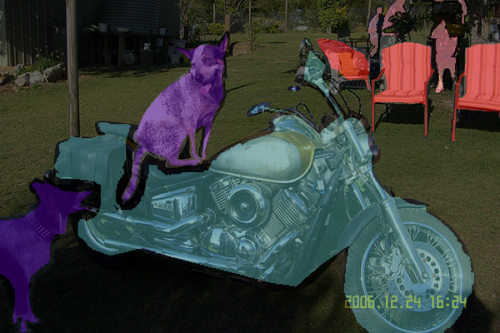} & & \includegraphics[height=0.3\textwidth]{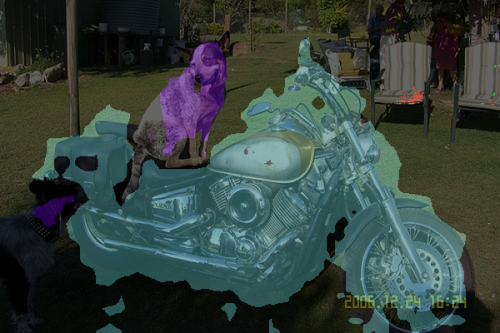} & &
            \includegraphics[height=0.3\textwidth]{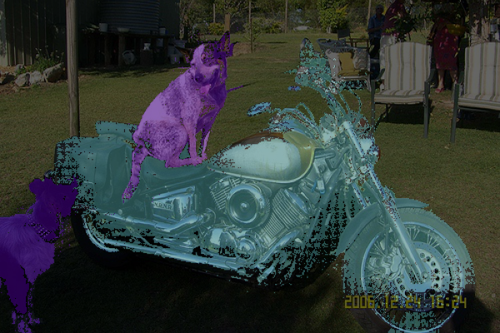} &&
            \includegraphics[height=0.3\textwidth]{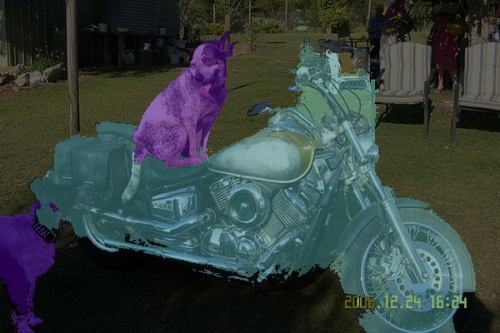} && \includegraphics[height=0.3\textwidth]{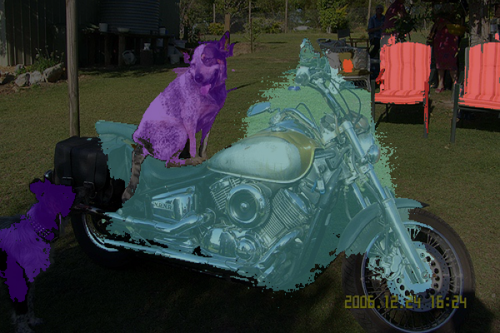} \\
             &&
            \includegraphics[height=0.3\textwidth]{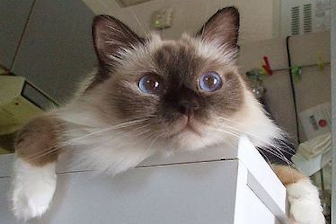} & &
            \includegraphics[height=0.3\textwidth]{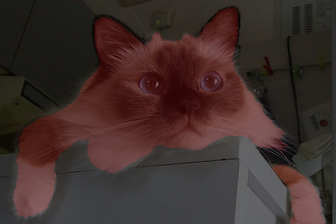} & & \includegraphics[height=0.3\textwidth]{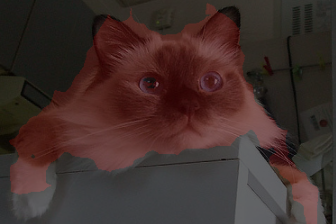} & &
            \includegraphics[height=0.3\textwidth]{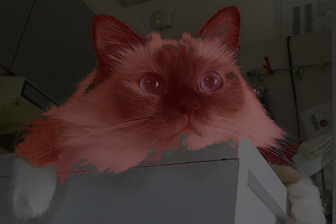} &&
            \includegraphics[height=0.3\textwidth]{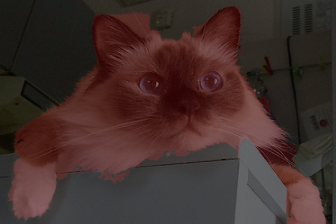} && \includegraphics[height=0.3\textwidth]{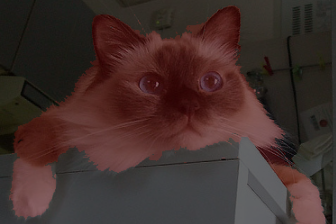} \\
            \parbox[t]{10mm}{\multirow{2}{*}[4.5em]{\rotatebox[origin=c]{90}{\huge\textbf{Pascal Context}}}} &&
            \includegraphics[height=0.3\textwidth]{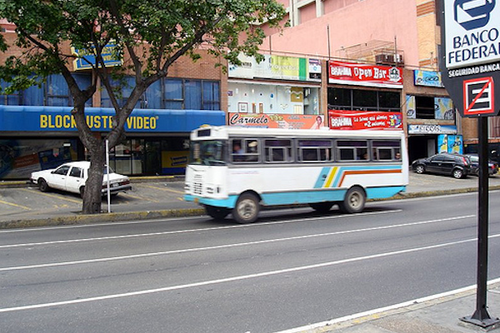} & &
            \includegraphics[height=0.3\textwidth]{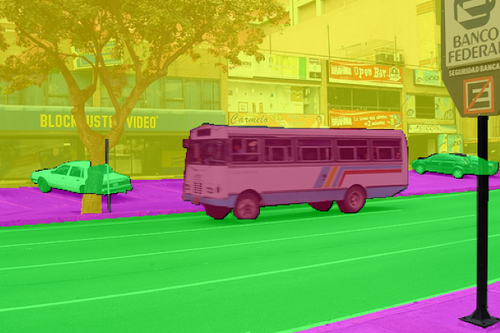} & & \includegraphics[height=0.3\textwidth]{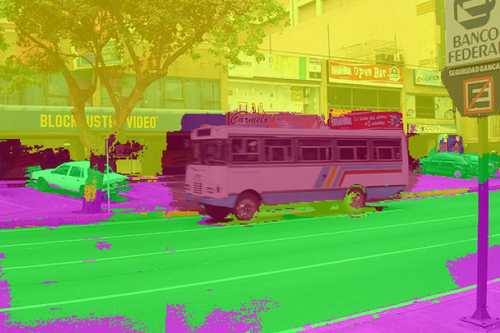} & &
            \includegraphics[height=0.3\textwidth]{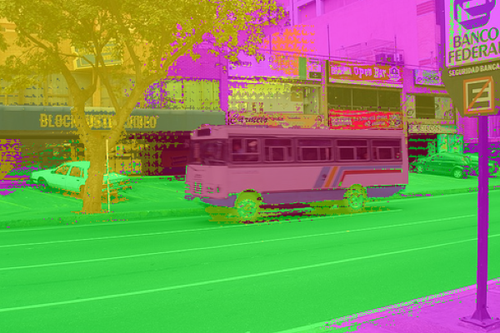} &&
            \includegraphics[height=0.3\textwidth]{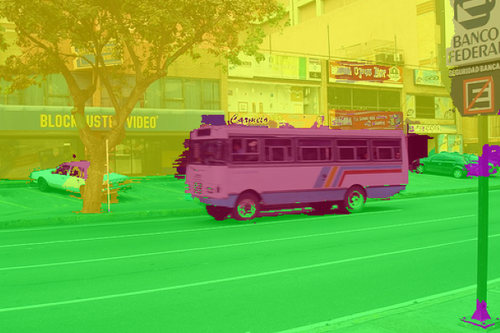} && \includegraphics[height=0.3\textwidth]{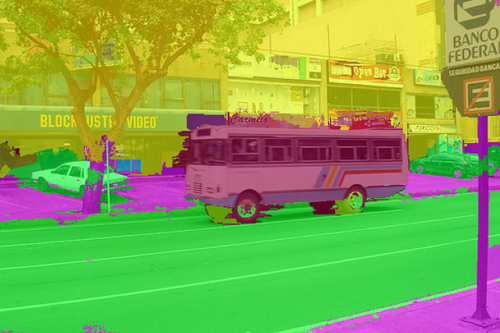} \\
             &&
            \includegraphics[height=0.3\textwidth]{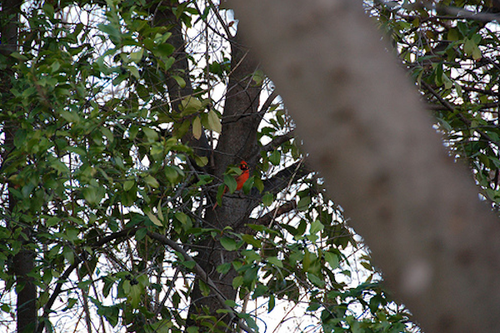} & &
            \includegraphics[height=0.3\textwidth]{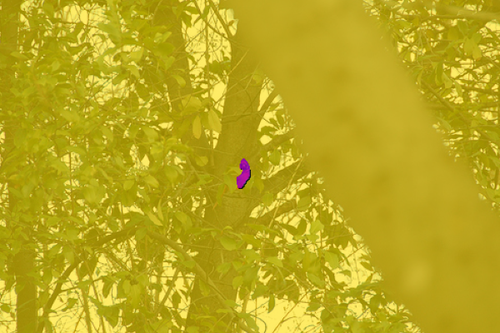} & & \includegraphics[height=0.3\textwidth]{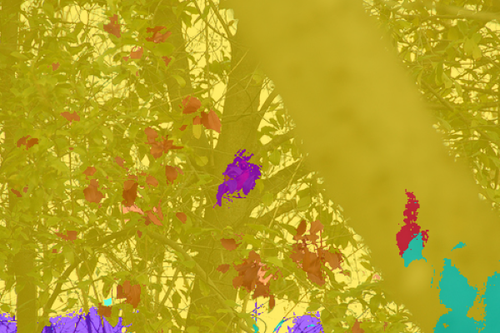} & &
            \includegraphics[height=0.3\textwidth]{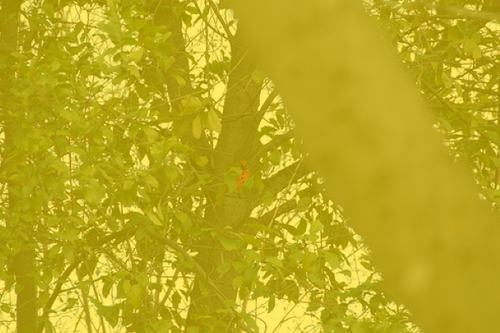} &&
            \includegraphics[height=0.3\textwidth]{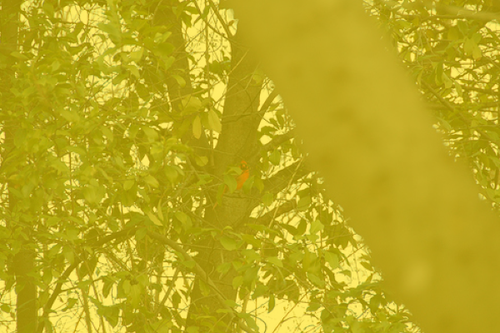} && \includegraphics[height=0.3\textwidth]{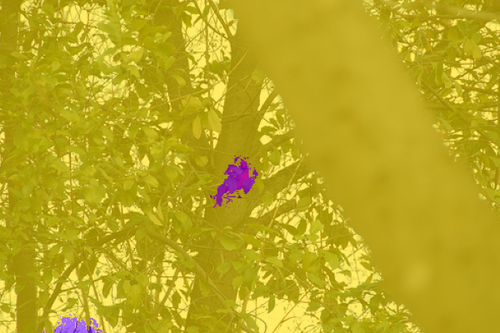} \\
           \parbox[t]{10mm}{\multirow{2}{*}[4em]{\rotatebox[origin=c]{90}{\huge\textbf{COCO Stuff}}}} &&
            \includegraphics[height=0.3\textwidth]{figures/qualitatives/stuff/2r_image.png} & &
            \includegraphics[height=0.3\textwidth]{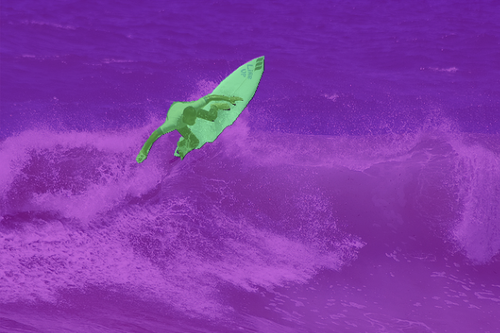} & & \includegraphics[height=0.3\textwidth]{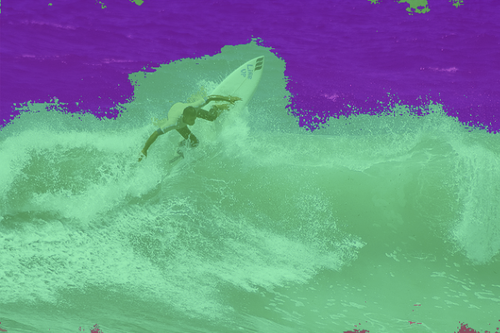} & &
            \includegraphics[height=0.3\textwidth]{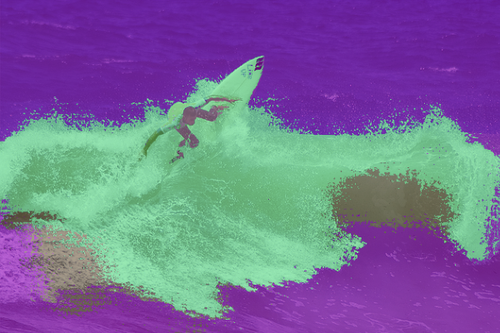} &&
            \includegraphics[height=0.3\textwidth]{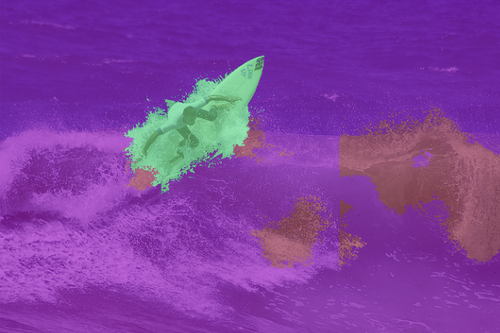} && \includegraphics[height=0.3\textwidth]{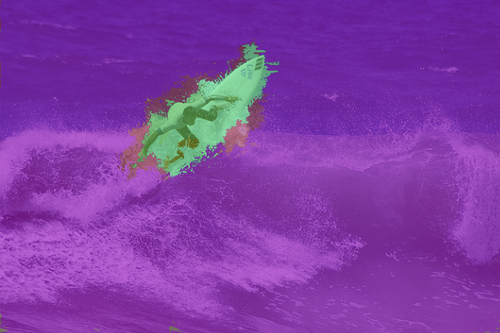} \\
            &&
            \includegraphics[height=0.3\textwidth]{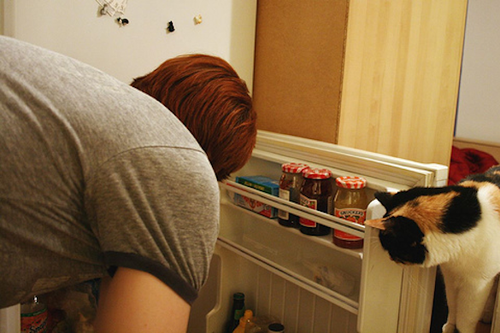} & &
            \includegraphics[height=0.3\textwidth]{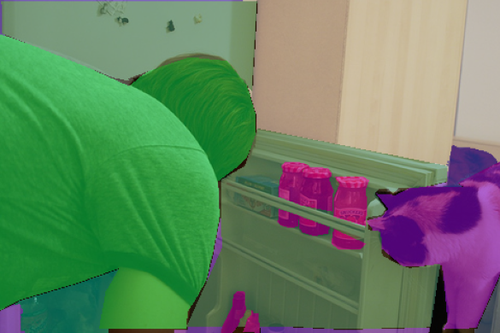} & & \includegraphics[height=0.3\textwidth]{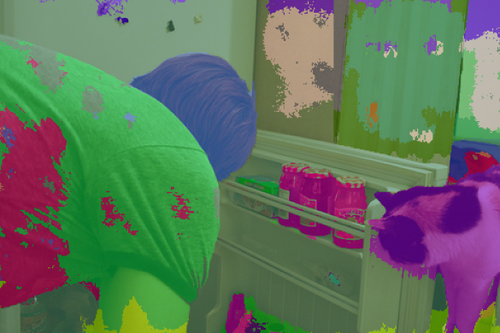} & &
            \includegraphics[height=0.3\textwidth]{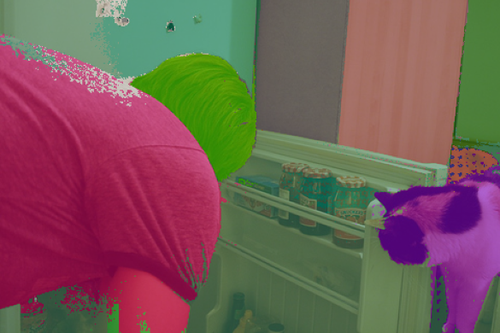} &&
            \includegraphics[height=0.3\textwidth]{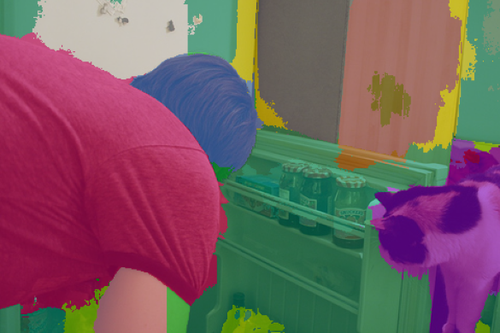} && \includegraphics[height=0.3\textwidth]{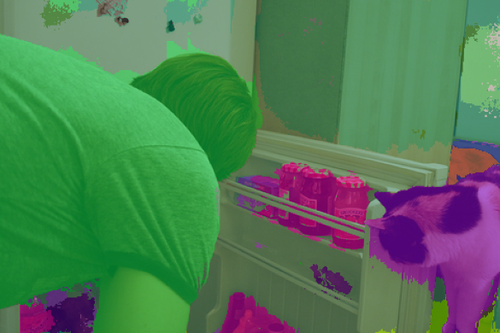} \\
            \parbox[t]{10mm}{\multirow{2}{*}[4em]{\rotatebox[origin=c]{90}{\huge\textbf{Cityscapes}}}} &&
            \includegraphics[height=0.3\textwidth]{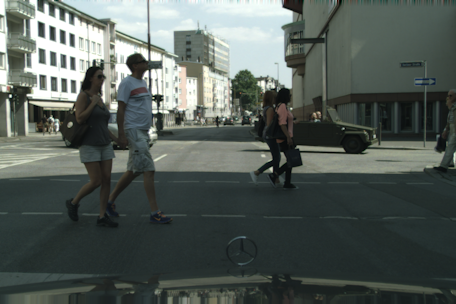} & &
            \includegraphics[height=0.3\textwidth]{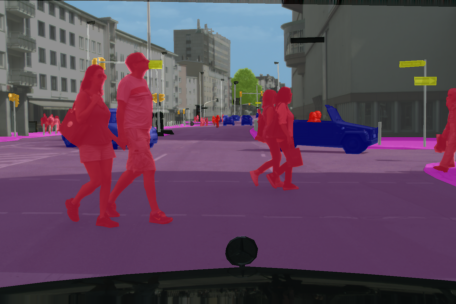} & & \includegraphics[height=0.3\textwidth]{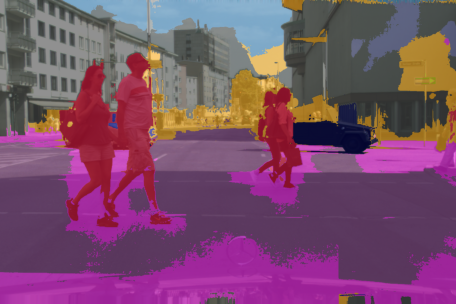} & &
            \includegraphics[height=0.3\textwidth]{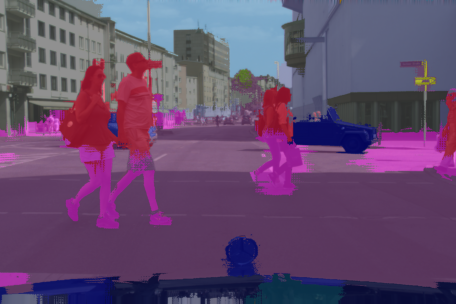} &&
            \includegraphics[height=0.3\textwidth]{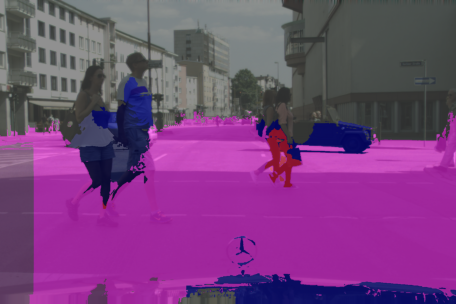} && \includegraphics[height=0.3\textwidth]{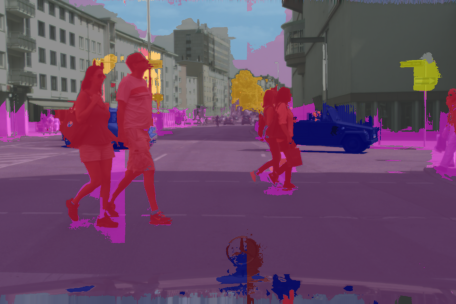} \\
            &&
            \includegraphics[height=0.3\textwidth]{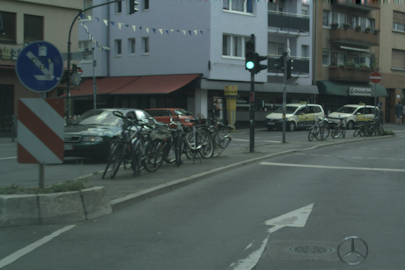} & &
            \includegraphics[height=0.3\textwidth]{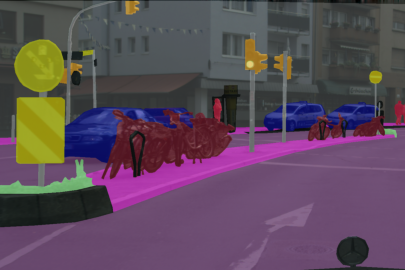} & & \includegraphics[height=0.3\textwidth]{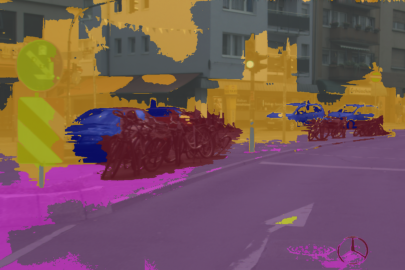} & &
            \includegraphics[height=0.3\textwidth]{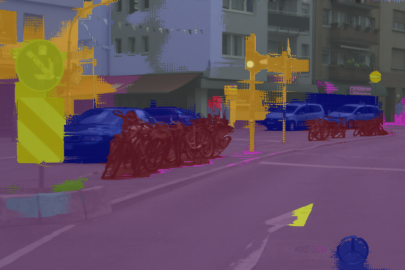} &&
            \includegraphics[height=0.3\textwidth]{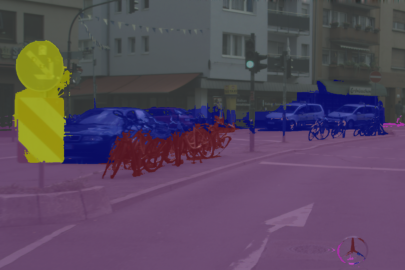} && \includegraphics[height=0.3\textwidth]{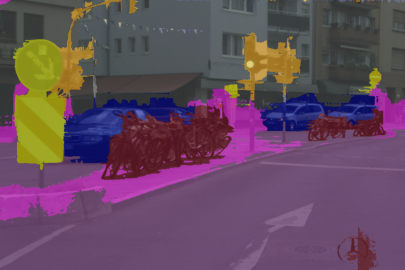} \\
           \parbox[t]{10mm}{\multirow{2}{*}[3em]{\rotatebox[origin=c]{90}{\huge\textbf{ADE20K}}}} &&
            \includegraphics[height=0.3\textwidth]{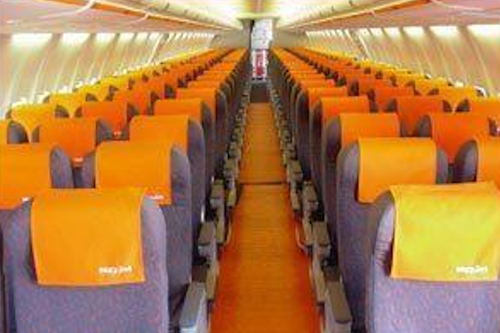} & &
            \includegraphics[height=0.3\textwidth]{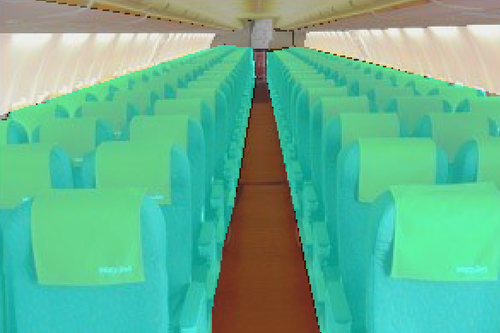} & & \includegraphics[height=0.3\textwidth]{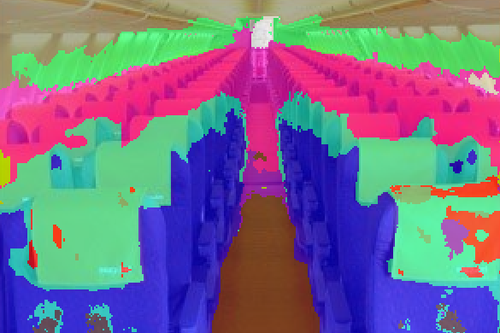} & &
            \includegraphics[height=0.3\textwidth]{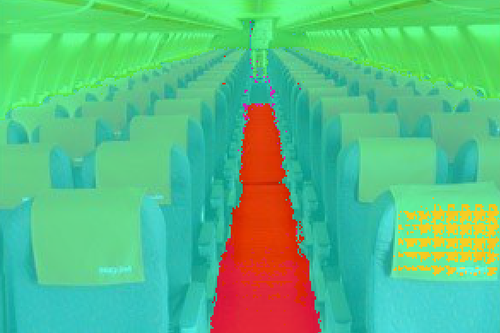} &&
            \includegraphics[height=0.3\textwidth]{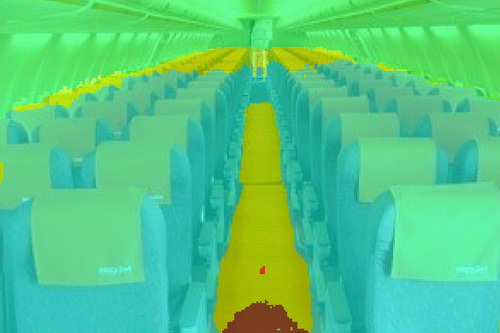} && \includegraphics[height=0.3\textwidth]{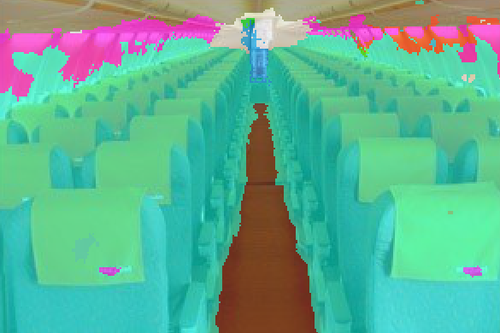} \\
             &&
            \includegraphics[height=0.3\textwidth]{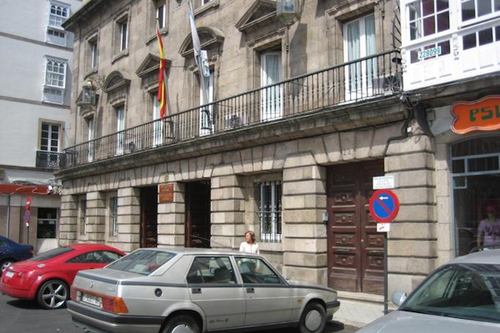} & &
            \includegraphics[height=0.3\textwidth]{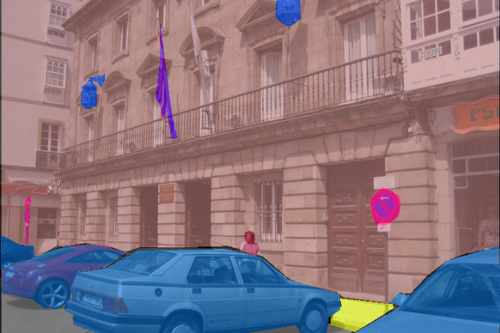} & & \includegraphics[height=0.3\textwidth]{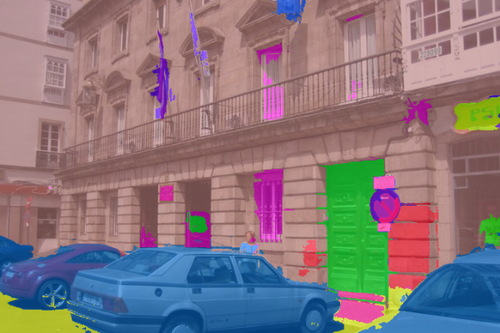} & &
            \includegraphics[height=0.3\textwidth]{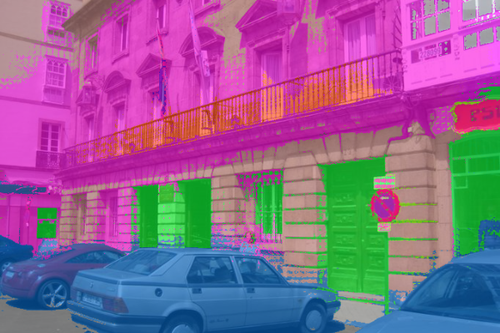} &&
            \includegraphics[height=0.3\textwidth]{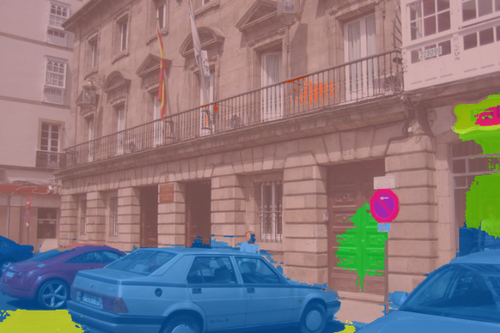} && \includegraphics[height=0.3\textwidth]{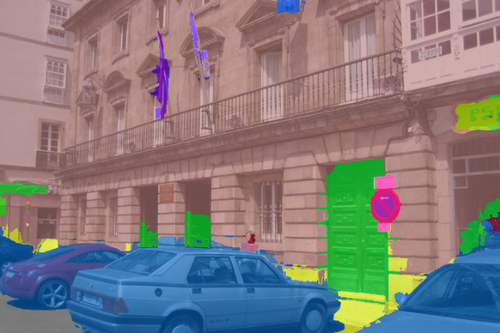} \\
        \end{tabular}
        }
    \vspace{-.2cm}
    \caption{Additional qualitative results of \ours in comparison with FreeDA~\cite{barsellotti2024training}, ProxyCLIP~\cite{lan2024proxyclip}, and CLIP-DINOiser~\cite{wysoczanska2024clip}.}
    \label{fig:qualitatives_supp}
    \vspace{-.4cm}
\end{figure*}

\tit{Comparison with State-of-the-Art Methods} Finally, in Fig.~\ref{fig:qualitatives_supp} we report a set of qualitative results on the five datasets used for the evaluation of the models, in addition to the qualitative depicted in Fig.~\ref{fig:qualitatives} of the main paper. We compare the segmentation masks of \ours with the ones of FreeDA~\cite{barsellotti2024training}, ProxyCLIP~\cite{lan2024proxyclip}, and CLIP-DINOiser~\cite{wysoczanska2024clip}, which represent our main competitors. In particular, we report a pair of images from Pascal VOC with background and eight pairs of images from Pascal Context, COCO Stuff, Cityscapes, and ADE20K, without background. As it can be seen, these qualitative results further highlight the impressive segmentation capabilities of \ours with both background and foreground categories.

\begin{figure*}[t]
    \centering
    \normalsize
        \resizebox{\linewidth}{!}{
        \setlength{\tabcolsep}{.08em}
        \begin{tabular}{c p{2mm} c p{0mm} c p{0mm} c p{0mm} c p{0mm} c p{0mm} c}
            \addlinespace[2mm]
            \textbf{\large Image} && \textbf{\large ViT-S} && \textbf{\large ViT-S Reg.} && \textbf{\large ViT-B} && \textbf{\large ViT-B Reg.} && \textbf{\large ViT-L} && \textbf{\large ViT-L Reg.} \\
            \addlinespace[1mm]
            \includegraphics[height=0.2\textwidth]{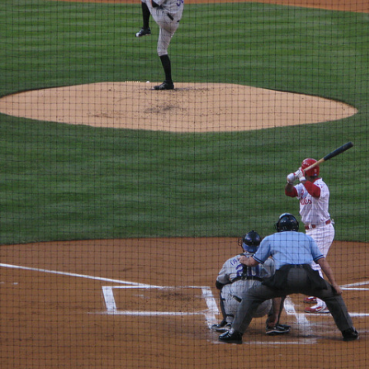} &&
            \includegraphics[height=0.2\textwidth]{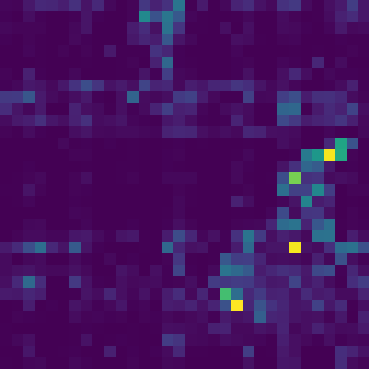} && \includegraphics[height=0.2\textwidth]{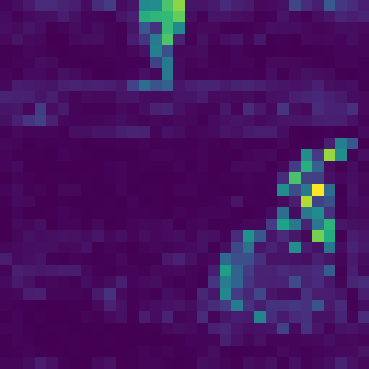} && \includegraphics[height=0.2\textwidth]{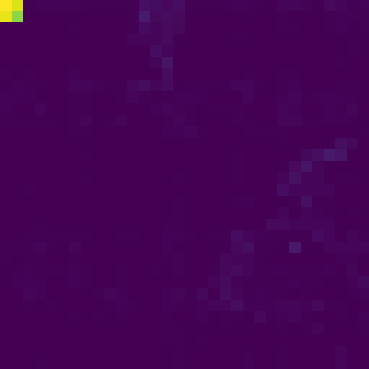} &&
            \includegraphics[height=0.2\textwidth]{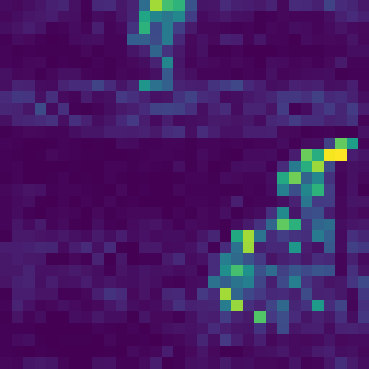} &&
            \includegraphics[height=0.2\textwidth]{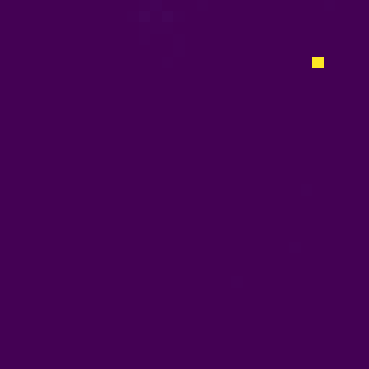} && \includegraphics[height=0.2\textwidth]{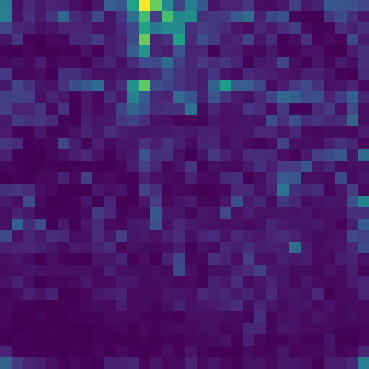} \\
        \end{tabular}
        }

    \resizebox{\linewidth}{!}{
        \setlength{\tabcolsep}{.08em}
        \begin{tabular}{c p{2mm} c p{0mm} c p{0mm} c p{0mm} c p{0mm} c p{0mm} c p{0mm} c p{0mm} c p{0mm} c p{0mm} c p{0mm} c p{0mm} c p{0mm} c p{0mm} c p{0mm} c p{0mm} c}
            \addlinespace[2mm]
            \toprule
            \addlinespace[5mm]
            \rotatebox{90}{\parbox[t]{2in}{\hspace*{\fill}\huge\textbf{ViT-S}\hspace*{\fill}}} &&
            \includegraphics[height=0.3\textwidth]{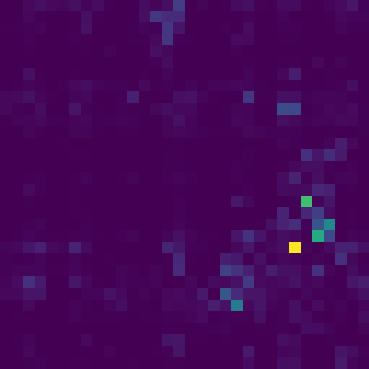} &&
            \includegraphics[height=0.3\textwidth]{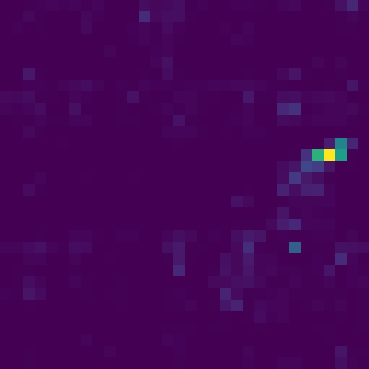} && \includegraphics[height=0.3\textwidth]{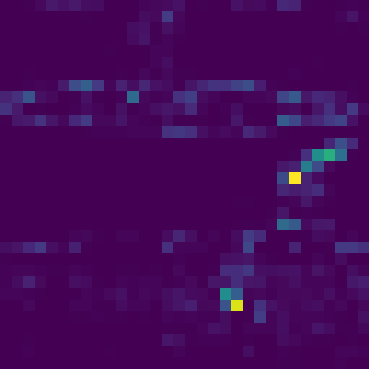} && \includegraphics[height=0.3\textwidth]{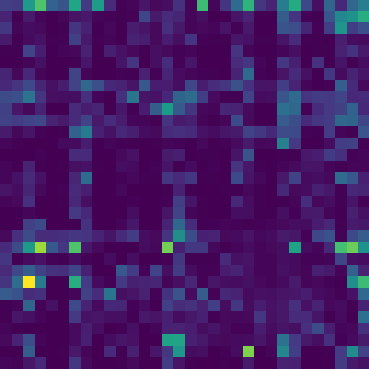} &&
            \includegraphics[height=0.3\textwidth]{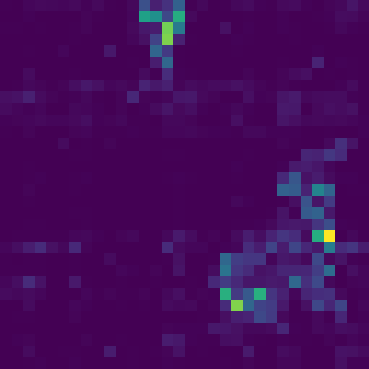} &&
            \includegraphics[height=0.3\textwidth]{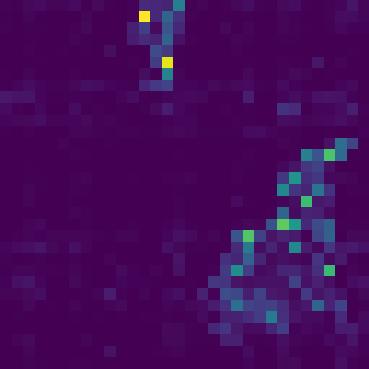}  \\
            \addlinespace[3mm]
            \rotatebox{90}{\parbox[t]{2in}{\hspace*{\fill}\huge\textbf{ViT-S Reg.}\hspace*{\fill}}} &&
            \includegraphics[height=0.3\textwidth]{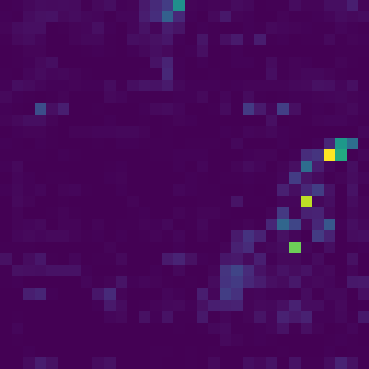} &&
            \includegraphics[height=0.3\textwidth]{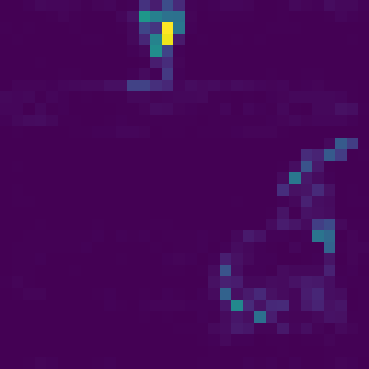} && \includegraphics[height=0.3\textwidth]{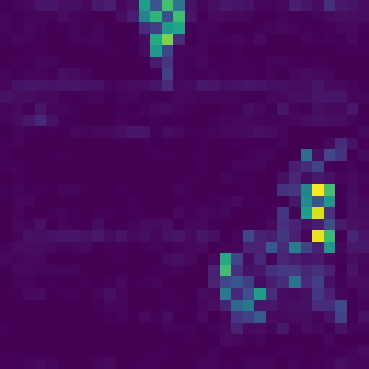} && \includegraphics[height=0.3\textwidth]{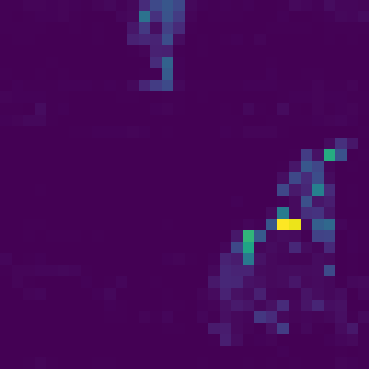} &&
            \includegraphics[height=0.3\textwidth]{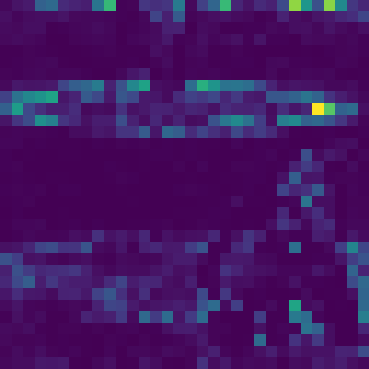} &&
            \includegraphics[height=0.3\textwidth]{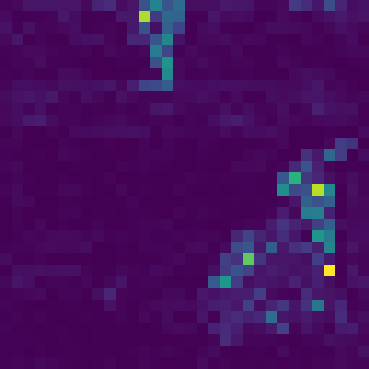}\\
            \addlinespace[5mm]
            \midrule
            \addlinespace[5mm]
            \rotatebox{90}{\parbox[t]{2in}{\hspace*{\fill}\huge\textbf{ViT-B}\hspace*{\fill}}} &&
            \includegraphics[height=0.3\textwidth]{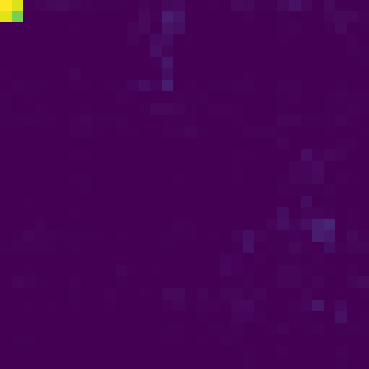} &&
            \includegraphics[height=0.3\textwidth]{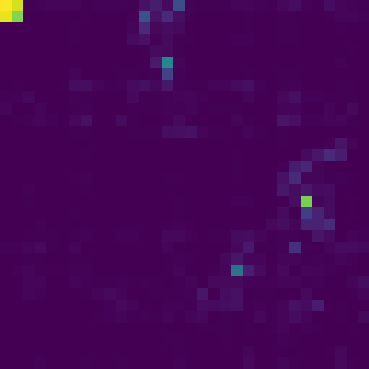} && \includegraphics[height=0.3\textwidth]{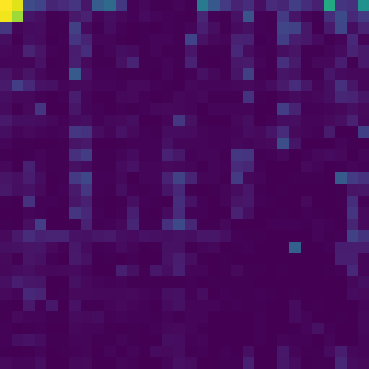} && \includegraphics[height=0.3\textwidth]{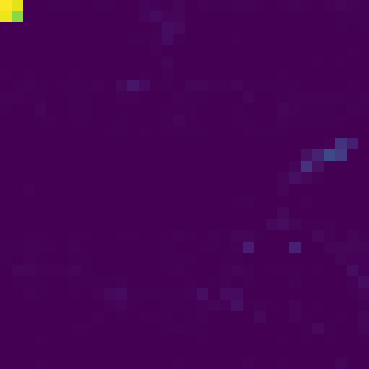} &&
            \includegraphics[height=0.3\textwidth]{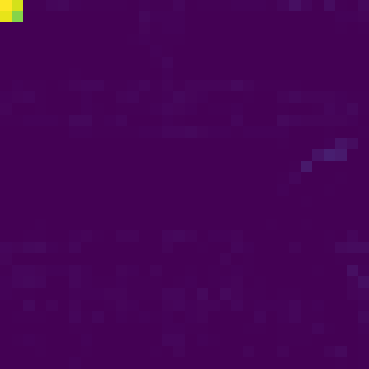} &&
            \includegraphics[height=0.3\textwidth]{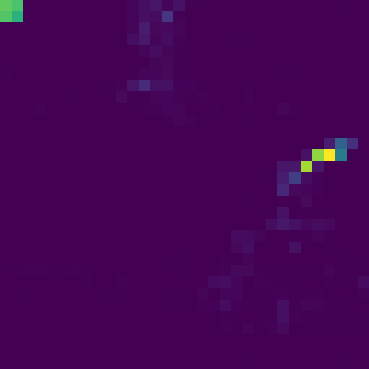} && \includegraphics[height=0.3\textwidth]{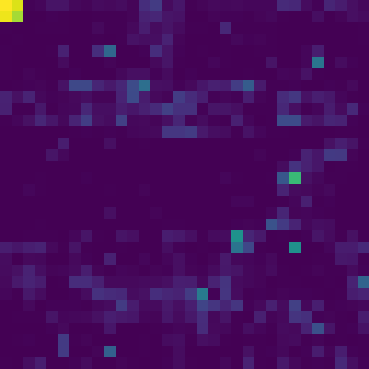} && \includegraphics[height=0.3\textwidth]{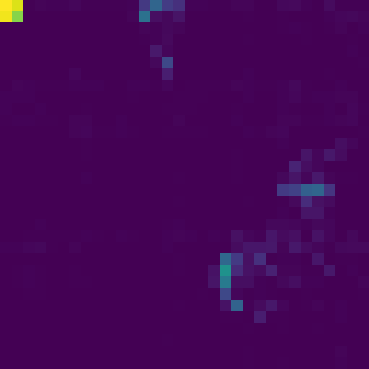}
            &&
            \includegraphics[height=0.3\textwidth]{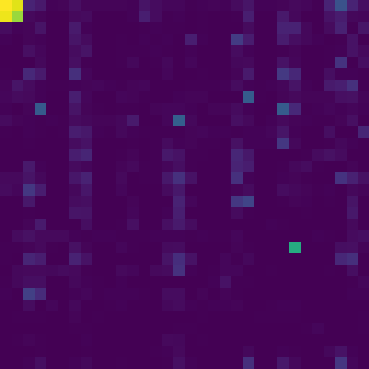} && \includegraphics[height=0.3\textwidth]{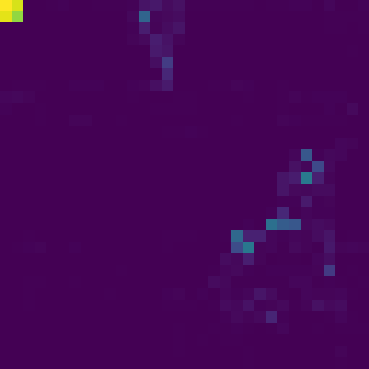} &&
            \includegraphics[height=0.3\textwidth]{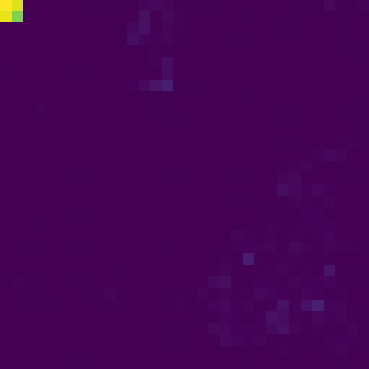} && \includegraphics[height=0.3\textwidth]{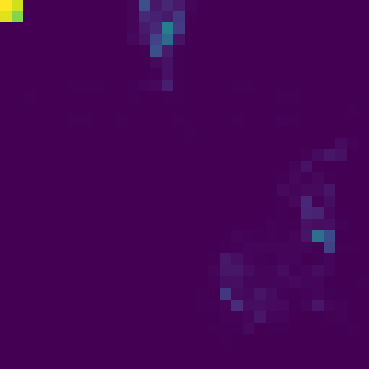} \\
            \addlinespace[3mm]
            \rotatebox{90}{\parbox[t]{2in}{\hspace*{\fill}\huge\textbf{ViT-B Reg.}\hspace*{\fill}}} &&
            \includegraphics[height=0.3\textwidth]{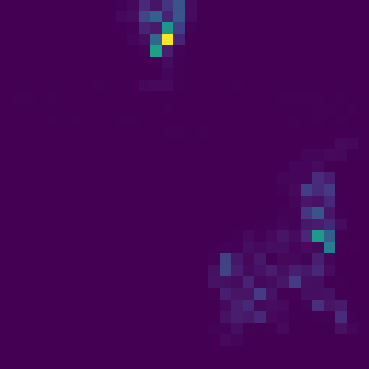} &&
            \includegraphics[height=0.3\textwidth]{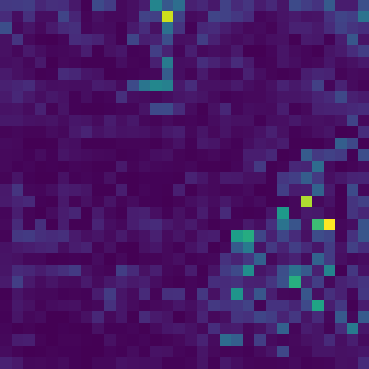} && \includegraphics[height=0.3\textwidth]{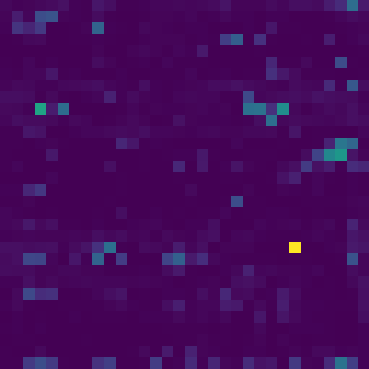} && \includegraphics[height=0.3\textwidth]{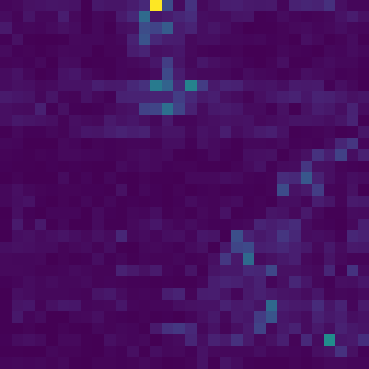} &&
            \includegraphics[height=0.3\textwidth]{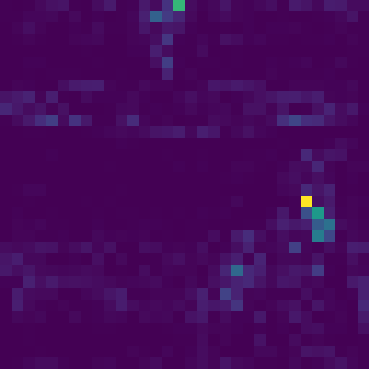} &&
            \includegraphics[height=0.3\textwidth]{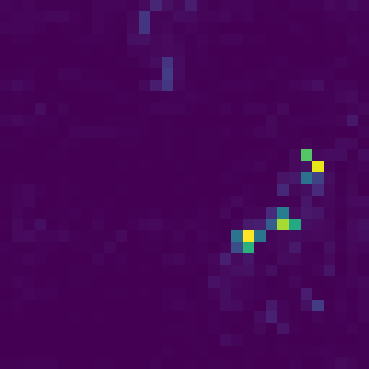} && \includegraphics[height=0.3\textwidth]{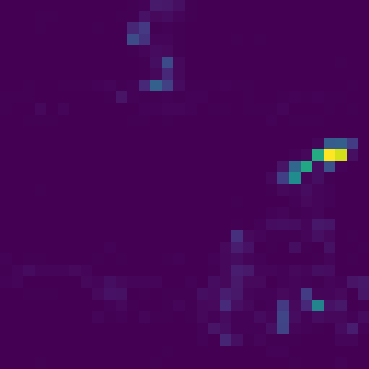} && \includegraphics[height=0.3\textwidth]{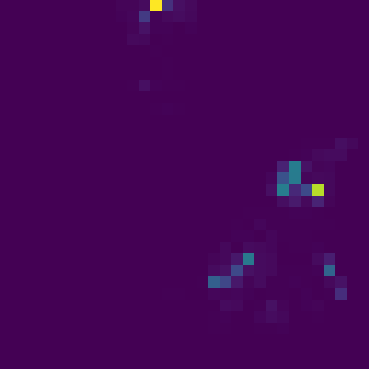}
            &&
            \includegraphics[height=0.3\textwidth]{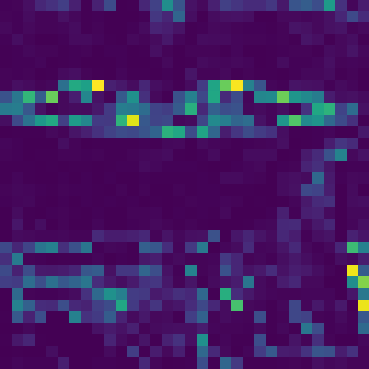} && \includegraphics[height=0.3\textwidth]{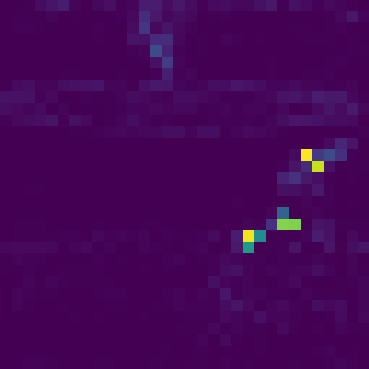} &&
            \includegraphics[height=0.3\textwidth]{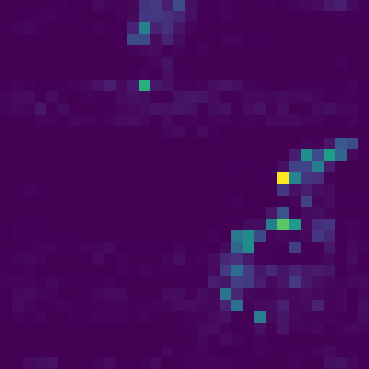} && \includegraphics[height=0.3\textwidth]{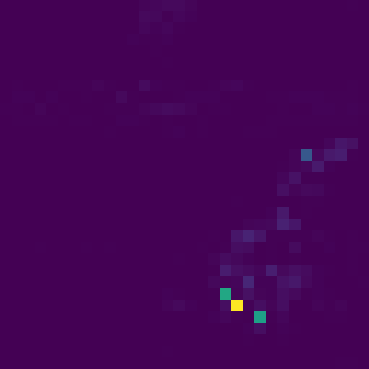} \\
            \addlinespace[5mm]
            \midrule
            \addlinespace[5mm]
            \rotatebox{90}{\parbox[t]{2in}{\hspace*{\fill}\huge\textbf{ViT-L}\hspace*{\fill}}} &&
            \includegraphics[height=0.3\textwidth]{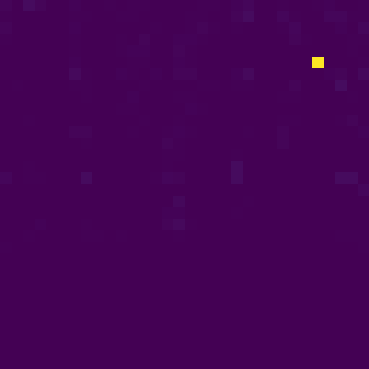} &&
            \includegraphics[height=0.3\textwidth]{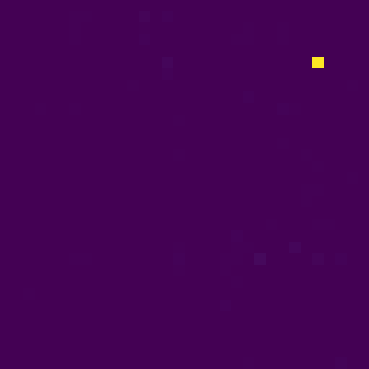} && \includegraphics[height=0.3\textwidth]{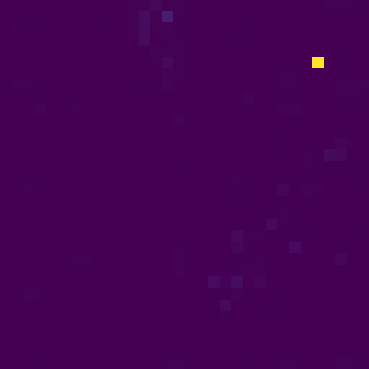} && \includegraphics[height=0.3\textwidth]{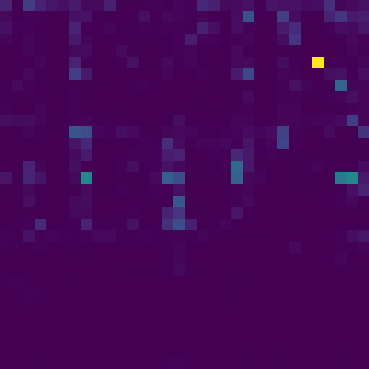} &&
            \includegraphics[height=0.3\textwidth]{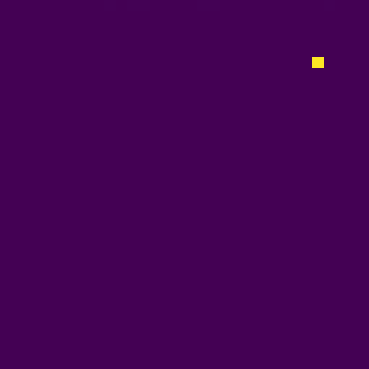} &&
            \includegraphics[height=0.3\textwidth]{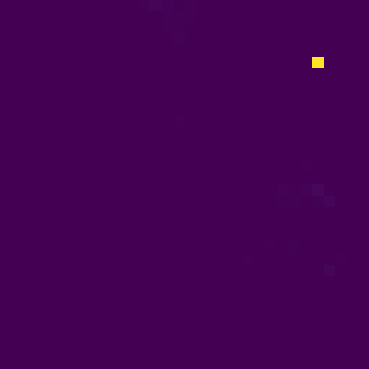} && \includegraphics[height=0.3\textwidth]{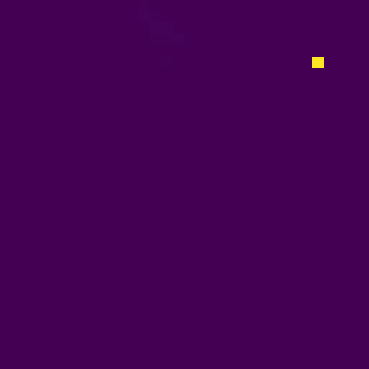} && \includegraphics[height=0.3\textwidth]{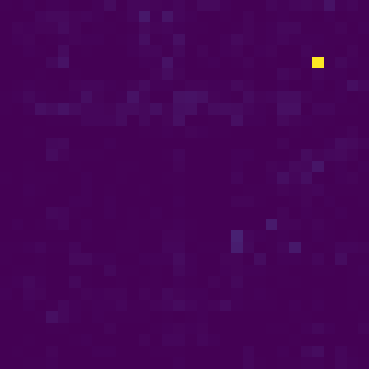} &&
            \includegraphics[height=0.3\textwidth]{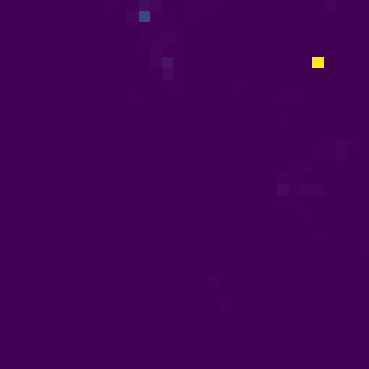} && \includegraphics[height=0.3\textwidth]{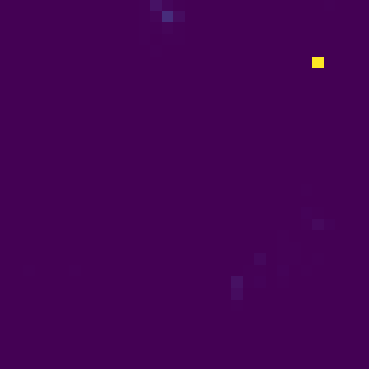} &&
            \includegraphics[height=0.3\textwidth]{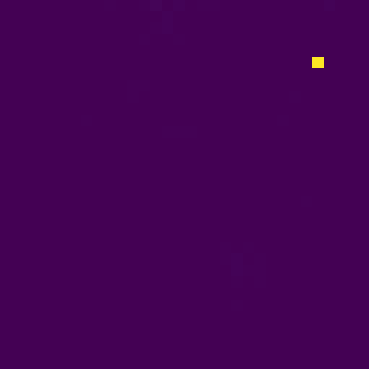} && \includegraphics[height=0.3\textwidth]{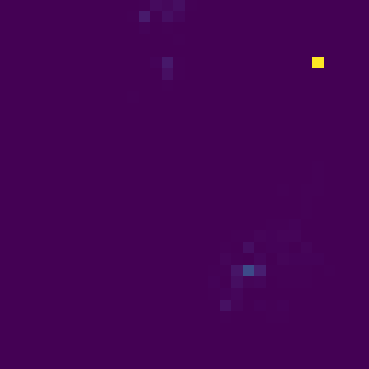} && \includegraphics[height=0.3\textwidth]{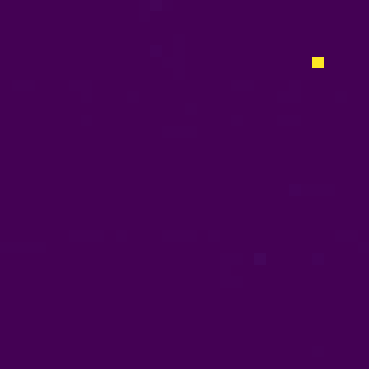} &&
            \includegraphics[height=0.3\textwidth]{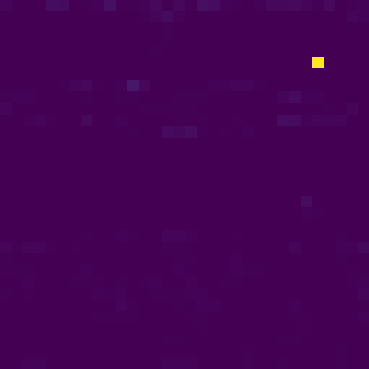} &&
            \includegraphics[height=0.3\textwidth]{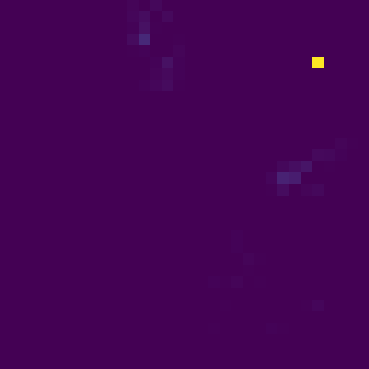} && \includegraphics[height=0.3\textwidth]{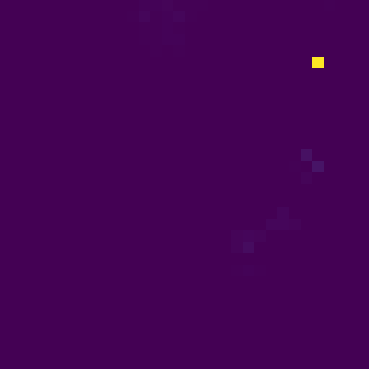} \\
            \addlinespace[3mm]
            \rotatebox{90}{\parbox[t]{2in}{\hspace*{\fill}\huge\textbf{ViT-L Reg.}\hspace*{\fill}}} &&
            \includegraphics[height=0.3\textwidth]{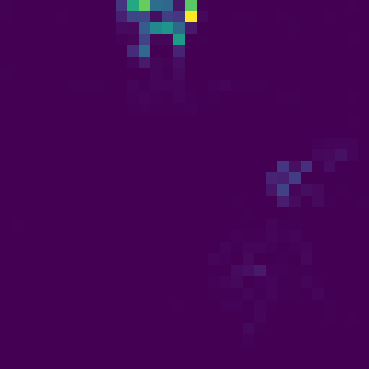} &&
            \includegraphics[height=0.3\textwidth]{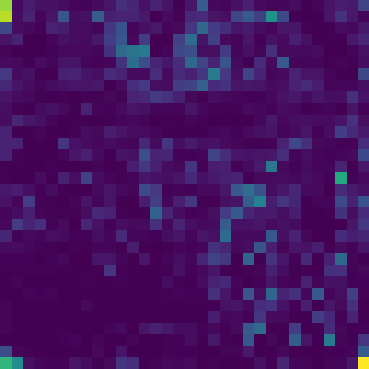} && \includegraphics[height=0.3\textwidth]{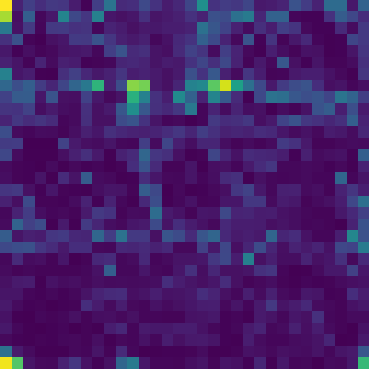} && \includegraphics[height=0.3\textwidth]{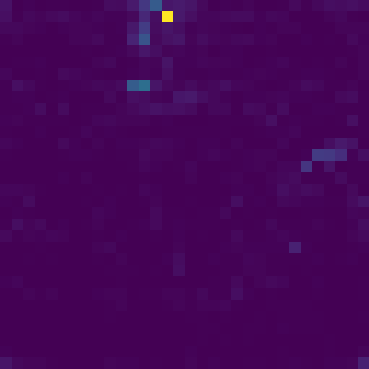} &&
            \includegraphics[height=0.3\textwidth]{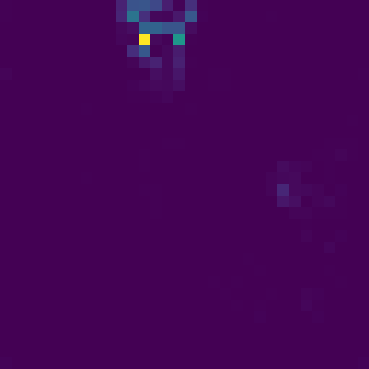} &&
            \includegraphics[height=0.3\textwidth]{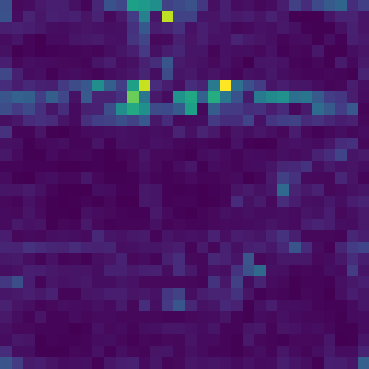} && \includegraphics[height=0.3\textwidth]{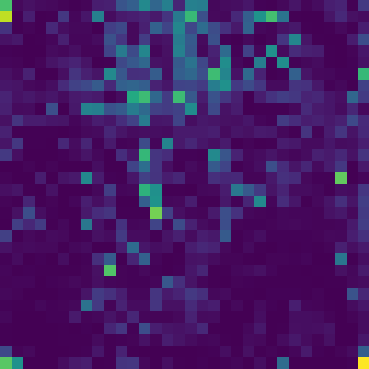} && \includegraphics[height=0.3\textwidth]{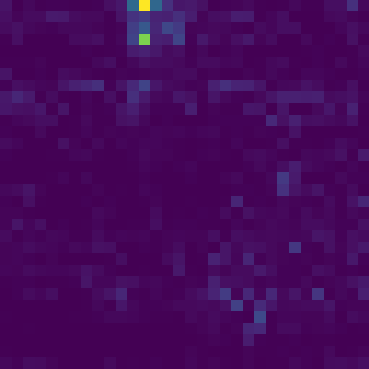} &&
            \includegraphics[height=0.3\textwidth]{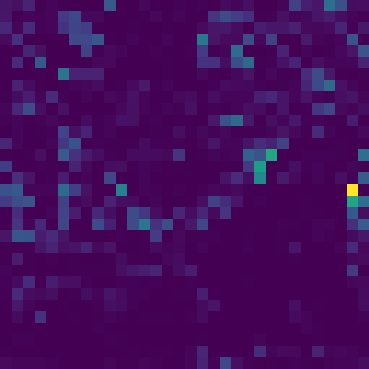} && \includegraphics[height=0.3\textwidth]{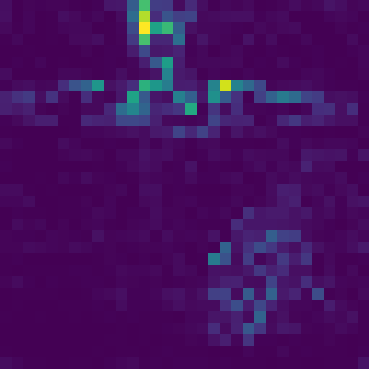} &&
            \includegraphics[height=0.3\textwidth]{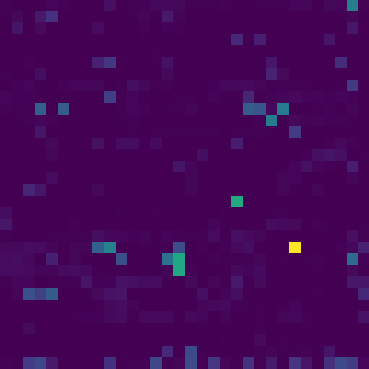} && \includegraphics[height=0.3\textwidth]{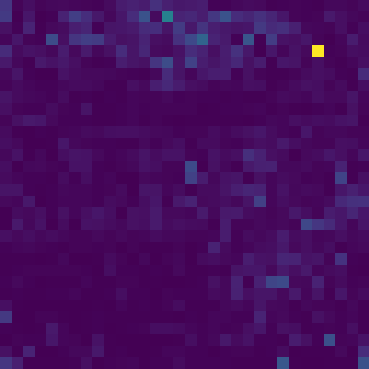} && \includegraphics[height=0.3\textwidth]{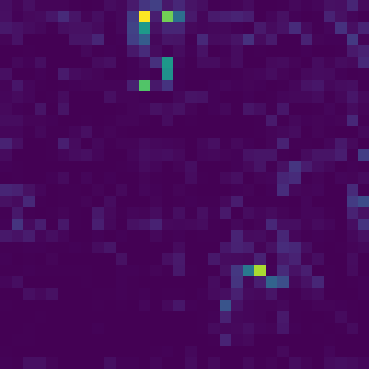} &&
            \includegraphics[height=0.3\textwidth]{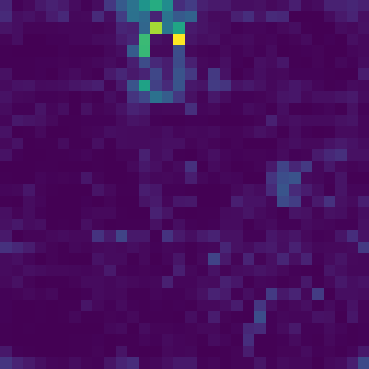} &&
            \includegraphics[height=0.3\textwidth]{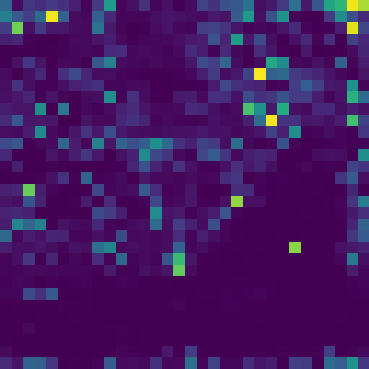} && \includegraphics[height=0.3\textwidth]{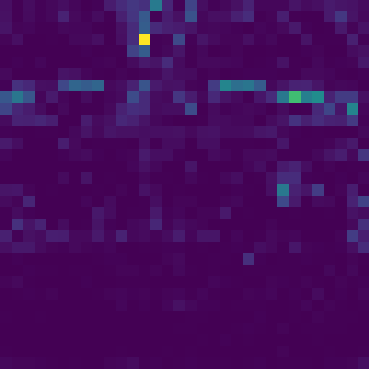} \\

        \end{tabular}
        }
    \vspace{-.15cm}
    \caption{Comparison of DINOv2 with and without registers across different visual backbones (ViT-S, ViT-B, and ViT-L). The results highlight how the ViT-B and ViT-L backbones without registers exhibit artifacts that introduce noise during the alignment process.}
    \label{fig:reg_vs_no_reg_v2}
    \vspace{-.5cm}
\end{figure*}

\begin{figure*}[t]
    \centering
    \footnotesize
    \resizebox{\linewidth}{!}{
        \setlength{\tabcolsep}{.08em}
        \begin{tabular}{c p{2mm} c p{0mm} c p{0mm} c p{0mm} c}
            \addlinespace[2mm]
            \textbf{Image} && \textbf{DINOv2} && \textbf{DINO} && \textbf{MAE} && \textbf{CLIP}\\
            \addlinespace[1mm]
            \includegraphics[height=0.2\textwidth]{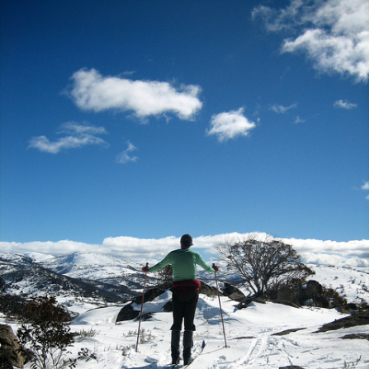} &&
            \includegraphics[height=0.2\textwidth]{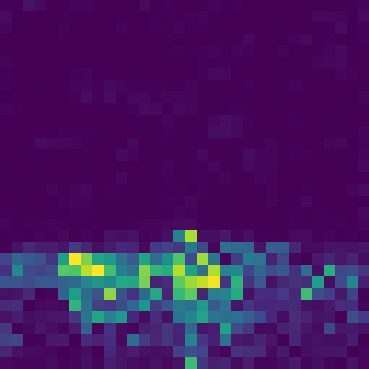} && \includegraphics[height=0.2\textwidth]{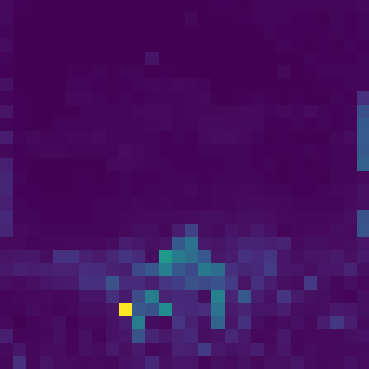} && \includegraphics[height=0.2\textwidth]{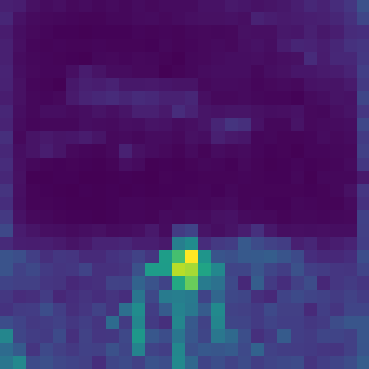} &&
            \includegraphics[height=0.2\textwidth]{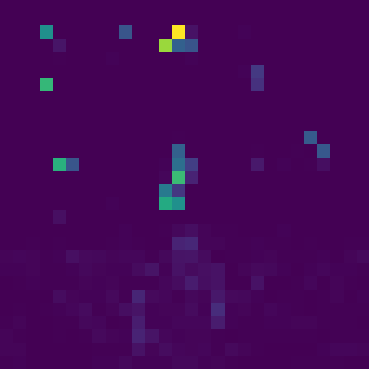} \\
        \end{tabular}
        }

    \resizebox{\linewidth}{!}{
        \setlength{\tabcolsep}{.12em}
        \begin{tabular}{c p{2mm} c p{0mm} c p{0mm} c p{0mm} c p{0mm} c p{0mm} c p{0mm} c p{0mm} c p{0mm} c p{0mm} c p{0mm} c p{0mm} c}
            \addlinespace[2mm]
            \toprule
            \addlinespace[5mm]
            \rotatebox{90}{\parbox[t]{2in}{\hspace*{\fill}\huge\textbf{DINOv2}\hspace*{\fill}}} &&
            \includegraphics[height=0.3\textwidth]{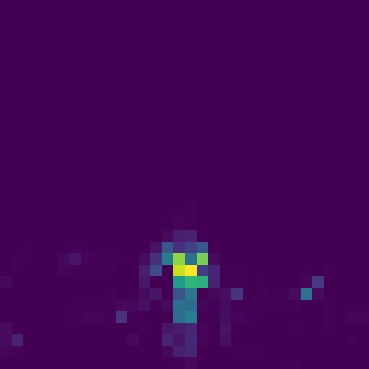} &&
            \includegraphics[height=0.3\textwidth]{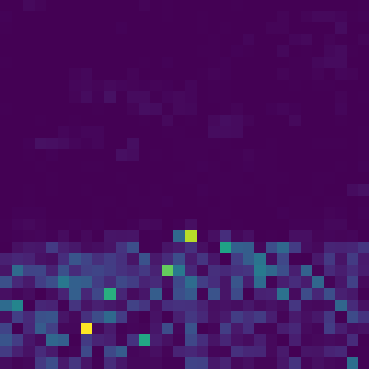} && \includegraphics[height=0.3\textwidth]{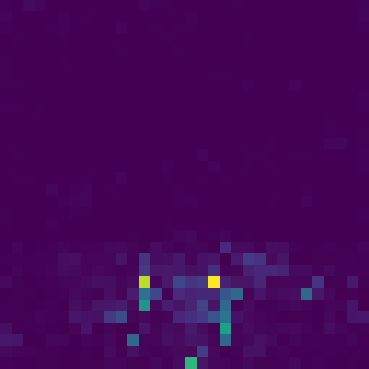} && \includegraphics[height=0.3\textwidth]{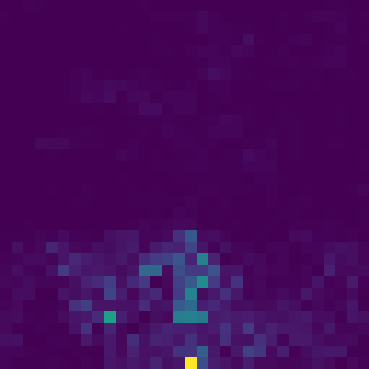} &&
            \includegraphics[height=0.3\textwidth]{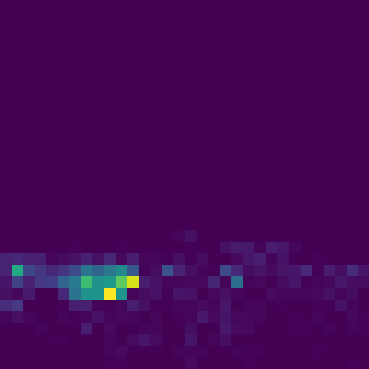} &&
            \includegraphics[height=0.3\textwidth]{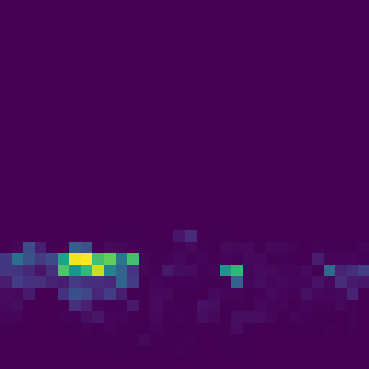} && \includegraphics[height=0.3\textwidth]{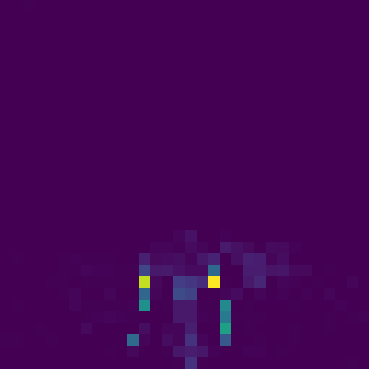} && \includegraphics[height=0.3\textwidth]{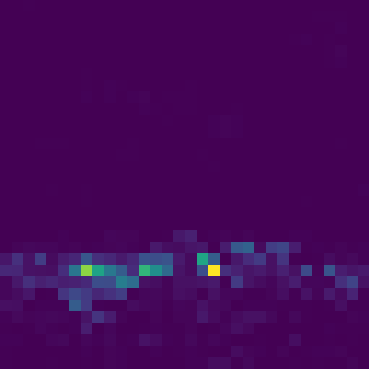}
            &&
            \includegraphics[height=0.3\textwidth]{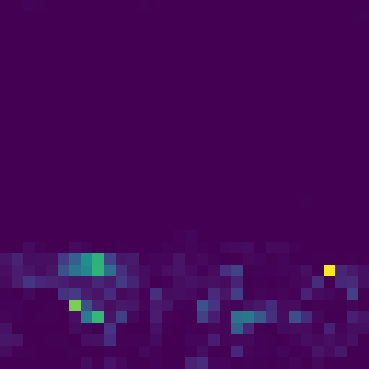} && \includegraphics[height=0.3\textwidth]{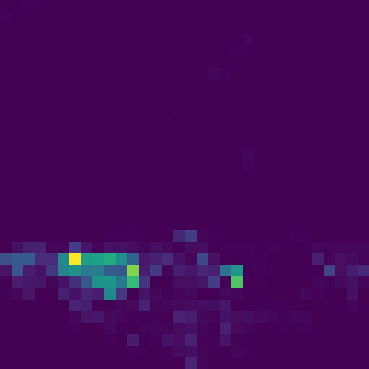} &&
            \includegraphics[height=0.3\textwidth]{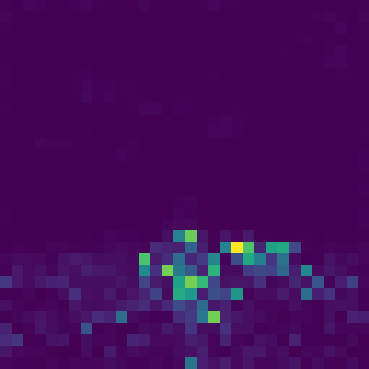} && \includegraphics[height=0.3\textwidth]{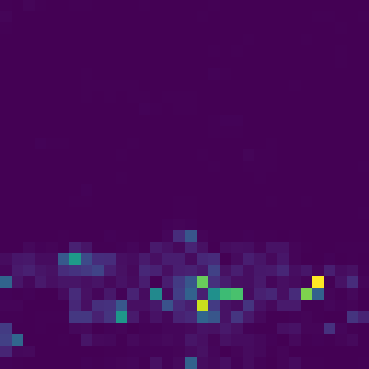} \\
            \addlinespace[5mm]
            \midrule
            \addlinespace[5mm]
            \rotatebox{90}{\parbox[t]{2in}{\hspace*{\fill}\huge\textbf{DINO}\hspace*{\fill}}} &&
            \includegraphics[height=0.3\textwidth]{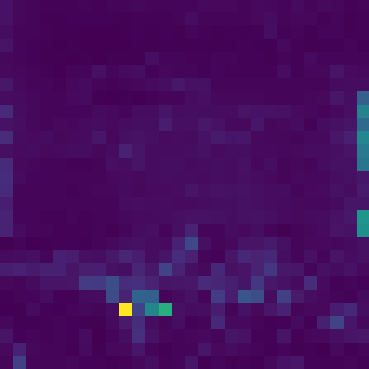} &&
            \includegraphics[height=0.3\textwidth]{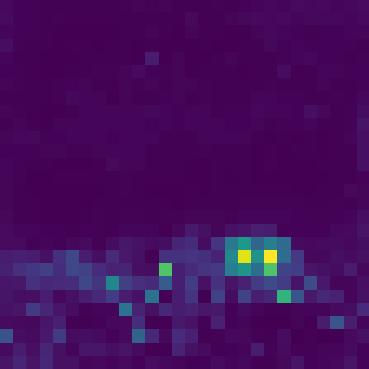} &&
            \includegraphics[height=0.3\textwidth]{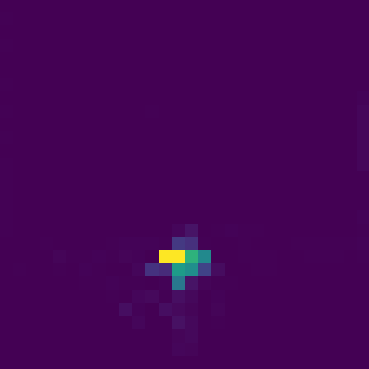} && \includegraphics[height=0.3\textwidth]{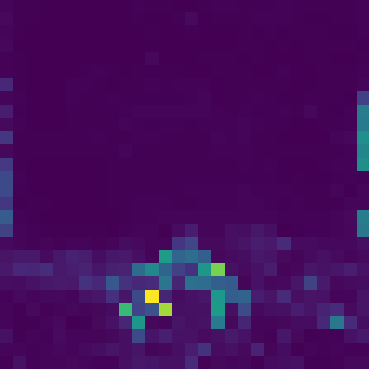} &&
            \includegraphics[height=0.3\textwidth]{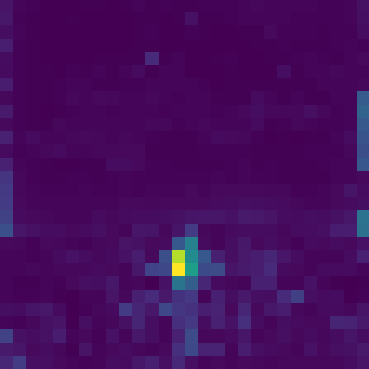} &&
            \includegraphics[height=0.3\textwidth]{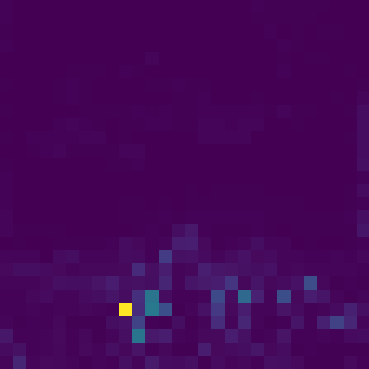} && \includegraphics[height=0.3\textwidth]{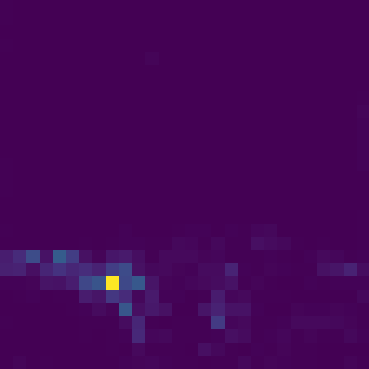} && \includegraphics[height=0.3\textwidth]{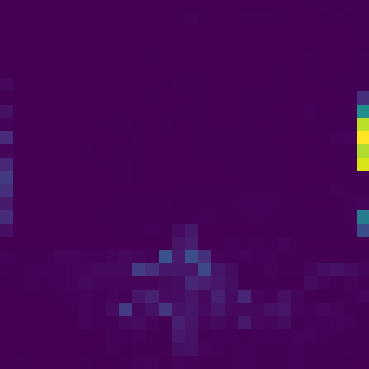}
            &&
            \includegraphics[height=0.3\textwidth]{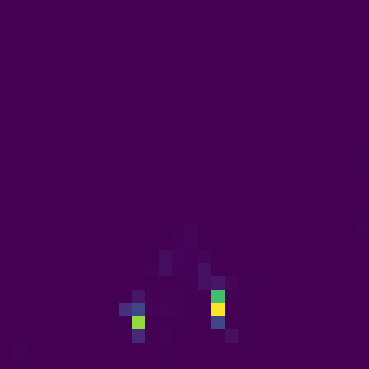} && \includegraphics[height=0.3\textwidth]{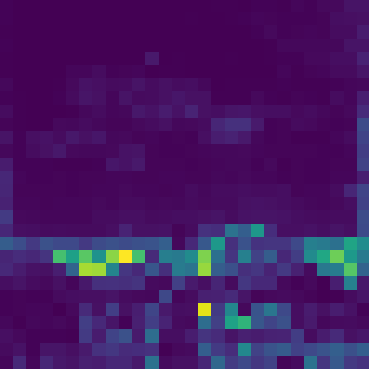} &&
            \includegraphics[height=0.3\textwidth]{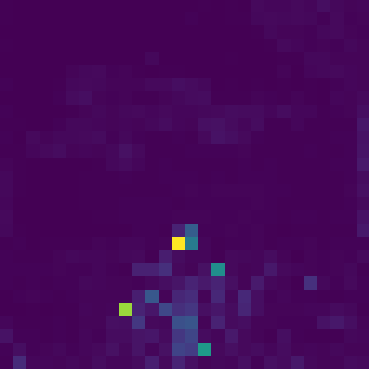} && \includegraphics[height=0.3\textwidth]{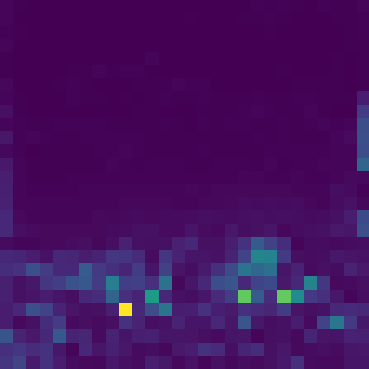} \\
            \addlinespace[5mm]
            \midrule
            \addlinespace[5mm]
            \rotatebox{90}{\parbox[t]{2in}{\hspace*{\fill}\huge\textbf{MAE}\hspace*{\fill}}} &&
            \includegraphics[height=0.3\textwidth]{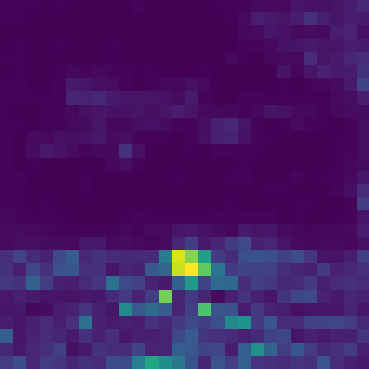} &&
            \includegraphics[height=0.3\textwidth]{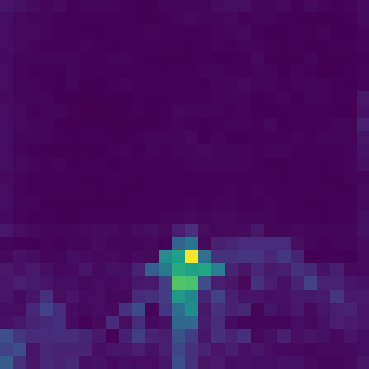} && \includegraphics[height=0.3\textwidth]{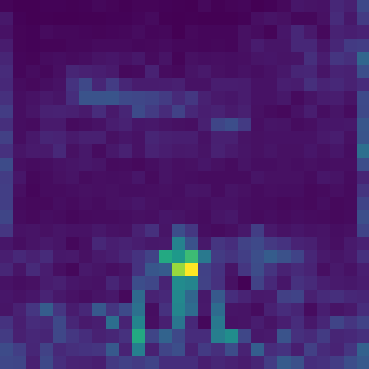} && \includegraphics[height=0.3\textwidth]{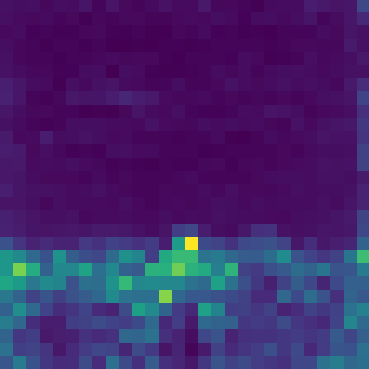} &&
            \includegraphics[height=0.3\textwidth]{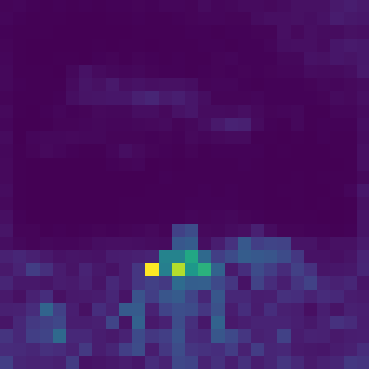} &&
            \includegraphics[height=0.3\textwidth]{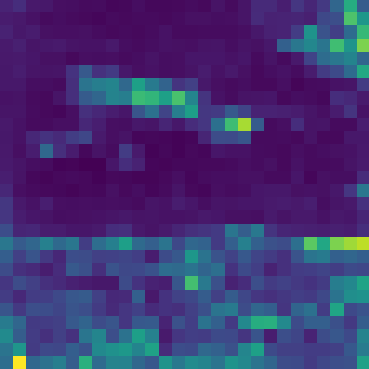} && \includegraphics[height=0.3\textwidth]{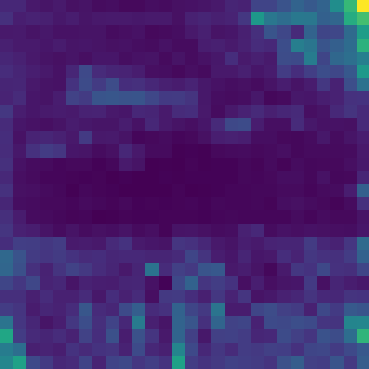} && \includegraphics[height=0.3\textwidth]{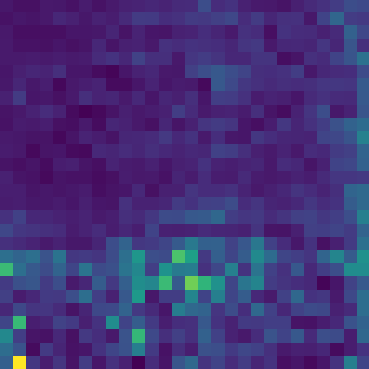}
            &&
            \includegraphics[height=0.3\textwidth]{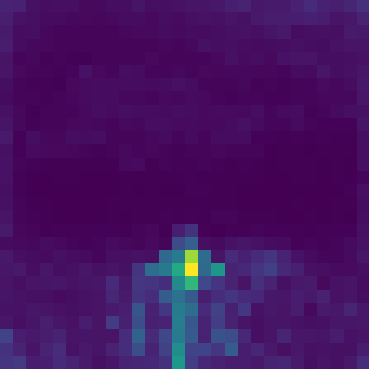} && \includegraphics[height=0.3\textwidth]{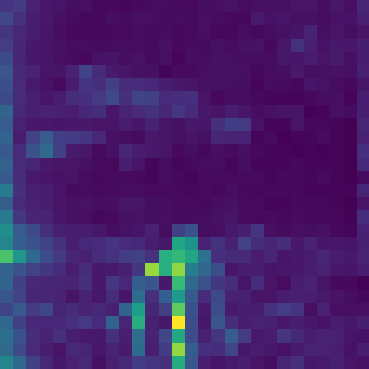} &&
            \includegraphics[height=0.3\textwidth]{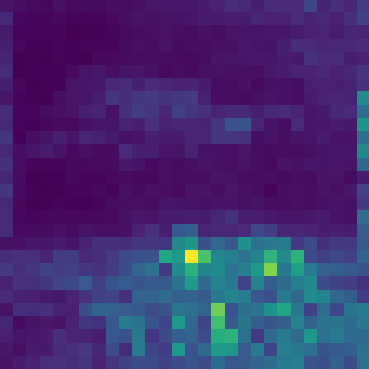} && \includegraphics[height=0.3\textwidth]{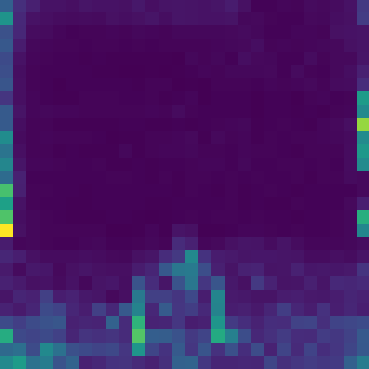} \\
            \addlinespace[5mm]
            \midrule
            \addlinespace[5mm]
            \rotatebox{90}{\parbox[t]{2in}{\hspace*{\fill}\huge\textbf{CLIP}\hspace*{\fill}}} &&
            \includegraphics[height=0.3\textwidth] {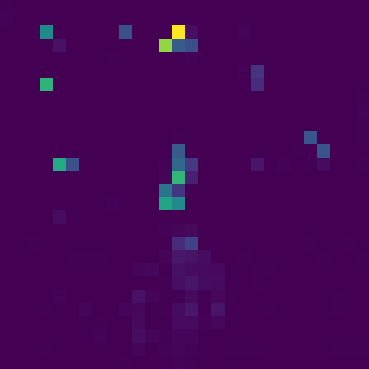} &&
            \includegraphics[height=0.3\textwidth]{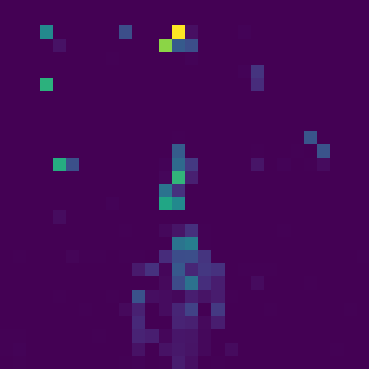} && \includegraphics[height=0.3\textwidth]{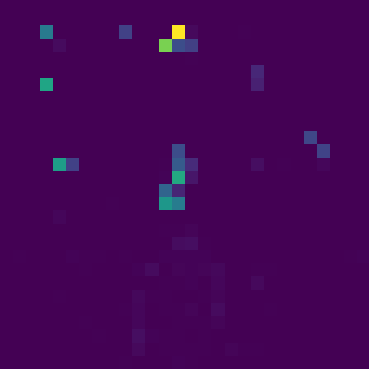} && \includegraphics[height=0.3\textwidth]{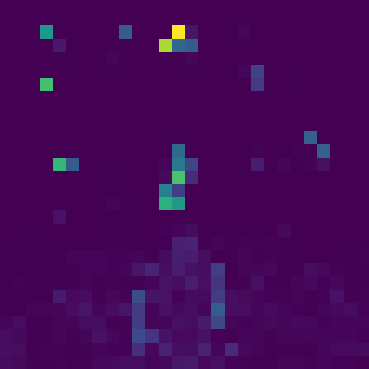} &&
            \includegraphics[height=0.3\textwidth]{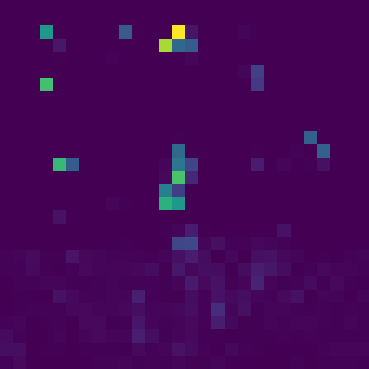} &&2
            \includegraphics[height=0.3\textwidth]{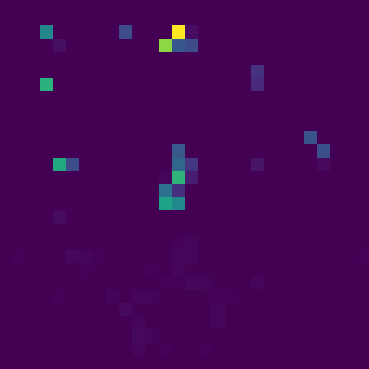} && \includegraphics[height=0.3\textwidth]{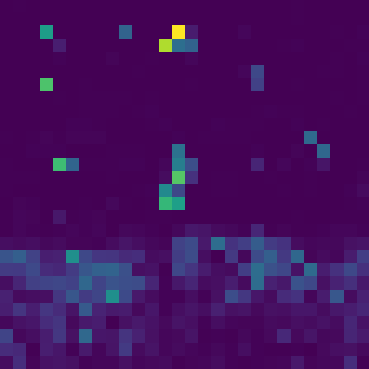} && \includegraphics[height=0.3\textwidth]{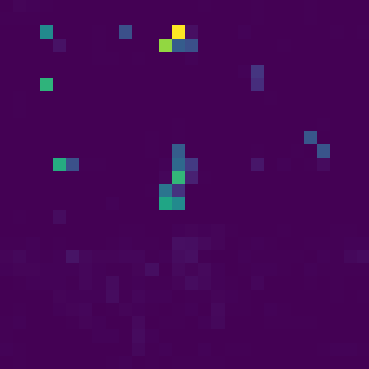}
            &&
            \includegraphics[height=0.3\textwidth]{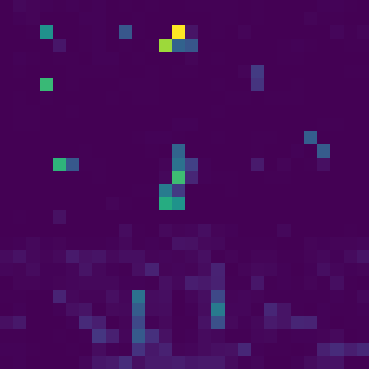} && \includegraphics[height=0.3\textwidth]{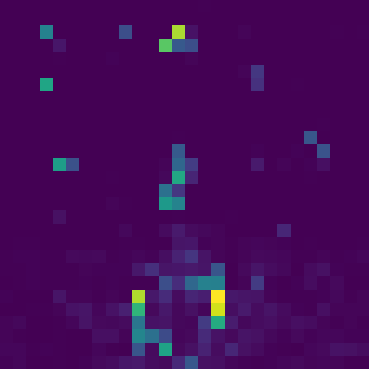} &&
            \includegraphics[height=0.3\textwidth]{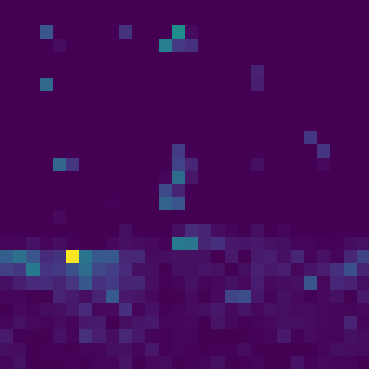} && \includegraphics[height=0.3\textwidth]{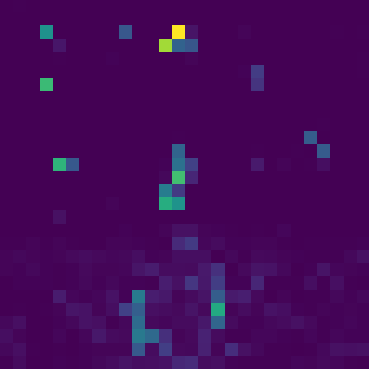} \\

        \end{tabular}
        }
    \vspace{-.15cm}
    \caption{Self-attention activations of different visual backbones (\ie,  DINOv2, DINO, MAE, CLIP).}
    \label{fig:backbone_attn}
    \vspace{-.4cm}
\end{figure*}

\end{document}